\documentclass[preprint,12pt]{elsarticle}

\usepackage{multirow}
\usepackage{amsmath,amssymb,amsfonts}
\usepackage{textcomp}
\usepackage[dvipsnames]{xcolor}
\usepackage{booktabs}
\usepackage{lineno}

\newcommand{\rev}[1]{\textcolor{black}{#1}}
\newenvironment{revblock}{\color{black}}{}
\newcommand{\limit}[1]{\textcolor{black}{#1}}

\journal{Robotics and Autonomous Systems}

\begin{document}

\begin{frontmatter}

\title{CineWild: Balancing Art and Robotics for Ethical Wildlife Documentary Filmmaking}
\author[upo]{Pablo Pueyo\corref{cor1}}
\ead{ppueyor@upo.es}
\author[upo]{Fernando Caballero}
\ead{fcaballero@upo.es}
\author[unizar]{Ana Cristina Murillo}
\ead{acm@unizar.es}
\author[unizar]{Eduardo Montijano}
\ead{emonti@unizar.es}
\cortext[cor1]{Corresponding author.}

\affiliation[upo]{organization={Service Robotics Lab, Universidad Pablo de Olavide},
            city={Seville},
            country={Spain}}
\affiliation[unizar]{organization={I3A, Universidad de Zaragoza},
            city={Zaragoza},
            country={Spain}}
            


\begin{abstract}
Unmanned aerial vehicles (UAVs) have become powerful tools across domains from industry to the arts. In documentary filmmaking, they offer dynamic, otherwise unreachable perspectives, transforming how stories are told. Wildlife documentaries especially benefit, yet UAVs also raise ethical concerns like the risk of disturbing the animals they aim to capture. This paper introduces CineWild, a model predictive control (MPC) formulation for autonomous aerial cinematography that combines robotics, cinematography, and ethics. Building on an MPC solution that jointly controls drone motion and camera focal length, CineWild dynamically adjusts flight paths and camera settings to balance cinematic quality with animal welfare. Key features include adaptive zoom for filming from acoustic and visual safe distances, path-planning that avoids an animal’s field of view, and smooth, low-noise maneuvers. CineWild exemplifies interdisciplinary innovation, bridging engineering, visual storytelling, and environmental ethics. 
We evaluate the formulation in simulation, with ablation studies and a comparison against a cinematographic baseline and a welfare-aware baseline under identical weights, and we demonstrate the perception and camera-actuation pipeline in real-world proof-of-concept trials. We will release the code upon acceptance.
\end{abstract}



\end{frontmatter}


\begin{figure}[!b]
\centering
\includegraphics[width=1\columnwidth]{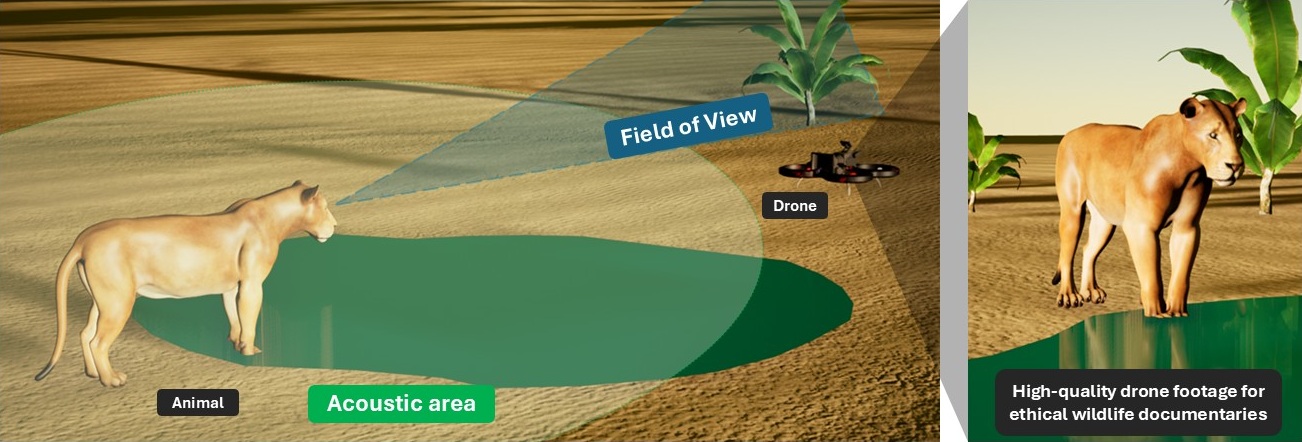} 
\caption{\footnotesize{\textbf{Overview of the CineWild system.} CineWild enables respectful wildlife storytelling by integrating drone-based cinematography with awareness of animals’ visual and acoustic sensing zones. The left panel shows a lion, our subject, within a green acoustic zone, where engine noise could cause disruption. The blue area represents the animal’s field of view, which CineWild avoids the drone from entering to prevent visual stress. The right panel features high-resolution footage captured from a safe distance, illustrating how the system preserves both cinematic quality and animal welfare. By navigating out of sensory disturbance zones, CineWild supports a more ethical and artistic approach to wildlife documentary filmmaking.}}
\end{figure}

\section{Introduction}
Drones have become essential tools for visual storytelling and environmental documentation, providing access to aerial perspectives in complex environments~\cite{kratky2021autonomous}. In the field of wildlife cinematography, UAVs offer a flexible and cost‑effective alternative to traditional methods such as camera cranes or manned aerial surveys, enabling the recording of animal behavior with reduced logistical effort~\cite{mitman2012reel, linchant2015unmanned}.

Despite these advantages, drone‑based wildlife filming raises a critical challenge: UAVs can disturb the animals being documented. Acoustic noise, abrupt motion, and visual intrusion may induce stress or behavioral changes, particularly in sensitive habitats or among endangered species~\cite{christie2016unmanned, mulero2017unmanned}. Although recent advances in autonomous UAVs have improved visual quality, ease of use, and data collection capabilities~\cite{schad2023opportunities}, most cinematographic platforms do not explicitly model animal welfare, typically relying on conservative operating guidelines rather than embedding ethical constraints directly into the control formulation. In particular, existing approaches rarely exploit autonomous control of camera intrinsics to decouple visual quality from physical proximity.

In this paper, we introduce \textit{CineWild}, an autonomous aerial cinematography framework that explicitly integrates animal welfare into the control loop by jointly controlling drone motion and camera parameters. Central to this approach is the automatic regulation of the camera focal length, which enables adaptive zoom to preserve framing and visual quality while maintaining safe acoustic and visual distances from the animal.  CineWild formulates wildlife filming as a nonlinear optimal control problem\rev{, solved as a model predictive controller (MPC),} that incorporates disturbance‑aware objectives derived from biological studies, accounting for proximity, visibility, and motion smoothness. While informed by prior work on autonomous cinematography~\cite{pueyo2024cinempc}, this  framework goes beyond existing approaches by formulating ethical considerations and camera intrinsic control directly into the optimization process. \begin{revblock}
CineWild keeps the state, dynamics and cinematographic objectives of CineMPC~\cite{pueyo2024cinempc}; its contributions are the welfare-aware formulation and its evaluation:
\begin{itemize}
\item A welfare-aware MPC formulation for Ethical
Wildlife Documentary Filmmaking, with cost terms and hard constraints for proximity, visibility and motion smoothness. 
\item A species-parametrised model of the disturbance zones (safe, acoustic and visual distances, animal field of view).
\item Simulation experiments demonstrating welfare-aware filming: CineWild keeps its distance from the animal, reducing disturbing noise, and stays out of its field of view while preserving the requested shots. The solution is compared with a cinematographic baseline and a baseline with a safety distance and adaptive zoom.
\item An analysis of the formulation: ablation study to demonstrate the contribution of each welfare term, the choice of prediction horizon, the robustness to the start pose and the sensitivity to the weights and species thresholds, and the recovery when the animal turns its head.
\item A real-world proof-of-concept trial of the perception and camera-actuation pipeline.

\end{itemize}

\rev{These results show that the disturbance caused by an autonomous filming drone can be reduced within the controller itself, compensating physical proximity for zoom, while preserving the composition of the shot. The rest of the paper reviews related work (Sec.~2), presents the formulation (Sec.~3) and reports the simulation and real-world experiments (Sec.~4).}

\end{revblock}
\section{Related Work}

\rev{\paragraph{Wildlife monitoring with UAVs}}
Traditional wildlife documentation has long relied on direct observation, manual tracking, and fixed recording systems such as camera traps and audio recorders, enabling passive data collection and long-term monitoring~\cite{cutler1999using}. However, these tools are typically stationary and require frequent maintenance\rev{, which can itself disturb wildlife}~\cite{burton2022wildlife}. These limitations have motivated the adoption of mobile sensing platforms capable of collecting data over larger areas while reducing direct human presence. Among such technologies, Unmanned Aerial Vehicles (UAVs), or drones, have emerged as a particularly promising solution due to their flexibility, rapid deployment, and ability to access remote or difficult terrain. Drones can efficiently survey large regions while minimizing ground-level disturbance, making them valuable tools for wildlife observation~\cite{chabot2015methods}. Autonomous drones can execute pre-programmed missions without direct human control~\cite{han2015use}, which enhances operational safety and efficiency, particularly in remote or hazardous environments~\cite{hodgson2013unmanned, bonatti2019towards}. Studies have explored computer vision techniques for species detection and tracking~\cite{aliane2025drones}, and drones have been successfully deployed to record mammals, birds, and terrestrial fauna in otherwise inaccessible regions~\cite{shah2020multidrone, hodgson2013unmanned}.

\rev{\paragraph{Drone disturbance and its mitigation}}
However, obtaining visually compelling footage often requires operating close to animals, creating a conflict between image quality and non-intrusive observation. The presence of drones, especially at low altitudes, can trigger flight responses, stress, or even nest abandonment~\cite{ditmer2015bears, afridi2025impact}. Propeller noise from multirotor UAVs has been shown to influence animal behavior and can provoke noticeable behavioral changes~\cite{bennitt2022behavioral, mulero2017unmanned, afridi2026wildlife}. These findings have raised growing ethical concerns regarding the use of aerial platforms for wildlife filming and ecological research. To mitigate UAV-induced disturbance, several strategies have been proposed. Technical solutions include quieter drone designs, sound-dampening materials, and flight scheduling during periods of low animal activity~\cite{christie2016practical}. Researchers have also investigated behavioral thresholds to establish safe operating distances~\cite{duporge2021determination}. \rev{These guidelines are applied by the pilot; they are not part of how the drone or the camera is controlled.}

\begin{revblock}

As UAV technology has matured, research has increasingly focused on autonomous navigation, perception, and decision-making, and one of the most recent directions is aerial cinematography. Although primarily developed for filming human actors and general targets, these approaches provide a natural foundation for autonomous wildlife documentary production. Optimization-based methods plan drone and gimbal motion to satisfy framing and viewpoint objectives, for example with real-time MPC for single and multiple drones~\cite{nageli2017real}, with occlusion- and obstacle-aware planning in unstructured outdoor scenes~\cite{bonatti2019towards}, or with multi-UAV formations that coordinate filming and lighting~\cite{kratky2021autonomous}. CineMPC~\cite{pueyo2024cinempc} additionally controls the camera intrinsics (focal length, focus distance and aperture). While these approaches successfully automate camera operation and shot composition, and keep a safety distance to the filmed subject, none explicitly models the disturbance generated by the filming platform on the recorded subject. Table~\ref{tab:comparison} summarises the main differences.

\begin{table}[!t]
\centering
\caption{Qualitative comparison with autonomous aerial cinematography systems.}
\label{tab:comparison}
\footnotesize
\begin{tabular}{lccccc}
\toprule
 & Wildlife & Welfare terms & Zoom control & Subject FoV & Proximity  \\
\midrule
N\"ageli et al.~\cite{nageli2017real} & -- & -- & -- & -- & safety only \\
Bonatti et al.~\cite{bonatti2019towards} & -- & -- & -- & -- & safety only \\
Kr\'atk\'y et al.~\cite{kratky2021autonomous} & -- & -- & -- & -- & safety only \\
CineMPC~\cite{pueyo2024cinempc} & -- & -- & \checkmark & -- & safety only \\
CineWild (ours) & \checkmark & \checkmark & \checkmark & \checkmark & \checkmark \\
\bottomrule
\end{tabular}
\end{table}

CineWild embeds wildlife-aware objectives and constraints directly into an autonomous cinematography controller that optimizes both the drone motion and the camera intrinsics. By jointly considering cinematographic objectives, zoom control, subject visibility, and species-specific proximity zones, it enables detailed wildlife cinematography  while minimizing disturbance.
\end{revblock}

\section{Methodology}

CineWild is implemented within an MPC (Model Predictive Control) framework for autonomous aerial cinematography, where control inputs are optimized over a finite prediction horizon $N$. At each timestep $k$, the controller computes drone and camera inputs, $\mathbf{u}_{d,k}$ and $\mathbf{u}_{c,k}$, to jointly regulate camera extrinsic and intrinsic parameters in accordance with cinematographic objectives, while accounting for ethical constraints relevant to wildlife filming.
For perception, we reuse an existing pose‑estimation module to obtain the animal’s position and orientation, following prior validation in related work~\cite{pueyo2024cinempc}. Limitations associated with perception uncertainty, such as rapid head movements~\cite{wilson2020orientation}, are acknowledged but are beyond the scope of this paper.

\subsection{State Representation}
\label{sec:solution_state}
We retain the state formulation of the base method. The drone state $\mathbf{x}_{d,k}$, corresponding to the camera extrinsics, consists of position $\mathbf{p}_{d,k}\in\mathbb{R}^3$, velocity $\mathbf{v}_{d,k}\in\mathbb{R}^3$, and gimbal orientation $\mathbf{R}_{d,k}\in SO(3)$, which is decoupled from the drone body. The camera state is given by the focal length $f_k$, controlling optical zoom:
\begin{equation}
\mathbf{x}_{d,k} = (\mathbf{p}_{d,k},\,\mathbf{v}_{d,k},\,\mathbf{R}_{d,k}), \qquad
\mathbf{x}_{c,k} = (f_k).
\end{equation}

We consider a single target animal with state
\[
\mathbf{x}_{t,k} = (\mathbf{p}_{t,k},\,\mathbf{v}_{t,k},\,\mathbf{R}_{t,k}),
\]
where $\mathbf{p}_{t,k}\in\mathbb{R}^3$, $\mathbf{v}_{t,k}\in\mathbb{R}^3$, and $\mathbf{R}_{t,k}\in SO(3)$ denote its position, velocity, and orientation in the global frame.

The drone--target distance and relative rotation are
\[
d_{dt,k} = \|\mathbf{p}_{d,k}-\mathbf{p}_{t,k}\|, \qquad \mathbf{R}_{dt,k} = \mathbf{R}_{d,k}^\top \mathbf{R}_{t,k}.
\]
\subsection{Control Inputs}
\label{sec:control}
The MPC optimizes control actions that include the drone’s linear acceleration $\mathbf{a}_{d,k}$ and gimbal angular velocity $\mathbf{\Omega}_{d,k}$. For the camera, the MPC outputs the focal length velocity, $v_{f,k}$,
\begin{equation*}
\mathbf{u}_{d,k} = (\mathbf{a}_{d,k}, \mathbf{\Omega}_{d,k})\rev{,} \qquad 
\mathbf{u}_{c,k} = (v_{f,k}).
\end{equation*}

\subsection{System Dynamics}

The evolution of the drone and camera states is described by
a discrete-time model, detailed below for the translational,
rotational, and camera intrinsic components.

\begin{equation}
\begin{aligned}
\mathbf{p}_{d,k+1} &= \mathbf{p}_{d,k} + \Delta T\, \mathbf{v}_{d,k}, \quad
\mathbf{v}_{d,k+1} = \mathbf{v}_{d,k} + \Delta T\, \mathbf{a}_{d,k}, \\
\mathbf{R}_{d,k+1} &= \mathbf{R}_{d,k}\exp\!\big(\Delta T\,\boldsymbol{\Omega}_{d,k}^{\wedge}\big), \quad
f_{k+1} = f_{k} + \Delta T\, v_{f,k}.
\end{aligned}
\label{eq:dynamics}
\end{equation}

Translational motion is integrated using a forward Euler scheme.
The camera orientation, decoupled from the drone through a gimbal,
is represented by $\mathbf{R}_{d,k}\in SO(3)$ and updated via the
exponential map, where $(\cdot)^{\wedge}$ denotes the skew-symmetric
operator, ensuring orthonormality.


\begin{figure*}[!t]
\centering
\begin{tabular}{ccc} 
\includegraphics[width=0.30\textwidth]{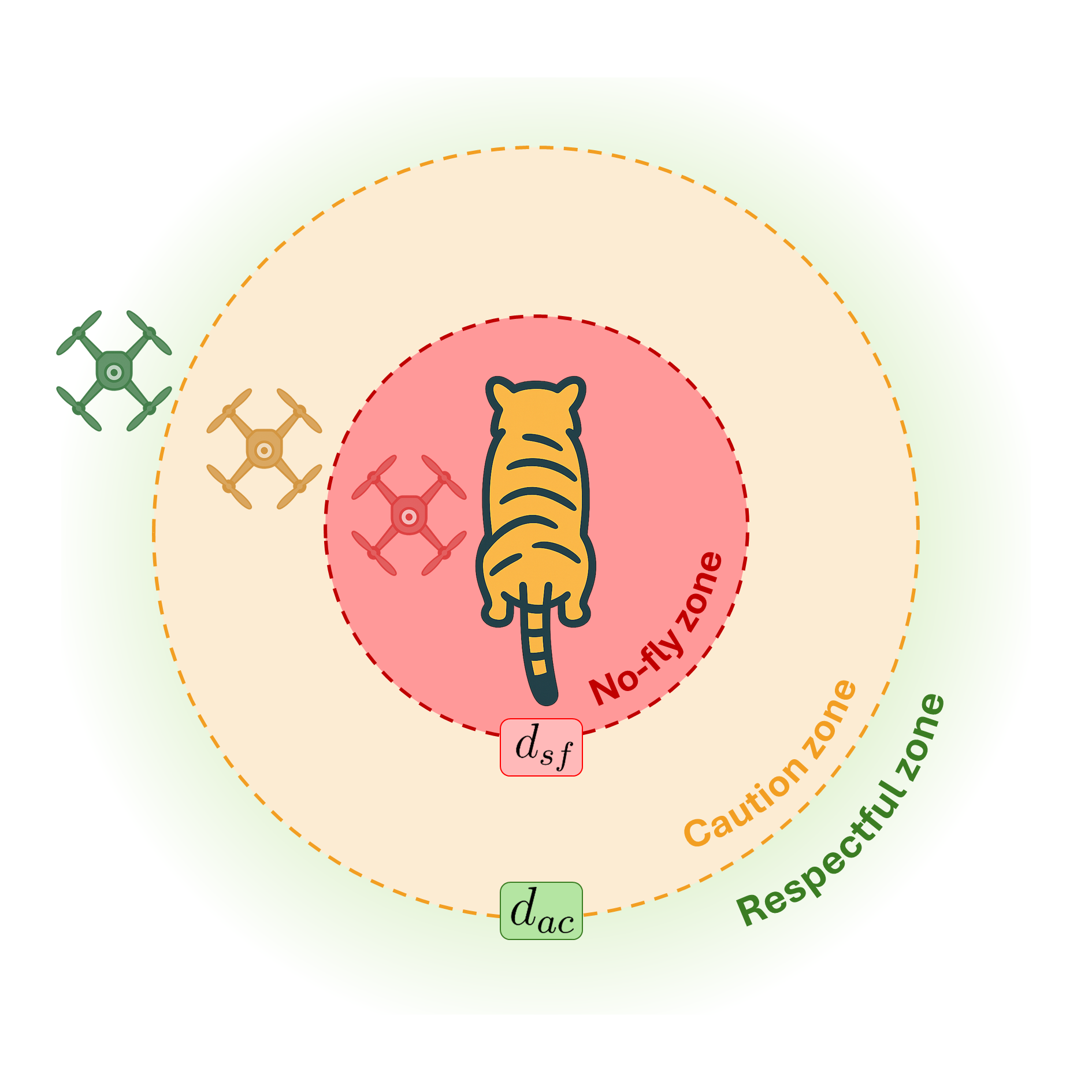} & \includegraphics[width=0.30\textwidth]{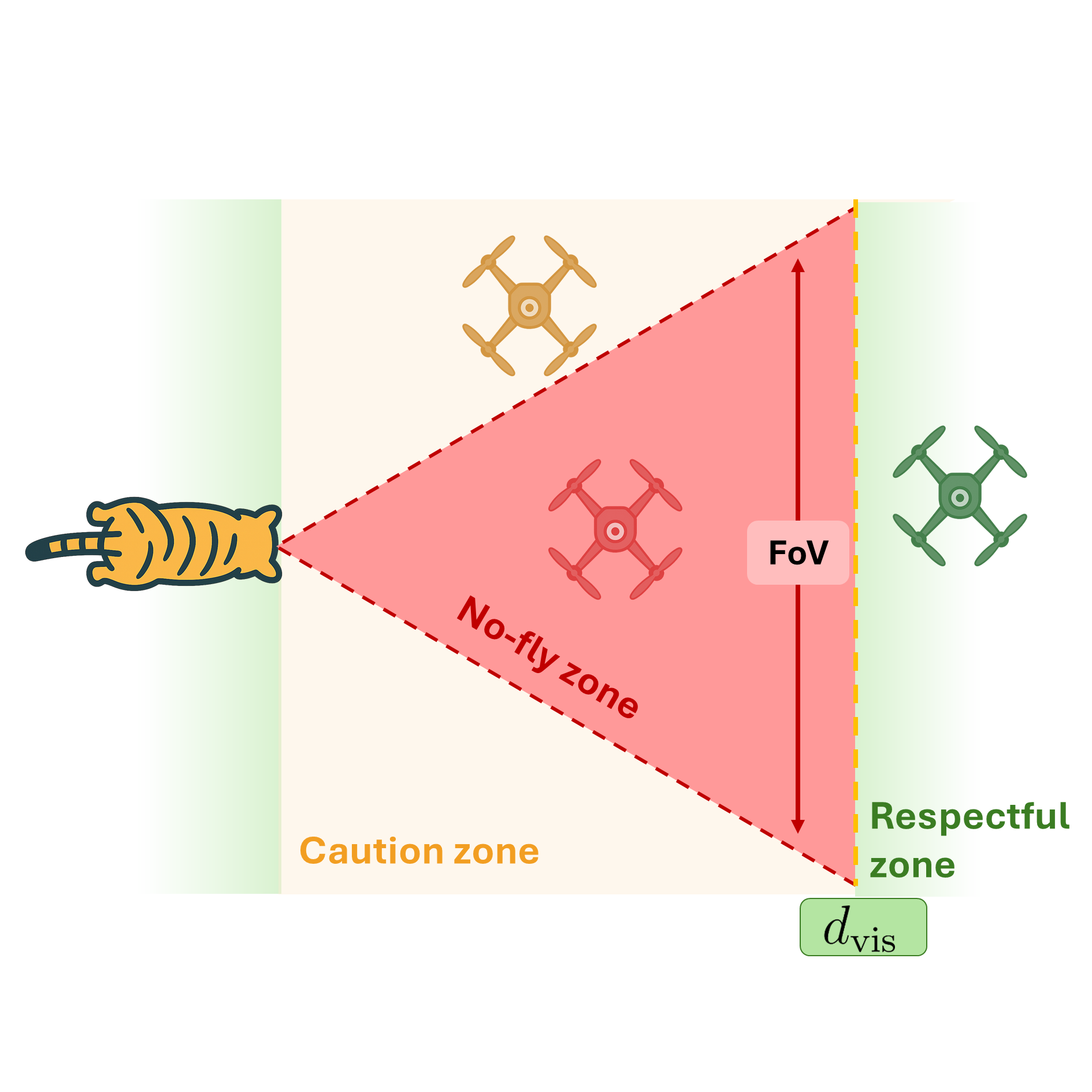} & \includegraphics[width=0.30\textwidth]{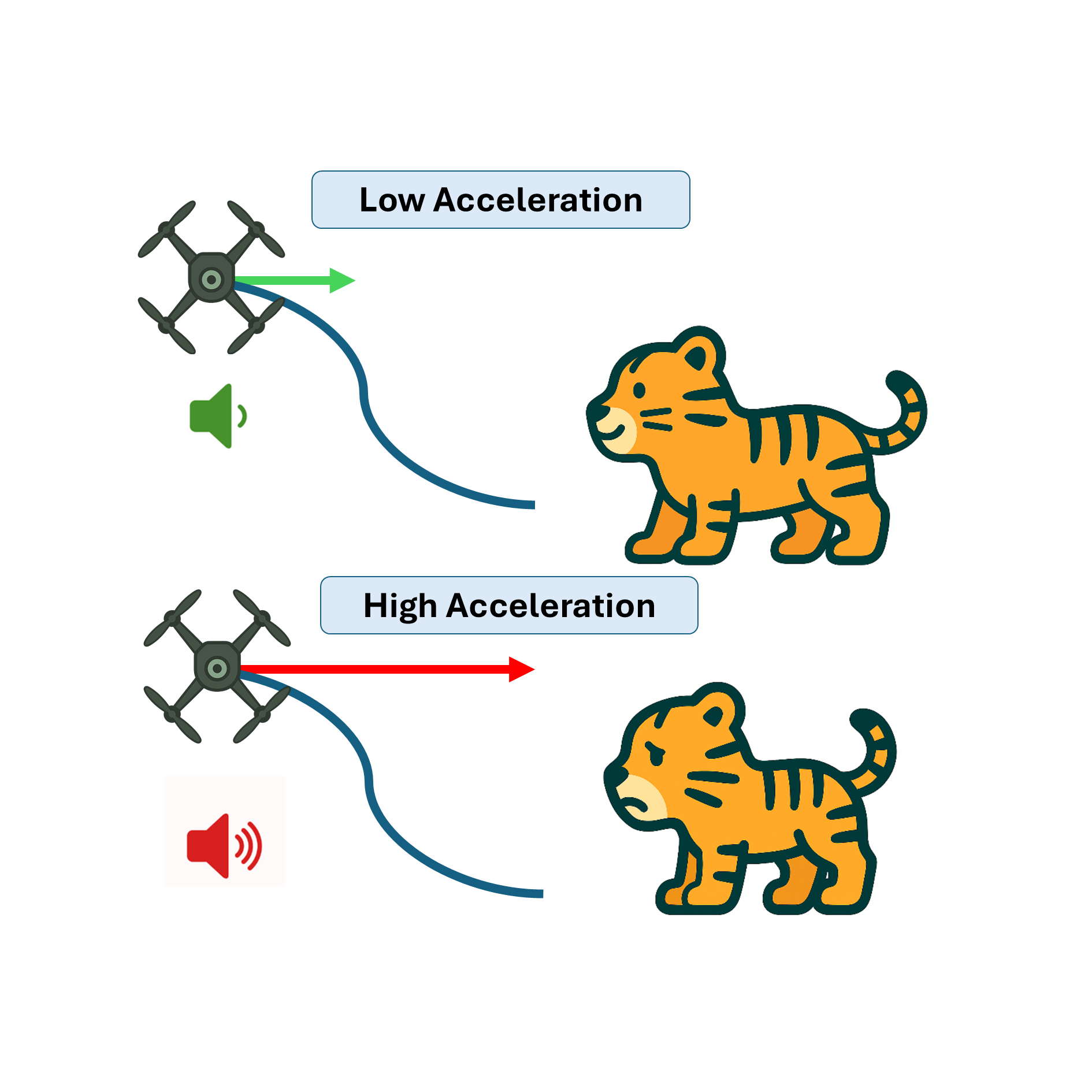} 
\\ \footnotesize (a) Proximity & \footnotesize (b) Visual Range & \footnotesize (c) Smooth Drone Motion \\[0.5em] \multicolumn{3}{c}{ 
\begin{tabular}{cc} 
\includegraphics[width=0.30\textwidth]{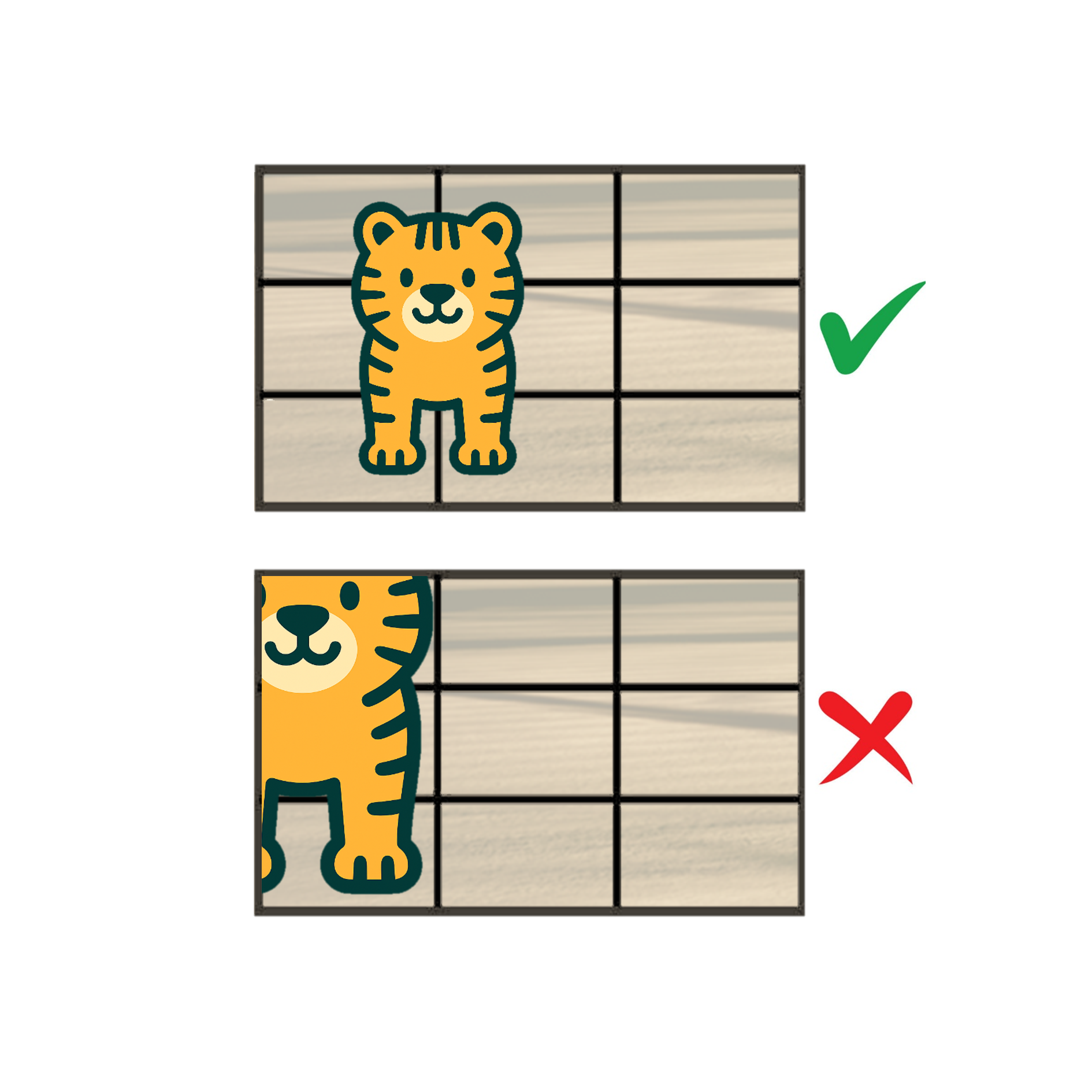} & \includegraphics[width=0.30\textwidth]{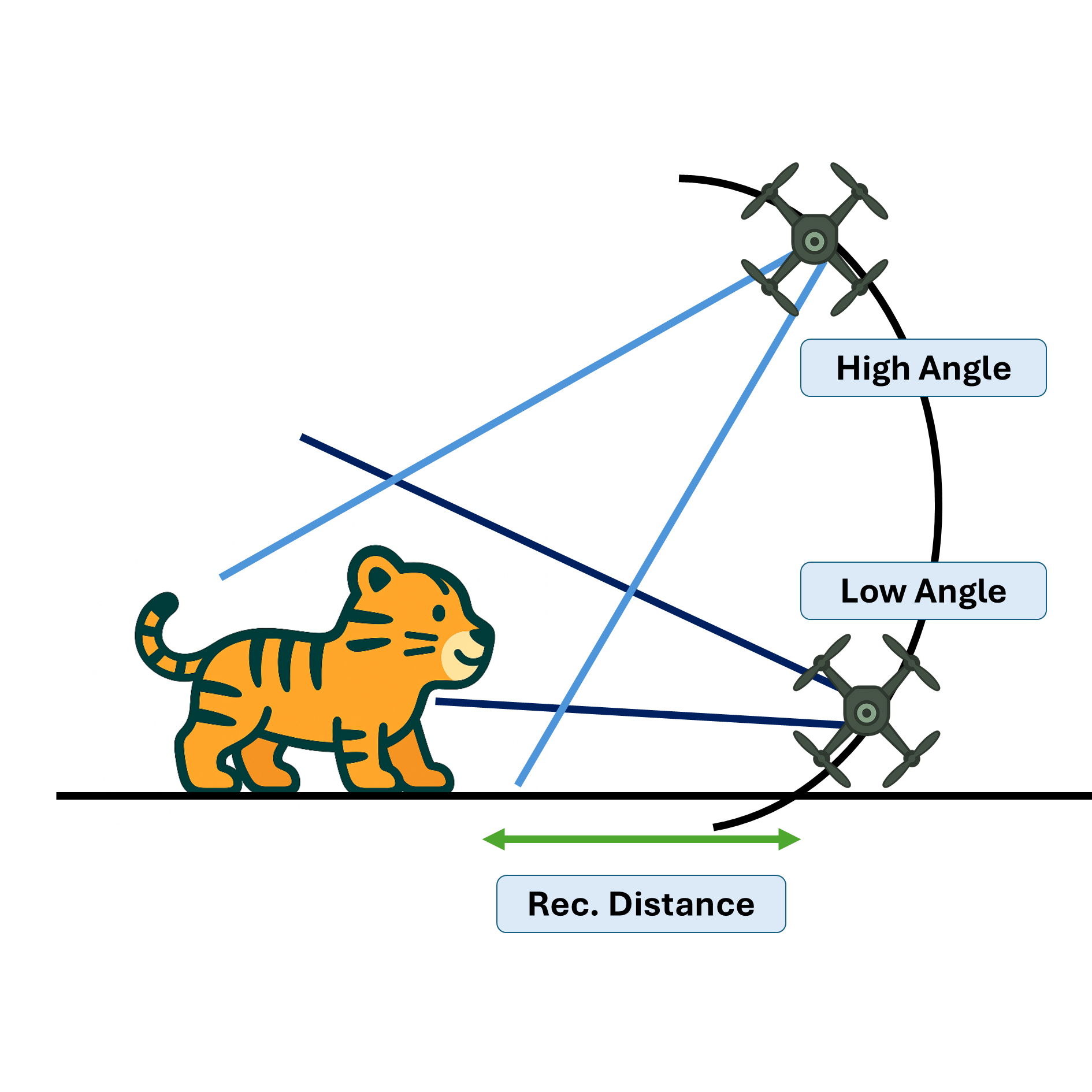} 
\\ \footnotesize (d) Target Image Position & \footnotesize (e) Filming Perspective \end{tabular} } 
\end{tabular}
\caption{\footnotesize{\textbf{CineWild Objectives Overview.}} 
Drone icons are color-coded by ethical impact: green denotes wildlife-safe zones, yellow caution areas, and red potential disturbance. 
\textbf{Ethical objectives:} (a) proximity-aware zoning based on acoustic thresholds \(d_{\text{sf}}\) and \(d_{\text{ac}}\); (b) field-of-view modeling using visual range \(d_{\text{vis}}\); (c) smooth-motion penalization to reduce disturbance. 
\textbf{Cinematographic objectives:} (d) image composition enforcing framing rules (e.g., rule of thirds); (e) perspective control of viewing angle and recording distance.}
\label{fig:costs_objectives}
\end{figure*}
\subsection{Optimization Problem}
We extend the cost function introduced in~\cite{pueyo2024cinempc}
by incorporating additional objectives and hard constraints
specifically tailored for ethical wildlife cinematography.
The resulting optimization problem is formulated as
\begin{equation}
\footnotesize
\begin{aligned}
\min_{\substack{\mathbf{u}_{d,k_0}..\mathbf{u}_{d,k_0+N}\\ \mathbf{u}_{c,k_0}..\mathbf{u}_{c,k_0+N}}} \quad
& \sum_{k=k_0}^{k_{0}+N}
\Big( J_{\text{ethics},k} + J_{\text{visual},k} \Big)\\ 
\textrm{s.t.} \quad 
& (\ref{eq:dynamics})  \\
& \mathbf{x}_{d,k} \in \mathcal{X}_d,\; \mathbf{x}_{c,k} \in \mathcal{X}_c \\
& \mathbf{u}_{d,k} \in \mathcal{U}_d,\; \mathbf{u}_{c,k} \in \mathcal{U}_c \\
& \mathbf{x}_{d,k} \notin \mathcal{X}_{\mathrm{ethics}}
\end{aligned}
\end{equation}

The cost function consists of two components: an ethical term
$J_{\text{ethics},k}$, which governs responsible behavior around
wildlife, and a visual term $J_{\text{visual},k}$, which enforces
cinematographic objectives.

The optimization is subject to system dynamics and feasibility
constraints on both states and controls, represented by $\mathcal{X}_d$, $\mathcal{X}_c$, $\mathcal{U}_d$, and $\mathcal{U}_c$.
These sets capture the platform's physical limits, such as bounded velocity, acceleration, gimbal motion, and focal length ranges. The forbidden region $\mathcal{X}_{\mathrm{ethics}}$ encodes hard
ethical requirements, including maintaining safe distances and avoiding the animal's field of view.

\subsection{Ethical Objectives}

The ethical layer is structured around two elements: hard constraints, which define conditions the drone must never violate, and soft objectives, which guide its behavior within the admissible space, see Fig.~\ref{fig:costs_objectives}. These components enforce safety around animals while encouraging smooth, minimally intrusive motion.

The hard constraints related to the No-Fly zones, $\mathcal{X}_{\mathrm{ethics}}$,
prevent the platform from entering areas where proximity or visibility by
the animal would cause unacceptable disturbance.
\begin{equation}
\mathcal{X}_{\text{ethics}} =
\mathcal{X}_{\text{prox}} \cup \mathcal{X}_{\text{FoV}}.
\end{equation}
These regions are referred to as \emph{No-Fly Zones} throughout the paper and in the associated figures.
\rev{With several animals in the scene, $\mathcal{X}_{\text{ethics}}$ becomes the union of the keep-out sets of all animals and $J_{\text{ethics}}$ the sum of their terms; the implementation already evaluates the terms per animal, but the experiments consider a single animal.}

Within the admissible region, comprising the Acceptable and Caution Areas, the ethical soft objective is defined as
\[
J_{\text{ethics},k} =
J_{\text{prox},k} + J_{\text{FoV},k} + J_{\text{smooth},k},
\]
encouraging the drone to maintain safe and acoustically responsible
distances, remain outside the animal’s field of view, and follow smooth trajectories.

Each ethical component, presented below, is grounded in empirical findings from wildlife–disturbance research, with the corresponding literature integrated directly into its subsection. This ensures that our formulation reflects established best practices in ethical wildlife cinematography, such as maintaining adequate distance, minimizing visual exposure, and favoring optical zoom over physical approach~\cite{mulero2017unmanned,christie2016unmanned}.
\subsection*{Proximity}
Wildlife is highly sensitive to noise and close-range approaches, and field studies show that drones may induce stress or behavioral changes when flown too close to animals~\cite{christie2016unmanned,linchant2015unmanned}. To mitigate these effects, we model proximity using a hard safety constraint and a distance-based cost that together regulate the drone--animal separation (Fig.~\ref{fig:costs_objectives}-a).

Safe operating distances vary significantly across species and depend on both auditory sensitivity and drone characteristics~\cite{vas2015approaching,duporge2021determination,afridi2025impact, afridi2026wildlife}. These findings motivate adjustable distance thresholds in our formulation. Let $d_{dt,k}$ denote the  distance between the drone and the animal, and define two thresholds that partition the space into three zones.

\paragraph{Hard Constraint}
To prevent disturbance due to close-range operation, we define
a No-Fly region corresponding to configurations where the drone
violates a minimum safe distance from the animal as
\begin{equation}
\mathcal{X}_{\mathrm{prox}}
=
\left\{
x_{d,k}
\;\middle|\;
d_{dt,k} < d_{\mathrm{sf}}
\right\}.
\end{equation}
The controller enforces the drone remains outside this region.
\paragraph{Cost Term $J_{\text{prox},k}$}
The acoustic distance $d_{\text{ac}}$ denotes the range within which drone noise may cause disturbance, defining a \textit{Caution Zone} ($d_{\text{sf}} \le d_{dt,k} < d_{\text{ac}}$) and a \textit{Respectful Zone} ($d_{dt,k} \ge d_{\text{ac}}$). The proximity cost is defined as
\begin{equation}
  J_{\text{prox},k} = w_{\text{prox}}
  \begin{cases}
    \rev{w_{\text{sf}}(d_{dt,k} - d_{\text{sf}})^2 + C_{\text{sf}},} & \rev{d_{dt,k} < d_{\text{sf}},}\\[3pt]
    w_{\text{ac}}(d_{dt,k} - d_{\text{ac}})^2 + 1, & d_{\text{sf}} \le d_{dt,k} < d_{\text{ac}},\\[3pt]
    e^{-0.5(d_{dt,k} - d_{\text{ac}})}, & d_{dt,k} \ge d_{\text{ac}}.
  \end{cases}
  \label{eq:J_prox}
\end{equation}
This formulation penalizes operation within the acoustic-disturbance region while encouraging observation from respectful distances, with weights $w_{\text{prox}}$ and $w_{\text{ac}}$ controlling the relative importance and sensitivity of the proximity objective.
\rev{The acoustic distance $d_{\text{ac}}$ is a geometric proxy: we do not model sound pressure, propagation or rotor noise, whose dependence on rotor effort is only captured indirectly by the smoothness term $J_{\text{smooth}}$. The thresholds are species parameters (Table~\ref{table:weights}). Responses to drones are species-specific~\cite{afridi2025impact}: giraffes are among the most reactive species, responding to drones flown lower than 60\,m and closer than 100\,m horizontally~\cite{bennitt2019terrestrial}, and big cats such as tigers may chase or attack drones~\cite{rebolo2019drones}. The values used in our simulated scenes are illustrative and smaller than these field distances, to fit the simulated environments; changing them changes the behavior without retuning any weight (Table~\ref{table:robustness}).}
\rev{The first branch, with $w_{\text{sf}} = 10\,w_{\text{ac}}$ and $C_{\text{sf}} = w_{\text{ac}}(d_{\text{sf}}-d_{\text{ac}})^2+1$, makes $J_{\text{prox}}$ continuous at $d_{\text{sf}}$; it only acts on intermediate solver iterates, since $d_{dt,k} \ge d_{\text{sf}}$ is a hard constraint. $J_{\text{prox}}$ is also continuous at $d_{\text{ac}}$.}


\begin{figure}[!b]
\centering
\includegraphics[width=0.8\columnwidth]{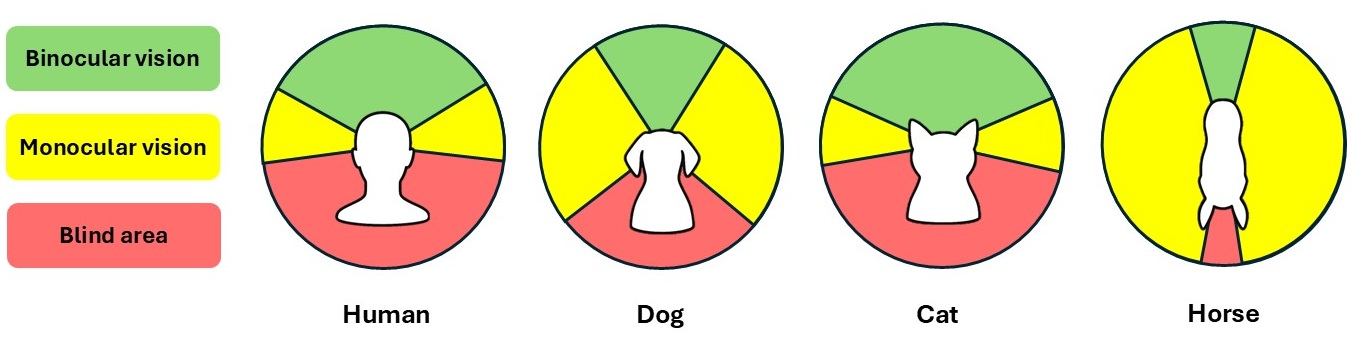} 

\caption{\footnotesize{\textbf{Species-Aware Field-of-View Modeling.} Visualization of field-of-view variations across animal species with different visual systems, accounting for forward-facing, lateral, and stereoscopic eye configurations~\cite{land2012animal}.}}
\label{fig:fov}
\end{figure}
\subsection*{Visual Range}
\label{sec:fov}
Visual disturbance is a major source of animal stress, as many species react strongly when a UAV enters their field of view, particularly at close distances~\cite{christie2016unmanned,linchant2015unmanned}. To mitigate visual exposure, we model visibility using a hard safety constraint and a soft cost that together regulate drone configurations relative to the animal’s gaze (Fig.~\ref{fig:costs_objectives}-b).

Animal visual systems vary widely across species, ranging from forward-facing to lateral or panoramic vision~\cite{land2012animal}. We capture this diversity by modeling the animal’s eye as a camera with adaptable intrinsic parameters, enabling species-aware visibility modeling, as illustrated in Fig.~\ref{fig:fov}.

\paragraph{Visibility Modeling}
The drone’s 3D position is projected onto the animal’s image plane using a pinhole camera approximation. The projected image coordinates $\mathbf{im}_{d,k}\in\mathbb{R}^2$ are given by
\[
\mathbf{im}_{d,k} = \lambda\,\mathbf{K}_{\text{eye}}\,\mathbf{p}_{td,k},
\]
where $\lambda$ normalizes homogeneous coordinates, $\mathbf{K}_{\text{eye}}$ denotes the intrinsic calibration matrix of the animal’s eye, and $\mathbf{p}_{td,k}$ is the relative drone position expressed in the eye frame,
\[
\mathbf{p}_{td,k} = \mathbf{R}_{t,k}^\top\!\left(\mathbf{p}_{d,k}-\mathbf{p}_{t,k}\right).
\]

The intrinsic matrix is defined as
\[
\mathbf{K}_{\text{eye}} =
\begin{bmatrix}
\beta_x f_{\text{eye}} & s & c_u \\
0 & \beta_y f_{\text{eye}} & c_v \\
0 & 0 & 1
\end{bmatrix},
\]
where $f_{\text{eye}}$ is the effective focal length of the animal’s visual system, $(c_u,c_v)$ is the optical center, $s$ the skew, and $\beta_x=\tfrac{W_{\text{px}}}{W_{\text{mm}}}$, $\beta_y=\tfrac{H_{\text{px}}}{H_{\text{mm}}}$ convert units from millimeters to pixels.

The resulting horizontal field of view is
\[
\text{FoV}_x = 2 \arctan\!\left(\frac{W_{\text{mm}}}{2f_{\text{eye}}}\right),
\]
and analogously for the vertical direction, enabling flexible adaptation across species.
\rev{The intrinsic matrix is used here as a compact way of expressing the horizontal and vertical angles of view of the animal (FoV$_x$, FoV$_y$) and their centre; it is equivalent to specifying these angles directly. A single pinhole projection can only represent fields of view below $180^\circ$. Species with lateral or panoramic vision (Fig.~\ref{fig:fov}) would require an angular 
model or several virtual cameras; the experiments in this paper use \limit{a forward-facing predator (in Experiment~1 the visibility terms are never active)}, and we leave panoramic species to future work.}

\paragraph{Hard Constraint}
To prevent visual disturbance at close range, we define a No-Fly
visibility region as
\begin{equation}
\mathcal{X}_{\mathrm{FoV}}
=
\left\{
x_{d,k}
\;\middle|\;
\begin{aligned}
& \rev{d_{dt,k} < d_{\mathrm{vis}} \;\wedge\; p^{(z)}_{td,k} > 0 \;\wedge} \\
& \mathbf{im}_{d,k} \in [0,W_{px}] \times [0,H_{px}]
\end{aligned}
\right\}.
\end{equation}
This region includes configurations where the drone is closer than
$d_{\mathrm{vis}}$, located in front of the animal, and its projection
lies inside the animal’s field of view. The controller avoids these
configurations by ensuring that the system state does not enter
$\mathcal{X}_{\mathrm{FoV}}$ during optimization.

\begin{revblock}
\paragraph{Smooth reformulation}
The pinhole description of $\mathcal{X}_{\mathrm{FoV}}$ divides by the depth $p^{(z)}_{td,k}$, which fails when the drone is directly next to the animal ($p^{(z)}_{td,k}=0$) and wrongly projects points behind it into the image. We therefore give the optimizer an equivalent condition without the division. In front of the animal, keeping the drone beyond the lateral image border, $im^{(x)}_{d,k} \le \epsilon$ (with a small margin $\epsilon$ px outside the image), is the same as keeping the angle between the animal's gaze and the direction to the drone larger than half the field of view, a linear condition on their relative position:

\begin{equation}
p^{(x)}_{td,k} - \kappa\, p^{(z)}_{td,k} \le 0, \qquad
\kappa = \frac{\epsilon\, W_{\text{mm}}/W_{\text{px}} - W_{\text{mm}}/2}{f_{\text{eye}}}.
\end{equation}
A smooth gate $\sigma_k = \tfrac{1}{2}\big(1+\tanh(p^{(z)}_{td,k}/\delta)\big)$, close to one in front of the animal and to zero behind it, deactivates this condition where the animal cannot see:
\begin{equation}
g_{\text{FoV},k} = \sigma_k \big(p^{(x)}_{td,k} - \kappa\, p^{(z)}_{td,k}\big) - (1-\sigma_k)\, m \;\le\; 0,
\label{eq:fov_smooth}
\end{equation}
with $\delta = 0.5$\,m and margin $m = 1$\,m. The constraint is differentiable everywhere, leaves rear viewpoints free, and is imposed on the whole horizon while the drone is closer than $d_{\text{vis}}$. \limit{The drone is kept on a fixed side of the view cone (the left one in our experiments); choosing the side online is left to future work.}
\end{revblock}


\paragraph{Cost Term $J_{\text{FoV},k}$}
Outside the No-Fly Zone, a soft cost reduces the likelihood of visual exposure. When the drone is close and in front of the gaze but remains outside the field of view, a penalty encourages motion away from the optical center:
\begin{equation}
J_{\text{FoV},k} =
w_{\text{FoV}}
\begin{cases}
\rev{\sigma_k}\, V_{\text{out}}, & d_{dt,k} < d_{\text{vis}} \ \text{and}\ \rev{im^{(x)}_{d,k} \le \epsilon},\\
0, & \text{otherwise}.
\end{cases}
\end{equation}
\rev{The in-front condition $p_{td,k}^{(z)} > 0$ is applied through the same smooth gate $\sigma_k$ as in Eq.~\eqref{eq:fov_smooth}, so the soft visibility cost vanishes smoothly behind the animal.}

The penalty is defined from the normalized image-plane distance
\[
\begin{aligned}
d_{\text{center}} &= \|\mathbf{im}_{d,k}-\mathbf{im}_{\text{center}}\|, \qquad
\hat{d} = \frac{d_{\text{center}}}{d_{\text{max}}},
\end{aligned}
\]
\[
\centering
d_{\text{max}} = \sqrt{c_u^2 + c_v^2}, \qquad
V_{\text{out}} = e^{-\hat{d}^2}
\]
where $\mathbf{im}_{\text{center}}=[c_u,c_v]^\top$. The weight $w_{\text{FoV}}$ controls the relative importance of visual avoidance during optimization.

\rev{\paragraph{Hard and soft visibility regions}
The two terms act in different regions in front of the animal and within $d_{\text{vis}}$: the hard constraint excludes projections inside the animal's field of view, while the soft term acts just outside it and encourages the drone to move farther from the image centre. Neither term is active behind the animal or beyond $d_{\text{vis}}$. The visibility cost $J_{\text{FoV}}$ is therefore discontinuous at $d_{\text{vis}}$, since moving just beyond this threshold removes the cost $V_{\text{out}}$. We retain this formulation because $d_{\text{vis}}$ represents a species-dependent threshold, but acknowledge that the optimizer may exploit this boundary. A smoother alternative would be to use the same distance-based gate as $\sigma_k$ to gradually activate the visibility term.}


\subsection*{Smooth Drone Motion}
Beyond proximity and visibility, abrupt UAV motions can also disturb wildlife. Telemetry and experimental studies indicate that sudden accelerations and rapid acoustic fluctuations can increase stress responses in many species~\cite{ditmer2015bears,mulero2017unmanned}. To mitigate these effects, we include a smooth‑motion cost that discourages aggressive maneuvers (Fig.~\ref{fig:costs_objectives}-c).

\paragraph{Cost Term $J_{\text{smooth},k}$}
Motion smoothness is encouraged by penalizing high accelerations, reducing noise peaks and mechanical vibrations:
\begin{equation}
  J_{\text{smooth},k} = w_{\text{smooth}}\left\| \mathbf{a}_{d,k} \right\|^2 .
\end{equation}
The weight $w_{\text{smooth}}$ controls the relative importance of smooth flight behavior within the overall optimization.

\subsection{Visual Objectives}

The cinematographic objectives are encoded through a set of soft cost terms that shape the visual quality of the footage. 
Following previous work~\cite{pueyo2024cinempc}, the visual component of the cost is decomposed into two elements:
\[
J_{\text{visual},k} = J_{\text{im},k} + J_{\text{p},k}.
\]
Together, these terms ensure that the resulting footage remains visually coherent, aesthetically pleasing, and compositionally consistent throughout the recording.
\label{sec_visual_obj}

\subsection*{Target Image Position}
The target image position objective $J_{\text{im},k}$ penalizes deviations of the target from a desired location in image space, ensuring consistent cinematographic framing throughout the recording process (Fig.~\ref{fig:costs_objectives}-d). By guiding the target toward a reference image position, this term allows the controller to adjust both camera intrinsic and extrinsic parameters for optimal composition.

Let $\mathbf{im}_{t,k}$ and $\mathbf{im}^*_{t,k}\in\mathbb{R}^2$ denote the actual and desired image coordinates of target $t$ at time step $k$, respectively. The image composition cost is defined as
\begin{equation}
J_{\text{im},k} = w_{\text{im}} \left\| \mathbf{im}_{t,k} - \mathbf{im}^*_{t,k} \right\|^2 ,
\end{equation}
where $w_{\text{im}}$ weights the importance of framing accuracy. The projection of the target into image coordinates follows the same procedure as in the visibility model.

\subsection*{Filming Perspective}
The filming perspective objective $J_{\text{p},k}$ regulates the camera--target viewpoint by enforcing preferred depth and orientation, ensuring a stable and contextually appropriate cinematographic perspective (Fig.~\ref{fig:costs_objectives}-e).

The perspective is characterized by the target depth $d_{dt,k}$ and the relative rotation $\mathbf{R}_{dt,k}$ between the camera and the target (see Sec.~\ref{sec:solution_state}). Desired values $d_{dt,k}^*$ and $\mathbf{R}_{dt,k}^*$ are incorporated via the cost
\begin{equation}
J_{\text{p},k} =
w_d \left( d_{dt,k} - d_{dt,k}^* \right)^2
+ w_R \left\| \mathbf{R}_{dt,k}^\top - \mathbf{R}_{dt,k}^* \right\|_F ,
\end{equation}
where $w_d$ and $w_R$ weight the importance of depth and orientation, respectively.
\rev{$\|\cdot\|_F$ is the Frobenius norm, $\|\mathbf{A}\|_F = \sqrt{\operatorname{tr}(\mathbf{A}^\top\mathbf{A})}$; on $SO(3)$ it is a chordal rather than a geodesic distance, which is smooth and sufficient to penalise viewpoint deviations. The distance term is kept even though zoom can compensate the size of the target in the image: changing distance and changing focal length are not equivalent in cinematography, since the distance determines the perspective and the relative scale of the background, while the focal length only scales the image. $d^*_{dt}$ is therefore part of the shot specification; when the distance is free, as in Experiment~1, $w_d = 0$.}

\begin{table}[!b]
\centering
\caption{Weights of Cost Terms for Simulation Experiments \rev{and species parameters}}
\begin{tabular}{|c|c|c|c|c|c|c|}
\hline
\textbf{Exp} & \(w_{\text{prox}}\) & \(w_{\text{FoV}}\) & \(w_{\text{smooth}}\) & \(w_{\text{im}}\) & \(w_{d}\) & \(w_{R}\) \\
\hline
E1 & 15 & \rev{1} & \rev{2} & 1 & 0 & 250 \\
E2 & \rev{15} & 1 & \rev{2} & 0.5 & 10 & 100 \\
\hline
\hline
 & \multicolumn{2}{c|}{\rev{$d_{\text{sf}}$}} & \multicolumn{2}{c|}{\rev{$d_{\text{ac}}$}} & \multicolumn{2}{c|}{\rev{$d_{\text{vis}}$}} \\
\hline
\rev{E1 (giraffe)} & \multicolumn{2}{c|}{\rev{5\,m}} & \multicolumn{2}{c|}{\rev{20\,m}} & \multicolumn{2}{c|}{\rev{12\,m}} \\
\rev{E2 (tiger)} & \multicolumn{2}{c|}{\rev{5\,m}} & \multicolumn{2}{c|}{\rev{8\,m}} & \multicolumn{2}{c|}{\rev{12\,m}} \\
\hline
\end{tabular}
\label{table:weights}
\end{table}

\section{Experiments}
This section presents the experimental validation of CineWild. We conduct two simulation experiments and two real-world proof-of-concept trials to evaluate whether the proposed formulation transfers to physical sensing and camera actuation. In simulation, CineWild is compared against two baselines under identical conditions: \rev{(i) CineMPC~\cite{pueyo2024cinempc}, the cinematographic controller on which CineWild builds, which jointly controls the drone and the focal length but has no welfare terms; and (ii) a welfare-aware baseline, which we refer to as CineMPC + bubble, implementing the most direct alternative to CineWild: a safety bubble around the animal. It has the same cinematographic costs as CineMPC, and its minimum drone--animal distance, imposed as a hard constraint at every step of the prediction horizon, is raised to the acoustic distance, $d_{dt} \ge d_{\text{ac}}$; it has no graded proximity cost, no visibility terms and no smoothness term. It therefore keeps the acoustic distance by construction, and the comparison with it isolates what the welfare terms of CineWild add beyond a safety distance combined with zoom.} Quantitative and qualitative results highlight the differences between the approaches in terms of ethical constraint satisfaction and cinematographic quality, revealing the limitations of baseline cinematographic control under dynamic wildlife-interaction scenarios.
For a more visual and qualitative representation of the results, we refer the reader to the supplementary video.

\subsection*{\textbf{Experiments in simulation}}

\subsection{Simulation Experimental Setup}
\label{sec:sim_setup}
{\textit{Platform.}} The experiments in simulation were run on a laptop running Ubuntu 20.04, equipped with an NVIDIA GeForce RTX 3070 graphics card. For the simulations, we employed \textit{CinemAirSim}~\cite{pueyo2020cinemairsim}, an extended version of the Unreal Engine-based robotics simulator \textit{AirSim}
, which enables configurable camera intrinsic parameters. Upon acceptance, the code for the platform will be released.

{\textit{Controller.}} The controller operates with a sampling time of $\Delta T = 0.4\,\text{s}$ and a prediction horizon of $N = 5$. Moreover, \rev{$N = 10$ is evaluated in Sec.~\ref{sec:ablation} (Table~\ref{table:horizon}).} At each timestep, the nonlinear program is solved using IPOPT\rev{, initialised with the current state and zero control inputs}. \rev{In Experiment~1, the median solve time per MPC step is 248\,ms.} \rev{Solver success means convergence to a local optimum that satisfies the constraints; since the problem is non-convex, it does not guarantee a globally optimal trajectory.}

{\textit{Weights and species.}} The weights used for each cost term in the simulation experiments are reported in Table~\ref{table:weights}.
\rev{The ethical weights $(w_{\text{prox}}, w_{\text{FoV}}, w_{\text{smooth}})$ are identical in all experiments; no ethical term is switched off for a particular scenario. Only the species parameters $(d_{\text{sf}}, d_{\text{ac}}, d_{\text{vis}})$ depend on the animal, and the cinematographic weights $(w_{\text{im}}, w_d, w_R)$ encode the requested shot\limit{; they were set per shot and not tuned systematically}. The piecewise terms $J_{\text{prox}}$ and $J_{\text{FoV}}$ are implemented with conditional expressions that are re-evaluated at every solver iteration, so the active branch always corresponds to the current iteration.}

\rev{\textit{Protocol.} Each configuration is run 10 times. In Experiment~1, the drone starts 13--15\,m behind the giraffe, inside the acoustic distance, with a random offset of up to $\pm1$\,m; the seed is the run number, so all configurations start from the same ten poses. In Experiment~2, the drone always starts from the same pose inside the animal's field of view. Runs last 150\,s (Experiment~1) and 90\,s (Experiment~2), set by the duration of the sequences. A run is discarded and repeated only if the simulator fails (the drone stops responding or the animal model jumps or falls through the ground).} \limit{Both scenes contain a single, slow or static animal; agile targets, groups of animals and cluttered environments are not evaluated and left for future work.}

\begin{revblock}
\textit{Metrics.} We report welfare metrics (minimum/mean drone--animal distance, time within $d_{\text{sf}}$ and $d_{\text{ac}}$, and time in the animal's field of view), cinematographic metrics (image-plane errors $e_u$ and $e_v$, viewpoint error $e_{\text{yaw}}$, viewpoint success, and focal length), and motion metrics (root-mean-square (RMS) acceleration $a_{\text{RMS}} = \sqrt{\tfrac{1}{K}\sum_{k=1}^{K} \|\mathbf{a}_{d,k}\|^2}$ over the $K$ logged samples of a run, related to rotor effort and noise). Tables report the mean over the runs of each configuration.
\end{revblock}

\subsection{Experiment 1: Proximity-Aware Control for Acoustic Disturbance Minimization}
This experiment demonstrates how CineWild protects animal welfare during filming while adhering to cinematographic rules. As a representative case, we recorded a giraffe walking across the savanna. This setup enables the evaluation of how the cost terms \( J_{\text{prox}} \) and \( J_{\text{smooth}} \) contribute to minimizing disturbance while maintaining high-quality footage.
\rev{The visibility terms are also included. Since the requested viewpoints are behind and beside the animal, the drone is never inside its field of view, and removing them has no clear effect in this experiment (Sec.~\ref{sec:ablation}).}

For this experiment, we define the acoustic and safe perimeters as $ d_{\text{ac}} = 20\,\text{m}$ and $d_{\text{sf}} = 5\,\text{m}
$. \rev{The drone starts 13--15\,m behind the giraffe, inside its acoustic distance, with a random offset of up to $\pm 1$\,m (Sec.~\ref{sec:sim_setup}).}

\subsubsection{Experiment objectives}
The experiment consists of three sequences, each with different control objectives.

\begin{figure*}[!b]
\centering
\setlength{\tabcolsep}{1.5pt}
\renewcommand{\arraystretch}{0.9}
\begin{tabular}{@{}cccc@{}}
    \includegraphics[width=0.24\textwidth]{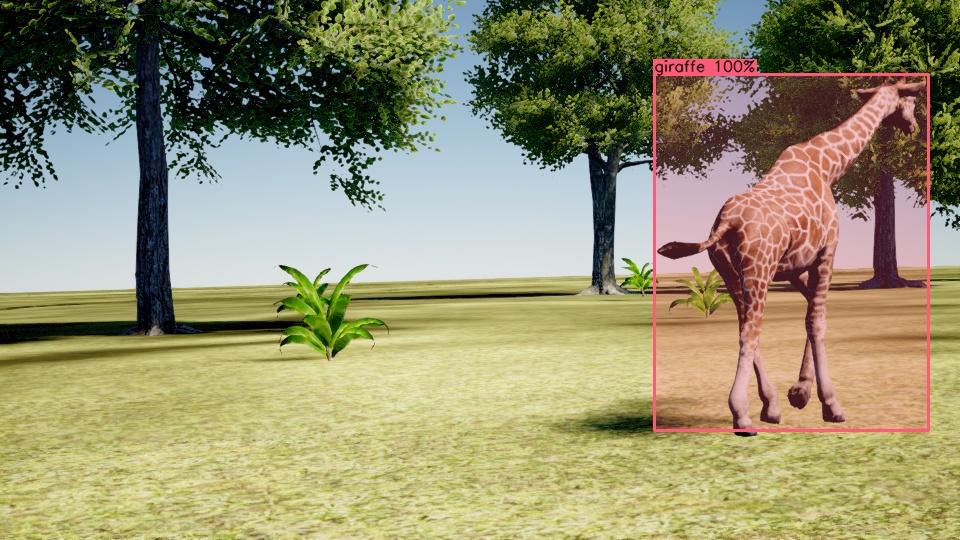} &
    \includegraphics[width=0.24\textwidth]{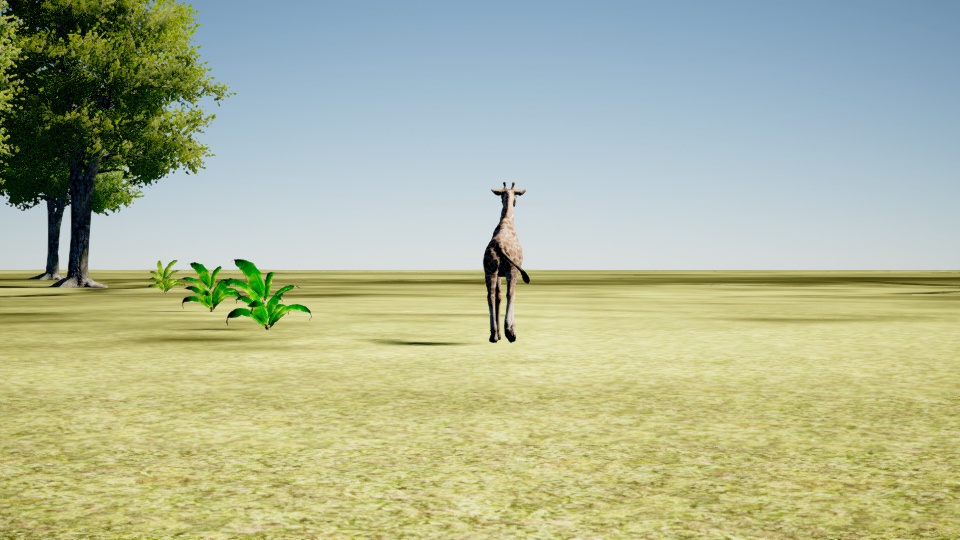} &
    \includegraphics[width=0.24\textwidth]{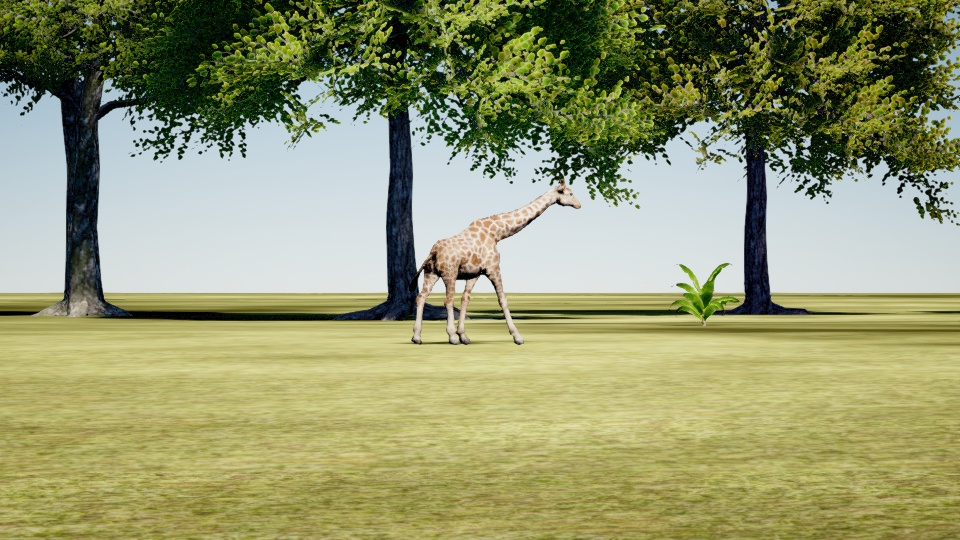} &
    \includegraphics[width=0.24\textwidth]{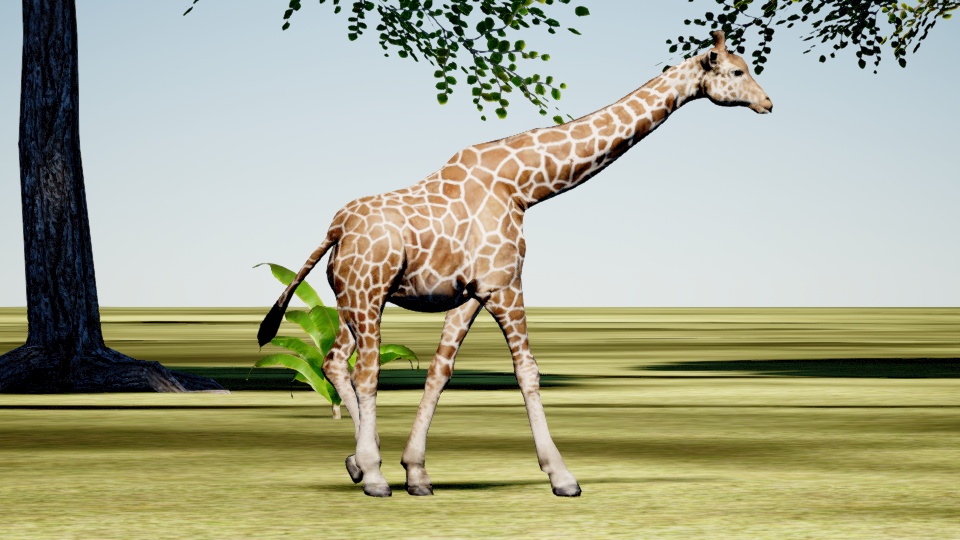} \\
    \includegraphics[width=0.24\textwidth]{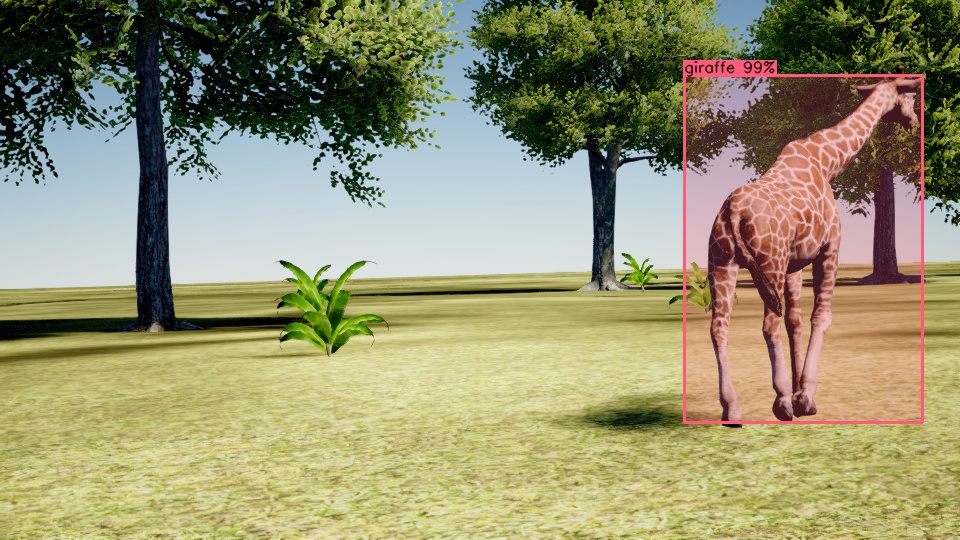} &
    \includegraphics[width=0.24\textwidth]{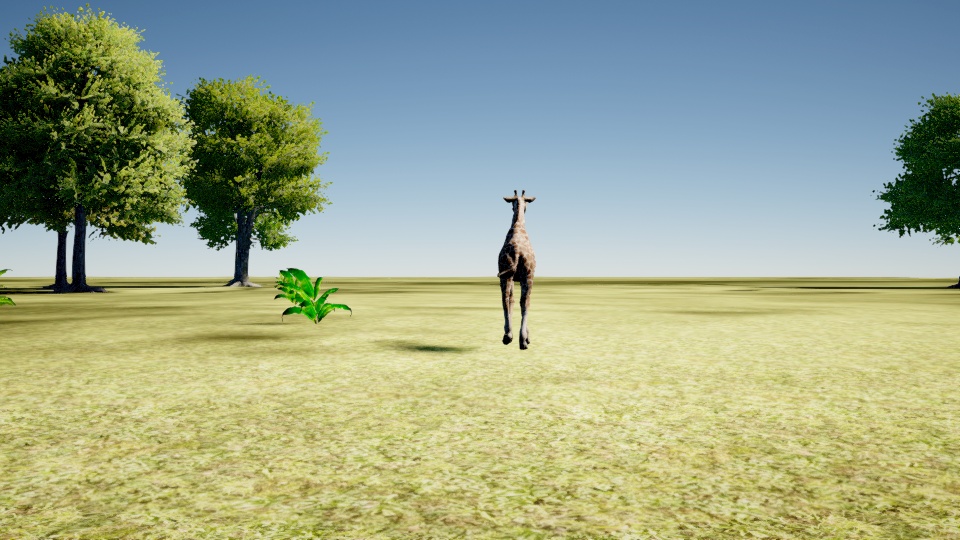} &
    \includegraphics[width=0.24\textwidth]{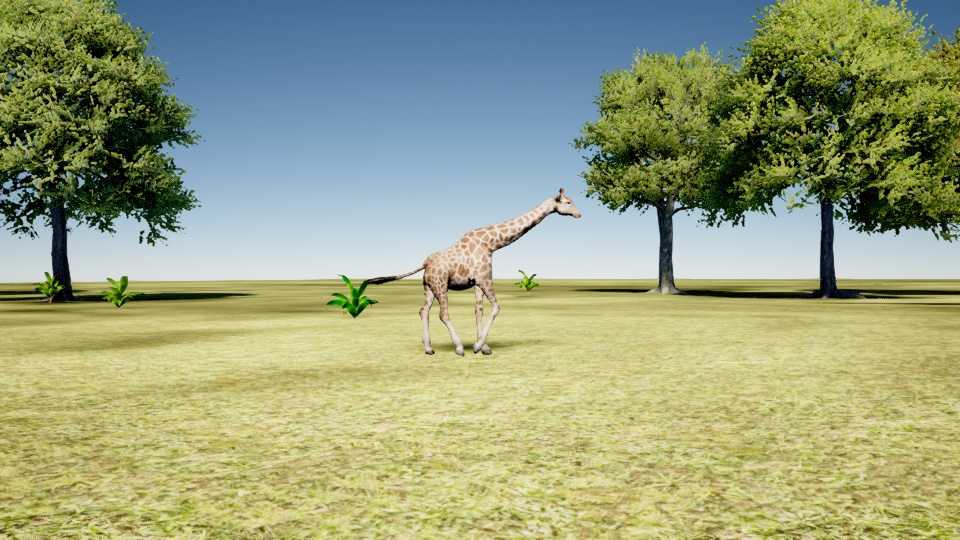} &
    \includegraphics[width=0.24\textwidth]{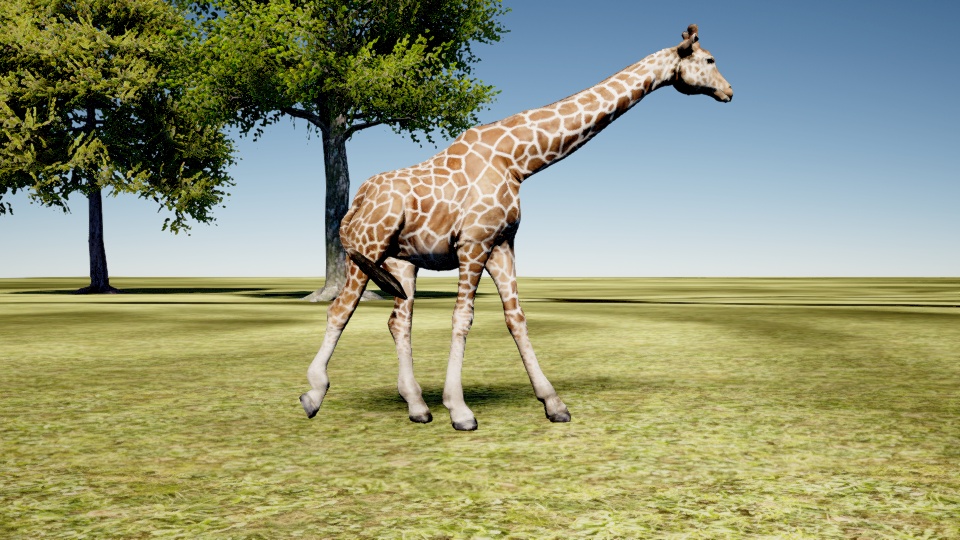} \\
    \multicolumn{4}{@{}p{\dimexpr\textwidth-3pt}@{}}{\centering\footnotesize {\textbf{CineWild} (top) and \textbf{CineMPC} (bottom): onboard camera.}} \\[4pt]
    \includegraphics[width=0.24\textwidth]{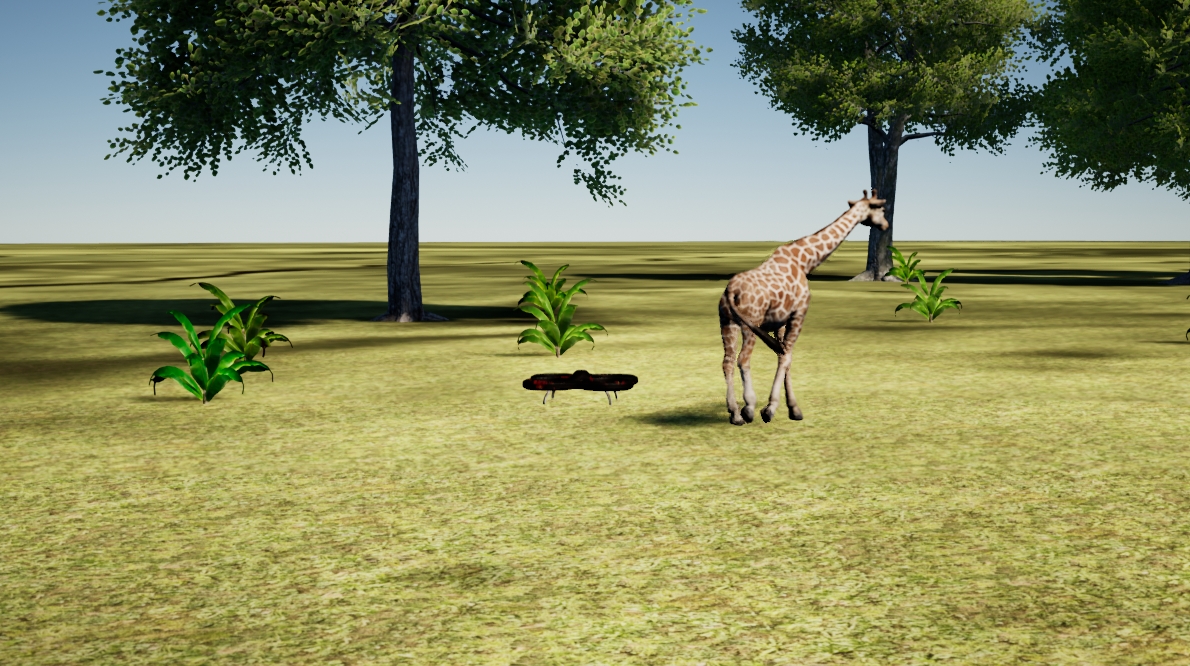} &
    \includegraphics[width=0.24\textwidth]{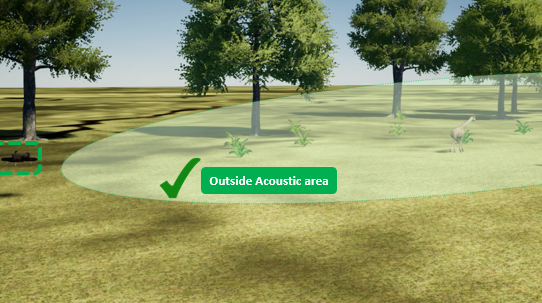} &
    \includegraphics[width=0.24\textwidth]{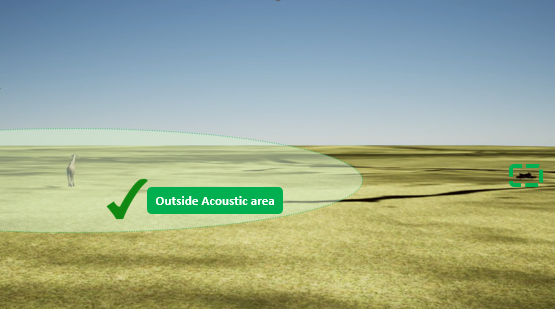} &
    \includegraphics[width=0.24\textwidth]{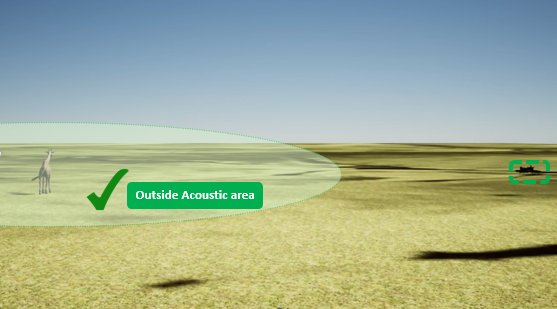} \\
    \includegraphics[width=0.24\textwidth]{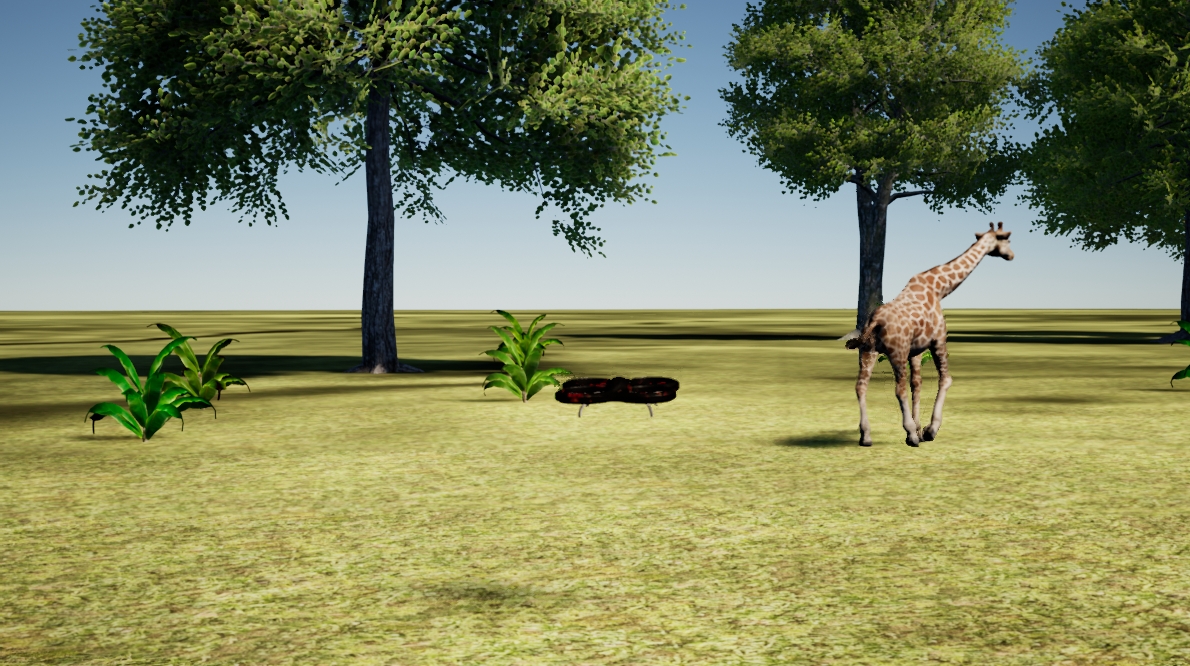} &
    \includegraphics[width=0.24\textwidth]{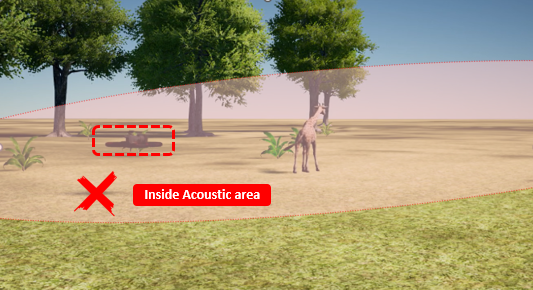} &
    \includegraphics[width=0.24\textwidth]{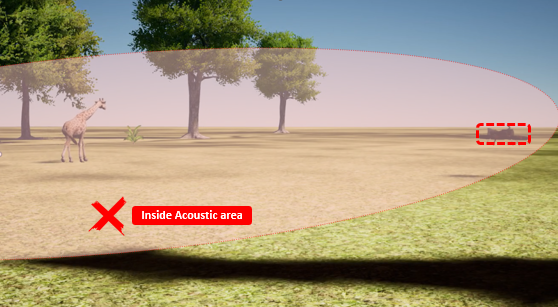} &
    \includegraphics[width=0.24\textwidth]{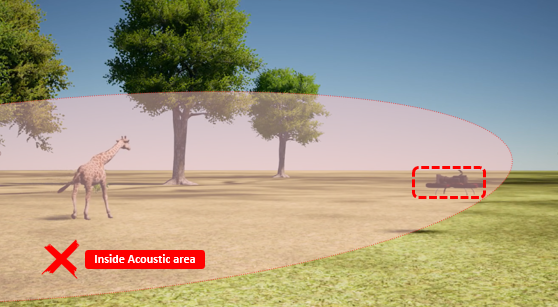} \\
    \multicolumn{4}{@{}p{\dimexpr\textwidth-3pt}@{}}{\centering\footnotesize {\textbf{CineWild} (top) and \textbf{CineMPC} (bottom): third-person view.}} \\
\end{tabular}
\caption{\footnotesize{\rev{\textbf{Experiment~1. Qualitative comparison of CineWild and CineMPC.} Frames are ordered left to right in time. The framing of the giraffe is similar with both methods, given the same composition instructions; notably, the third-person views show that CineWild is able to obtain it from outside the acoustic distance, zooming in instead of approaching the animal.}}}
\label{fig:exp1_combined}
\end{figure*}

\paragraph{Sequence 1: Initial Framing and Acoustic Compliance} 
In the first sequence, the drone operates within the acoustic perimeter, highlighting the effect of proximity constraints. The perspective objective $J_p$ enforces a rear-view configuration, while the image composition term $J_{\text{im}}$ applies rule-of-thirds framing. The proximity objective $J_{\text{prox}}$ drives the drone outward until the acoustic boundary is reached, restoring compliant operation while preserving framing. \rev{The initial and final states of the sequence are shown in the first and second columns of Fig.~\ref{fig:exp1_combined}.}

\paragraph{Sequence 2: Side-Angle Recording}
After reaching an acoustically safe distance, the desired viewpoint is shifted to a lateral configuration. The drone repositions smoothly while maintaining compliance with proximity constraints and preserving cinematographic quality. \rev{The end of the sequence is shown in the third column of Fig.~\ref{fig:exp1_combined}.}

\paragraph{Sequence 3: Close-Up Shot}
In the final sequence, a close-up view is achieved without reducing the drone--target distance. The image composition parameters are adjusted to increase target coverage, and the controller compensates by increasing the focal length. This enables detailed capture while leaving the proximity objective $J_{\text{prox}}$ unaffected. \rev{The end of the sequence is shown in the fourth column of Fig.~\ref{fig:exp1_combined}.}

\begin{figure*}[!t]
\centering
\includegraphics[width=0.90\textwidth]{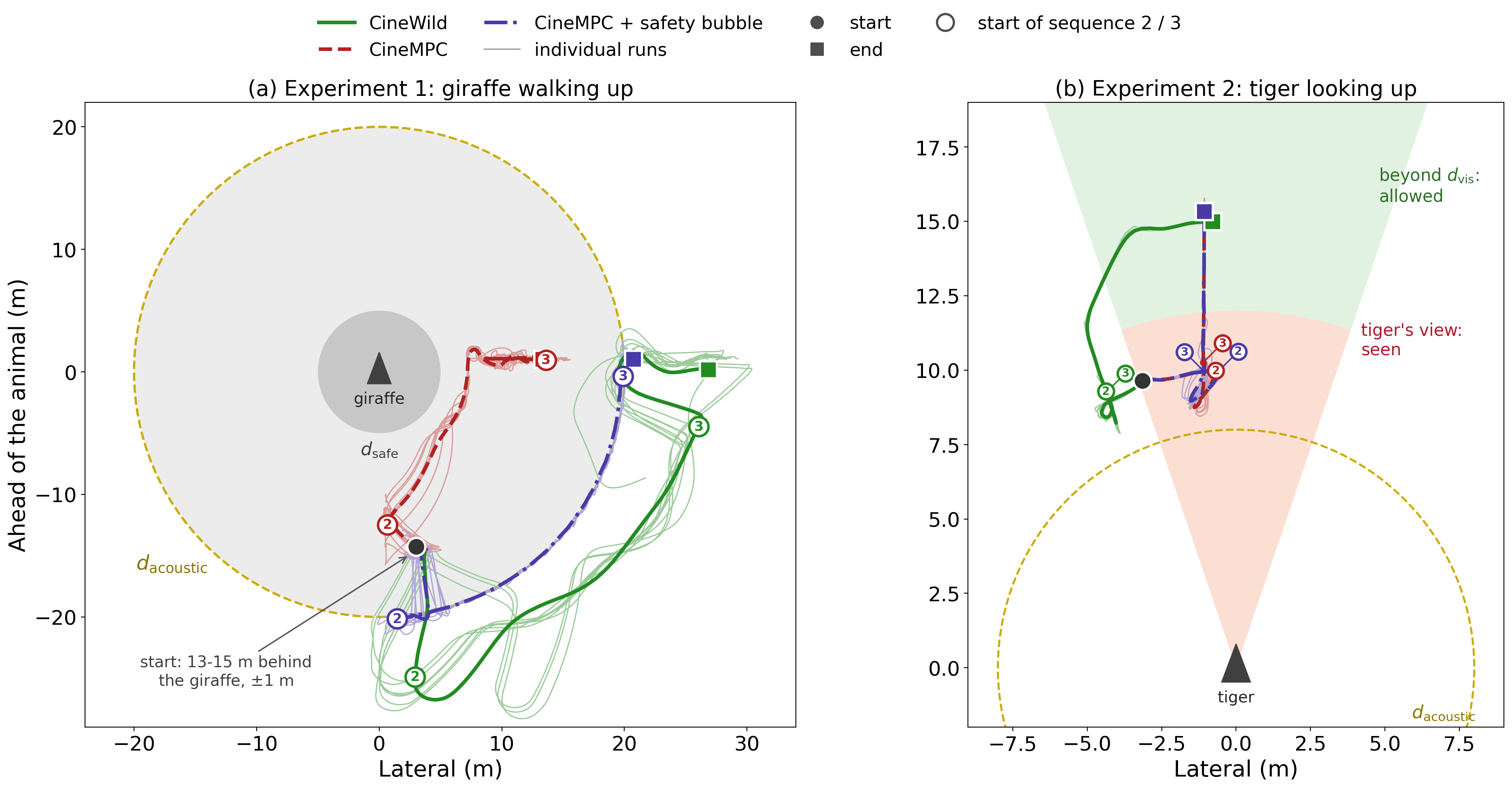}
\caption{\footnotesize{\rev{\textbf{Drone paths around the animal in E1 and E2.} Light lines: each of the 10 runs per method; thick lines: mean path). (a) Experiment~1: light grey, disturbance zone (closer than $d_{\text{ac}}$); dark grey, no-fly zone (closer than $d_{\text{sf}}$); numbered circles: where Sequences~2 and~3 begin. CineMPC approaches the giraffe and stays inside the acoustic distance; the bubble retreats to its edge and slides along it; CineWild retreats behind the animal and circles to the requested side view from outside $d_{\text{ac}}$. (b) Experiment~2: shaded, the tiger's field of view up to $d_{\text{vis}}$; dashed, $d_{\text{ac}}$. All methods start inside the field of view; CineWild leaves it and films from its side, while both baselines stay inside it until they move beyond $d_{\text{vis}}$ for Sequence~3.}}}
\label{fig:topdown}
\end{figure*}

\begin{figure*}[!t]
\centering
\includegraphics[width=\textwidth]{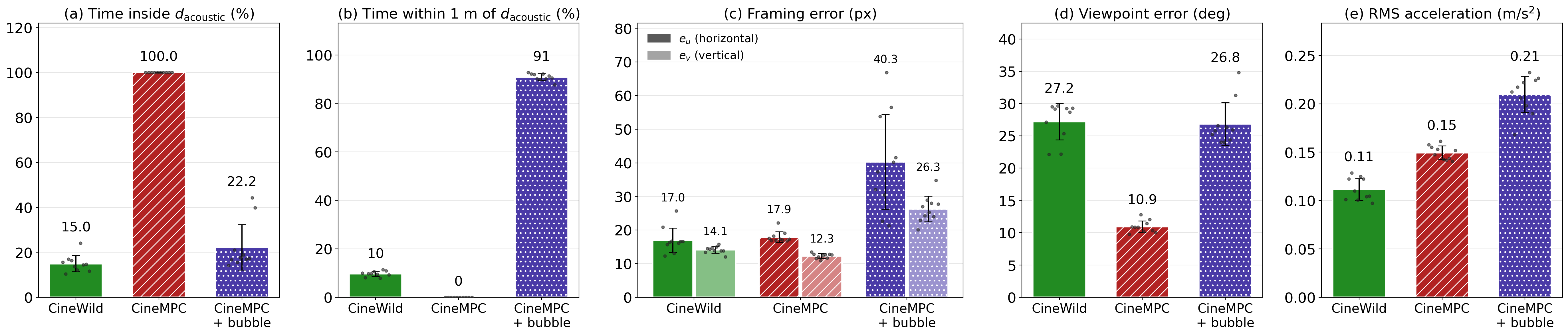}
\caption{\footnotesize{\rev{\textbf{Experiment~1. Comparison of the three methods} (mean $\pm$ std over 10 runs per method; dots: individual runs; lower is better in every panel). The bubble keeps the distance by staying on the edge of $d_{\text{ac}}$, with about twice the framing error and acceleration of CineWild; both pay a larger viewpoint error than CineMPC to stay far from the animal. (a) time inside $d_{\text{ac}}$; (b) time within 1\,m of $d_{\text{ac}}$; (c) framing error, horizontal $e_u$ and, lighter, vertical $e_v$; (d) viewpoint error; (e) RMS acceleration.}}}
\label{fig:exp1_bars}
\end{figure*}

\begin{figure*}[!t]
\centering
\includegraphics[width=0.9\textwidth]{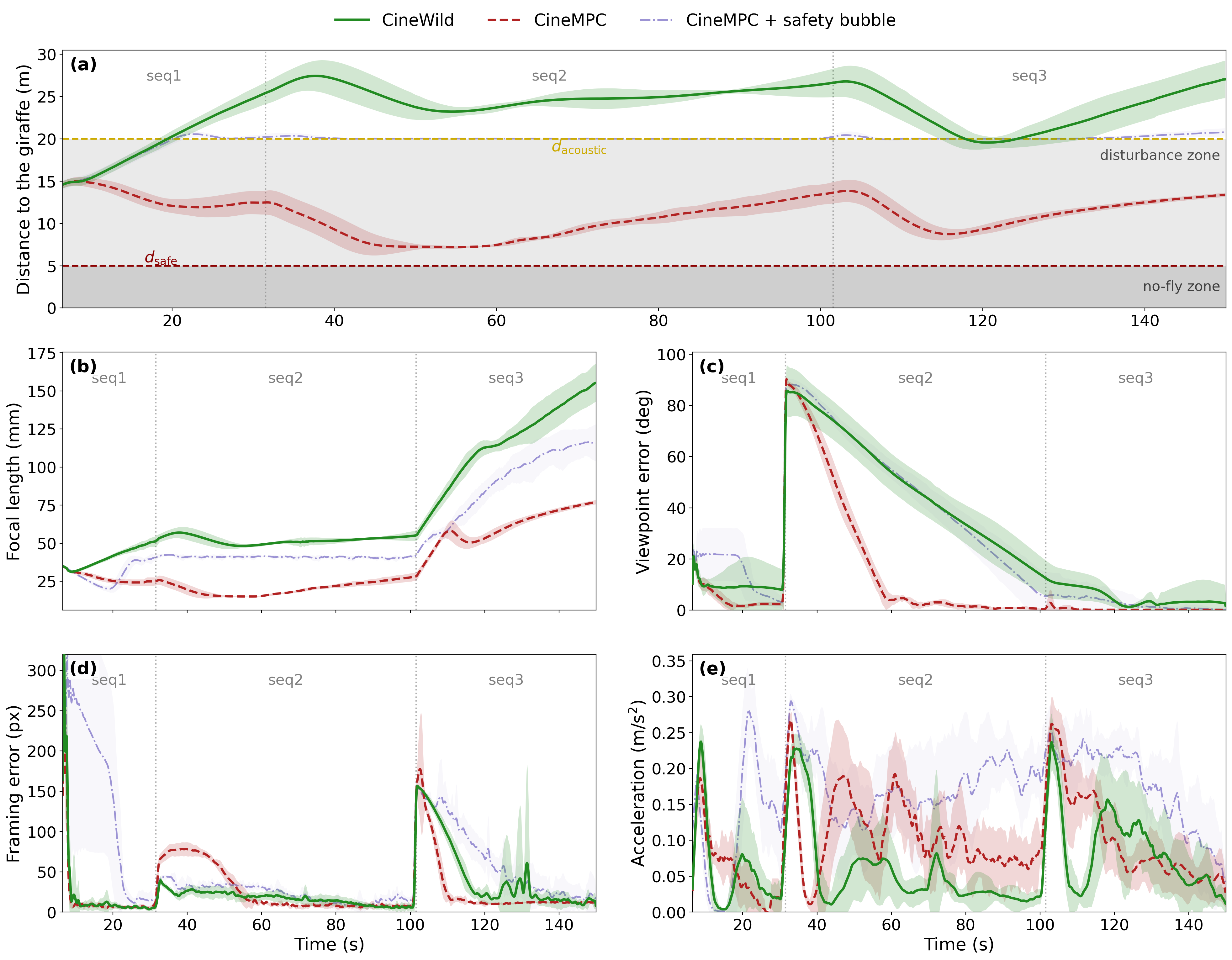}
\caption{\footnotesize{\rev{\textbf{Experiment~1. Quantitative comparison of CineWild + baselines over time} (mean $\pm$ std over 10 runs per method; vertical grey dotted lines: sequence changes; the welfare-aware baseline is drawn lighter). (a) Distance to the giraffe; light grey: disturbance zone, closer than $d_{\text{ac}}$; dark grey: no-fly zone, closer than $d_{\text{sf}}$. (b) Focal length, which compensates the distance to keep the requested framing. (c) Viewpoint error, difference between the requested and the achieved viewing angle around the animal; it jumps when a new viewpoint is requested (start of Sequence~2). (d) Framing error, image distance between the target and its requested position. (e) Acceleration, lowest on average for CineWild.}}}

\label{fig:exp1_quant}
\end{figure*}

\begin{table}[!t]
\centering
\caption{\rev{Experiment~1: mean over the 10 runs of each method (the spread across runs is shown in Fig.~\ref{fig:exp1_bars}), fixed weights (Table~\ref{table:weights}). Best value per row in bold. --: CineMPC never leaves the acoustic distance, so it is never near its boundary from outside.}}

\footnotesize
\setlength{\tabcolsep}{4pt}
\begin{tabular}{lccc}
\toprule
 & CineWild & CineMPC & CineMPC + bubble \\
\midrule
$d_{dt}$ min / mean [m] & \textbf{14.6} / \textbf{23.5} & 6.6 / 10.9 & 14.5 / 19.8 \\
time $d_{dt} < d_{\text{ac}}$ [\%] & \textbf{15.0} & 100 & 22.2 \\
time within 1\,m of $d_{\text{ac}}$ [\%] & \textbf{9.7} & -- & 90.9 \\
$e_u$ / $e_v$ [px] & \textbf{17.0} / 14.1 & 17.9 / \textbf{12.3} & 40.3 / 26.3 \\
$e_{\text{yaw}}$ [deg] & 27.2 & \textbf{10.9} & 26.8 \\
focal length [mm] & 71 & 35 & 55 \\
RMS acceleration [m/s$^2$] & \textbf{0.11} & 0.15 & 0.21 \\
\bottomrule
\end{tabular}
\label{table:exp1}
\end{table}

\subsubsection{Results}
Figure~\ref{fig:exp1_combined} compares first- and third-person views obtained with CineWild and the baseline. CineWild maintains consistent framing while operating outside the acoustic distance throughout all sequences, as confirmed by the third-person views. In contrast, the baseline frequently operates at closer distances, potentially violating proximity constraints, despite achieving visually comparable shots.
\begin{revblock}
Quantitative results are summarized in Table~\ref{table:exp1} and Figs.~\ref{fig:topdown}--\ref{fig:exp1_quant}. CineWild leaves the acoustic distance in all 10 runs and spends $15.0\%$  of the time inside it, mostly during the initial retreat, against 100\% for CineMPC, which never leaves it; neither enters the safe distance. As Fig.~\ref{fig:topdown}-a shows, CineMPC approaches the giraffe, whereas CineWild retreats behind it and circles to the side view from outside $d_{\text{ac}}$, at a mean distance of 23.5\,m. It compensates the distance with the zoom (mean focal length 71\,mm against 35\,mm, up to 155\,mm in the close-up, Fig.~\ref{fig:exp1_quant}-b), with similar framing errors (17.0/14.1 against 17.9/12.3\,px, Fig.~\ref{fig:exp1_quant}-d) and a lower RMS acceleration (0.11 against 0.15\,m/s$^2$, Figs.~\ref{fig:exp1_quant}-e and~\ref{fig:exp1_bars}). 

The CineMPC+bubble baseline also leaves the acoustic distance, but then moves around the constraint boundary: it spends 91\% of the time within 1\,m of $d_{\text{ac}}$ (10\% for CineWild)(Figs.~\ref{fig:exp1_bars}-b and ~\ref{fig:exp1_quant}-a). Its framing errors (40.3/26.3\,px) and RMS acceleration (0.21\,m/s$^2$) are about twice those of CineWild (Fig.~\ref{fig:exp1_quant}-d,e): while the drone is inside $d_{\text{ac}}$ its hard constraint is infeasible and the camera does not follow the shot, which produces framing errors of up to 300\,px during the initial retreat.

The price of the distance is the viewpoint: the mean viewpoint error of CineWild is 27.2$^\circ$, against 10.9$^\circ$ for CineMPC (Fig.~\ref{fig:exp1_quant}-c). It comes from Sequence~2, where the lateral viewpoint requires a turn of about $90^\circ$ around the animal: CineWild, like the bubble, performs it at more than 20\,m and brings the error below $30^\circ$ after 53\,s, against 19\,s for CineMPC at 8.8\,m. The requested rear view of Sequence~1 is reached in all runs of all methods.
\end{revblock}

\subsection{Experiment 2: Ethical Visibility-Aware Framing}
This experiment evaluates the visibility-aware formulation by recording a stationary tiger while keeping the drone outside the animal's field of view when operating nearby. A frontal viewpoint is desired through $J_p$, but the controller may relax it to avoid visual disturbance. The animal vision model uses
$(W_{\text{px}},H_{\text{px}},W_{\text{mm}},H_{\text{mm}})=(960,540,\rev{23.76,13.365})$,
$f_{\text{eye}}=35\,\mathrm{mm}$, and $d_{\text{vis}}=12\,\mathrm{m}$\rev{, i.e.\ a horizontal field of view of $37.5^\circ$, acting as an illustrative value}\limit{, narrower than the visual field of these kind of animals, which exceeds $180^\circ$~\cite{heesy2004relationship}; a wider field would push the drone further to the side or behind the animal}.
\rev{The proximity and smoothness terms are also included, with the same weights as in Experiment~1 (Table~\ref{table:weights}); for the tiger we use $d_{\text{ac}} = 8$\,m, below the requested filming distance, so that $J_{\text{prox}}$ only acts if the drone approaches closer than the shot requires. The drone always starts from the same pose, 10\,m in front of the tiger and just inside its field of view (Sec.~\ref{sec:sim_setup}).}

\paragraph*{Note: Infeasible Initialization}
\label{sec:heuristic}
The drone is intentionally initialized inside the animal's field of view, violating the hard constraint \rev{($\mathbf{x}_{d,k} \notin \mathcal{X}_{\mathrm{FoV}}$)} and causing temporary MPC infeasibility. A simple recovery heuristic moves the drone toward the left image boundary until feasibility is restored.
The heuristic gives no formal guarantee: it only moves the drone  out of the field of view and relies on the constraint becoming feasible again, which happened in all runs within 8.1--10.5\,s.

\subsubsection{Experiment objectives}
The experiment consists of three sequences, each with different control objectives.

\begin{figure*}[!b]
\centering
\setlength{\tabcolsep}{1.5pt}
\renewcommand{\arraystretch}{0.9}
\begin{tabular}{@{}cccc@{}}
    \includegraphics[width=0.24\textwidth]{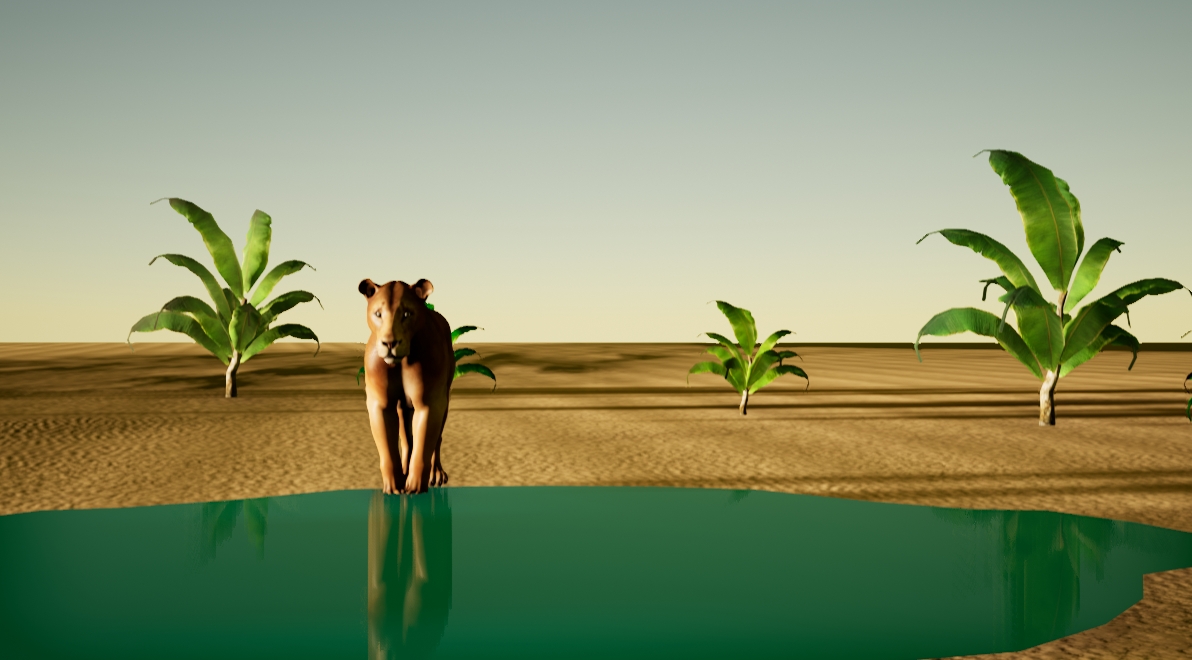} &
    \includegraphics[width=0.24\textwidth]{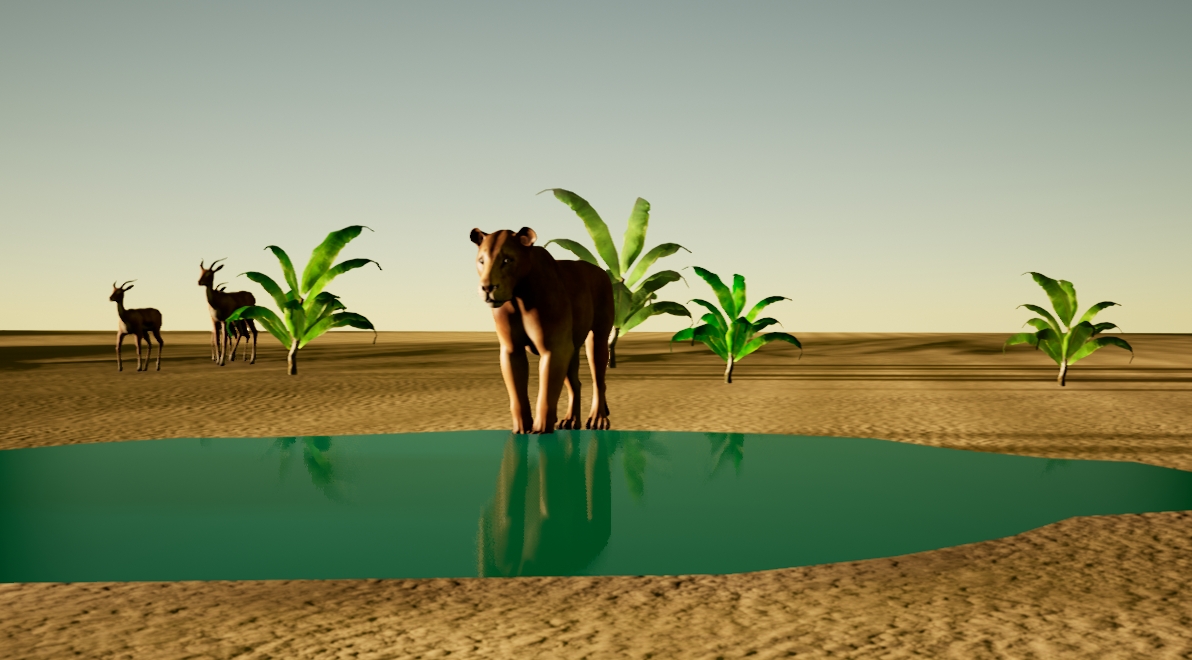} &
    \includegraphics[width=0.24\textwidth]{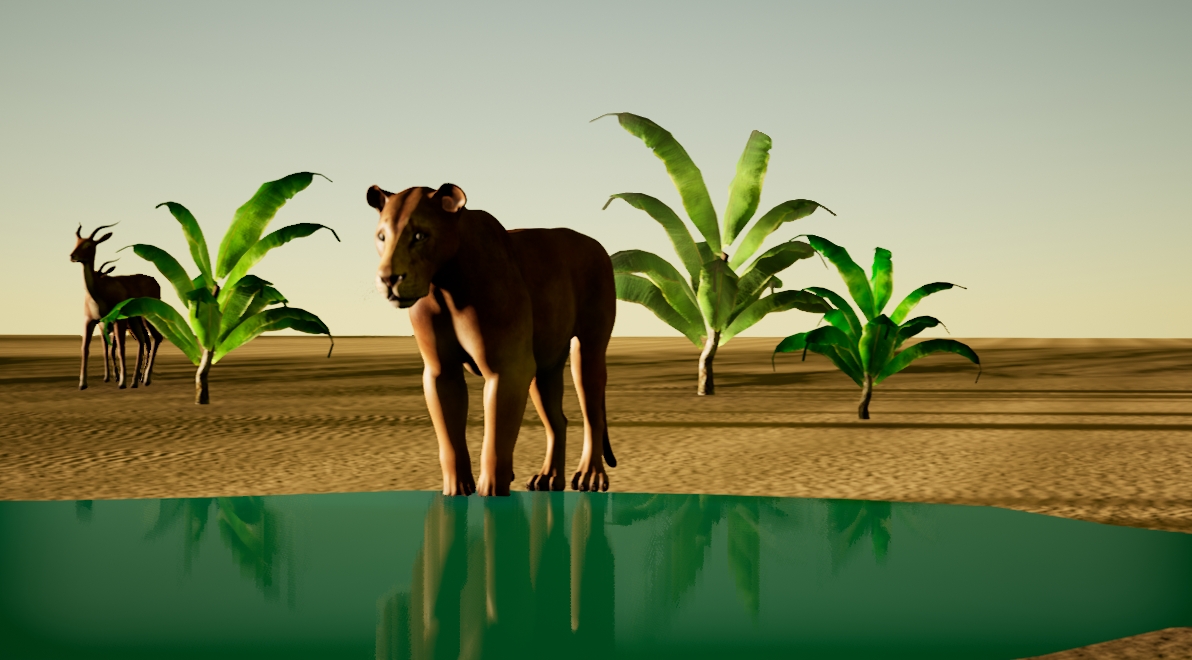} &
    \includegraphics[width=0.24\textwidth]{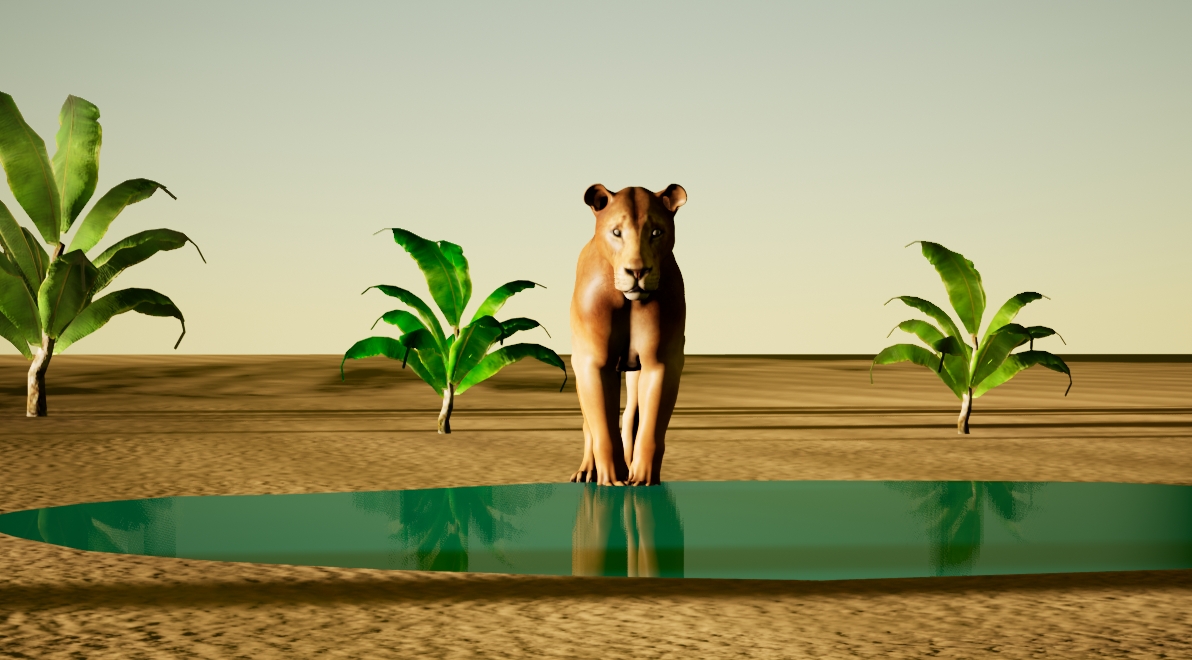} \\
    \includegraphics[width=0.24\textwidth]{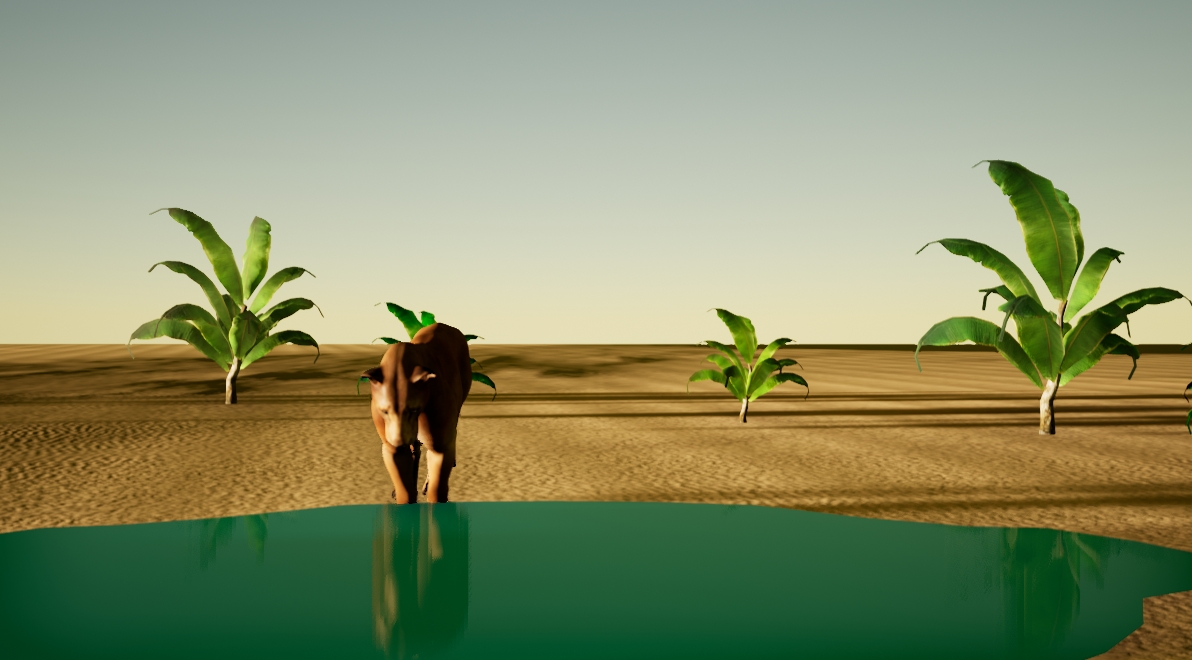} &
    \includegraphics[width=0.24\textwidth]{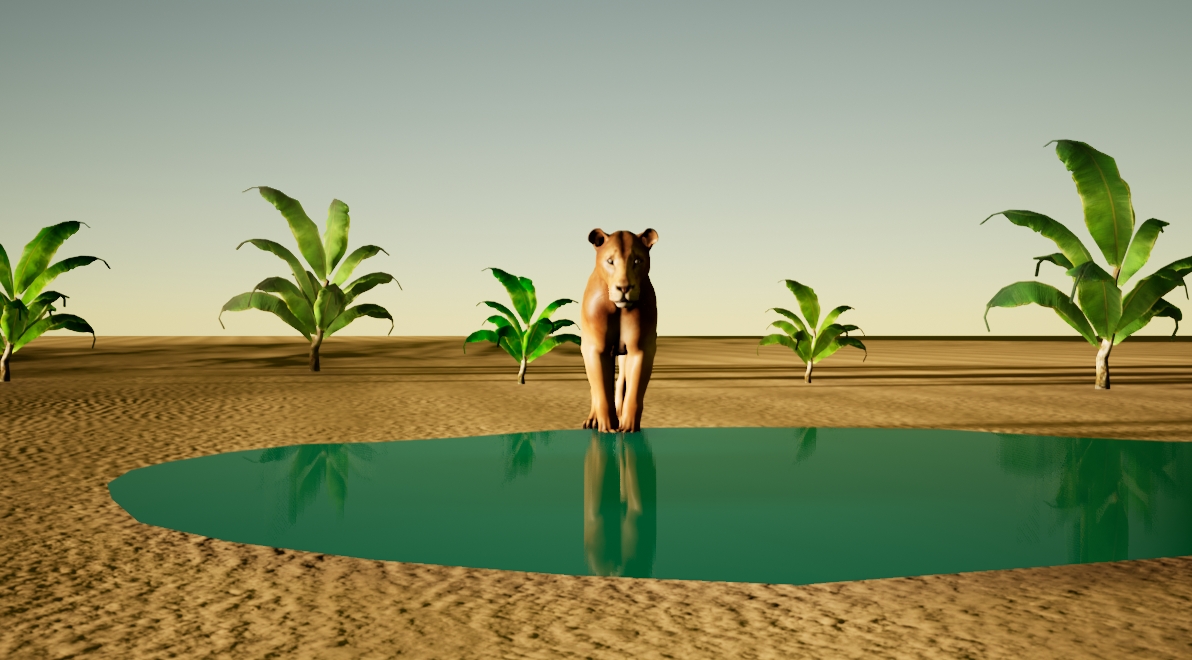} &
    \includegraphics[width=0.24\textwidth]{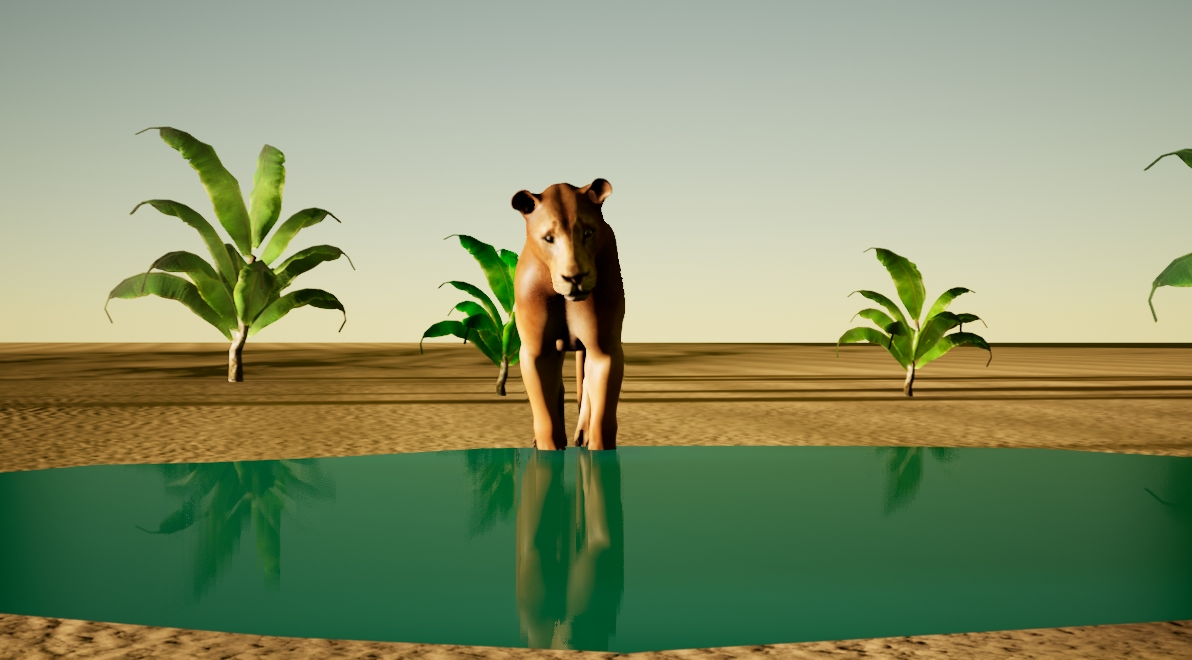} &
    \includegraphics[width=0.24\textwidth]{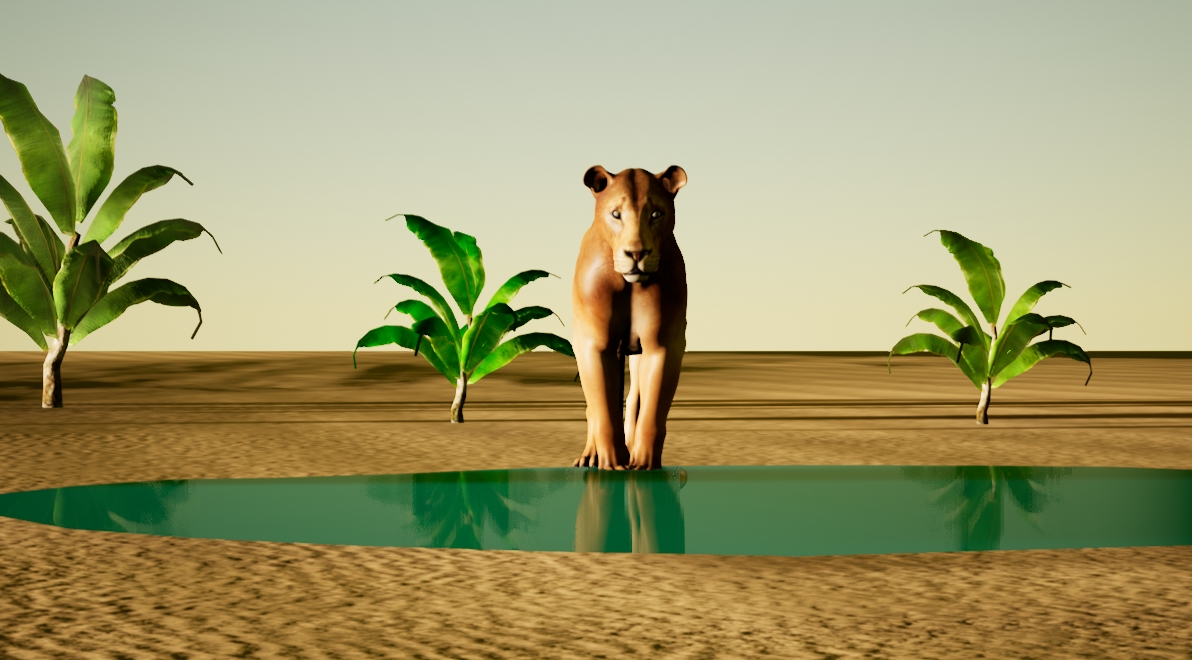} \\
    \multicolumn{4}{@{}p{\dimexpr\textwidth-3pt}@{}}{\centering\footnotesize {\textbf{CineWild} (top) and \textbf{CineMPC} (bottom): onboard camera.}} \\[4pt]
    \includegraphics[width=0.24\textwidth]{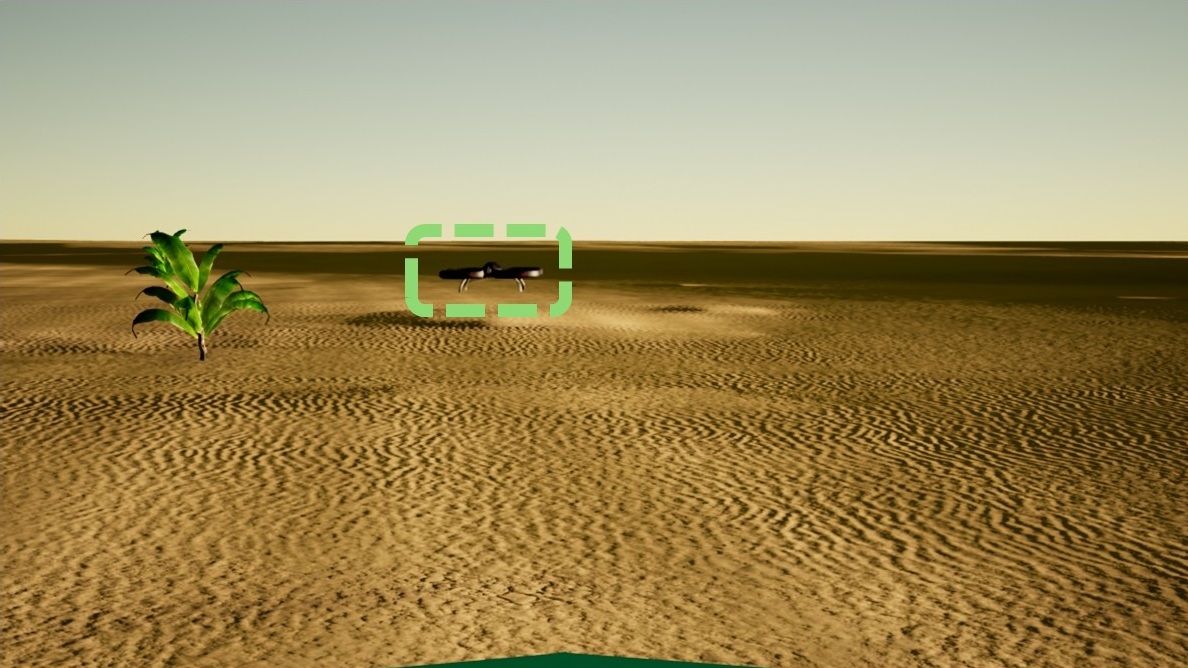} &
    \includegraphics[width=0.24\textwidth]{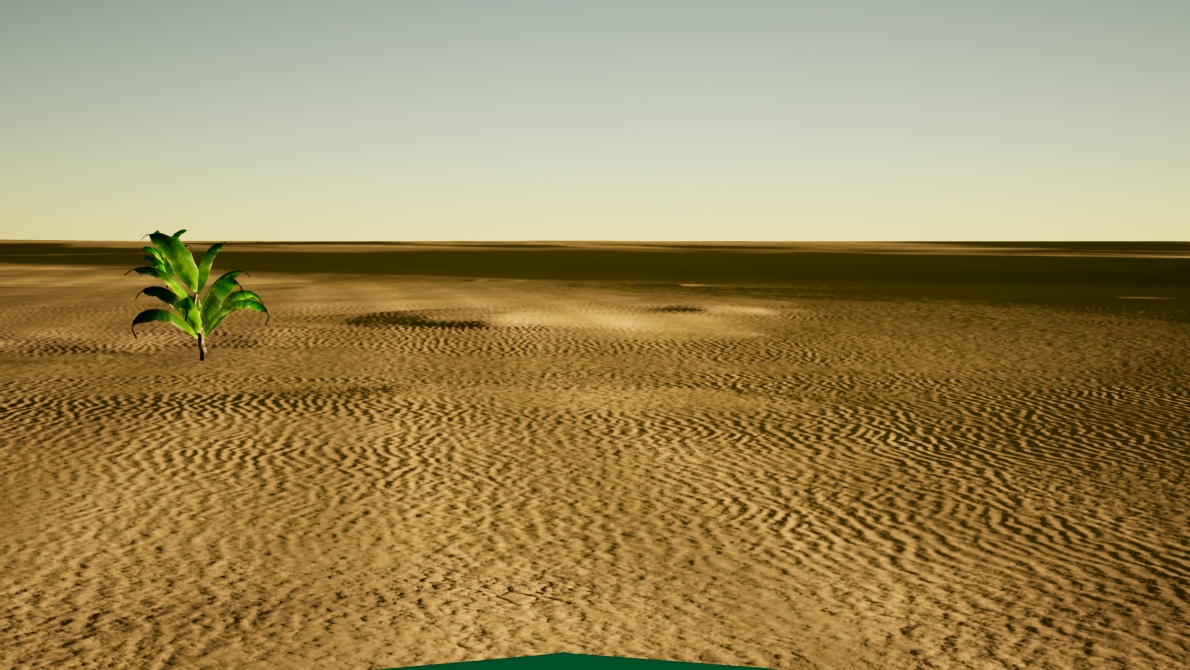} &
    \includegraphics[width=0.24\textwidth]{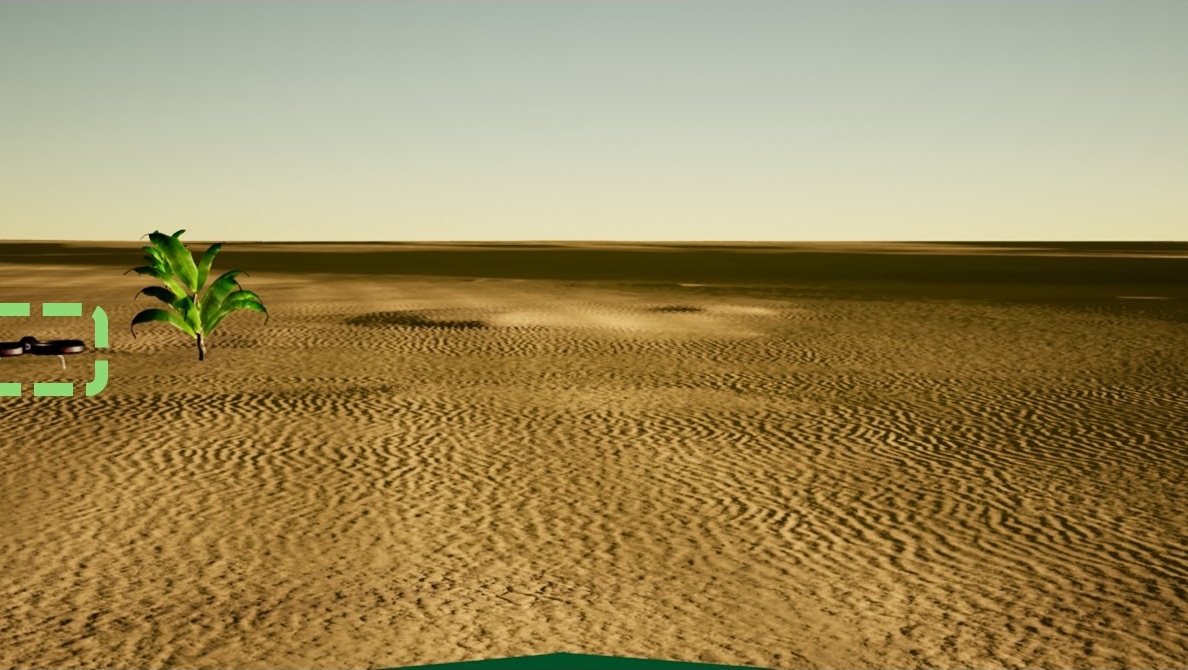} &
    \includegraphics[width=0.24\textwidth]{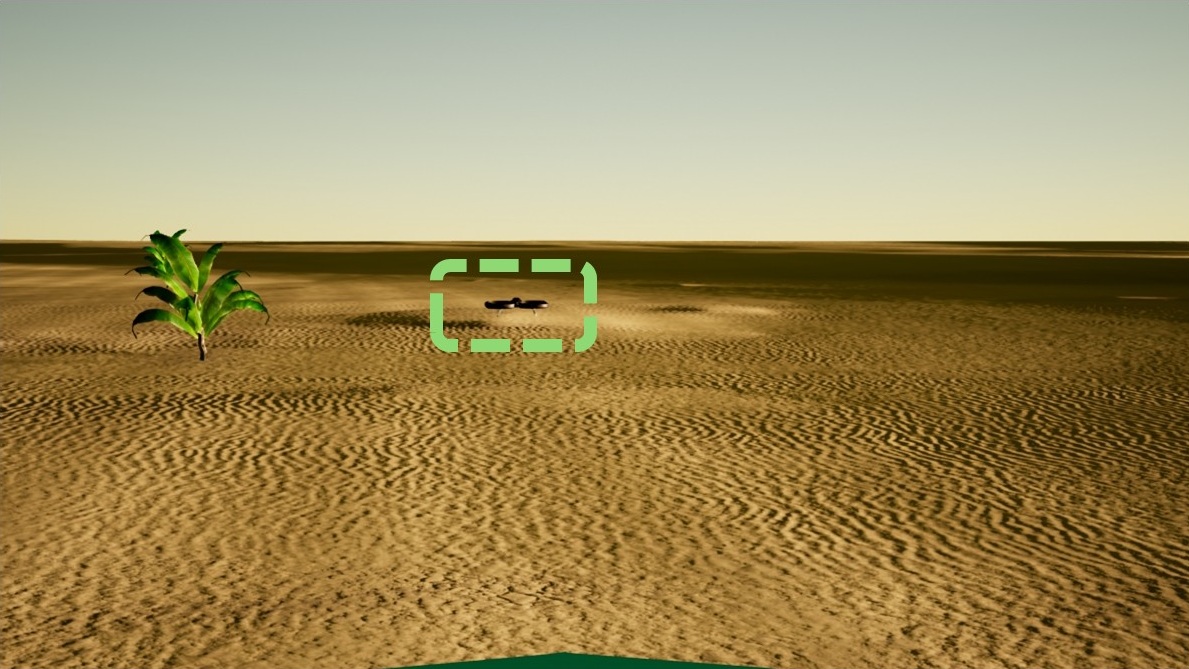} \\
    \includegraphics[width=0.24\textwidth]{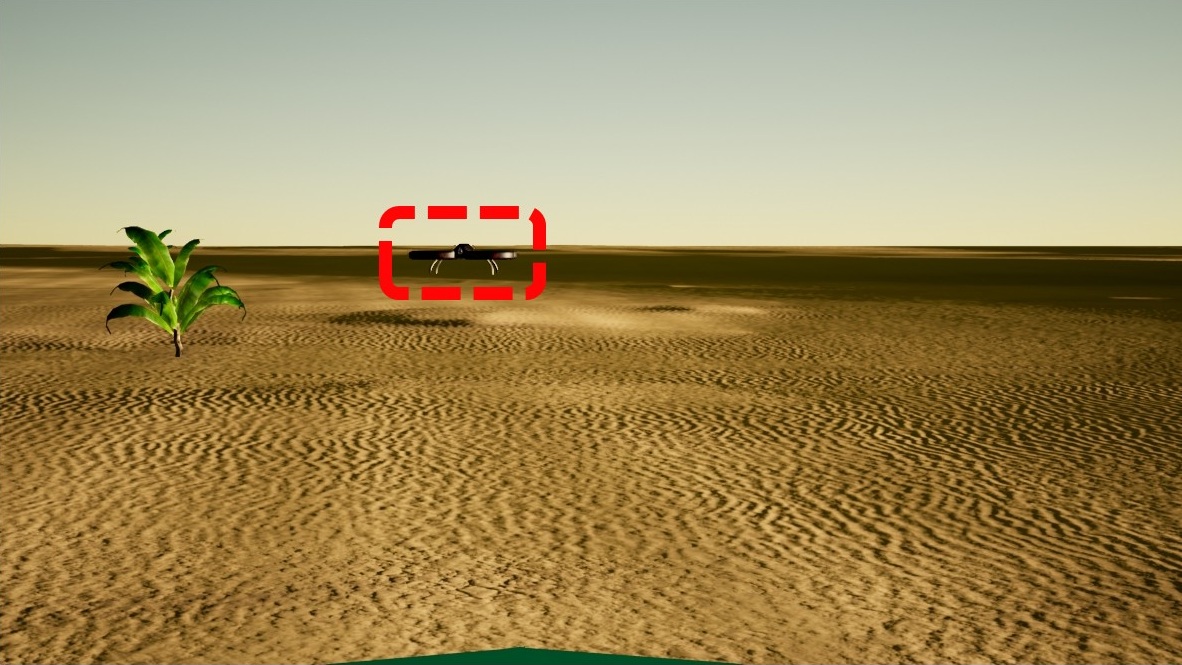} &
    \includegraphics[width=0.24\textwidth]{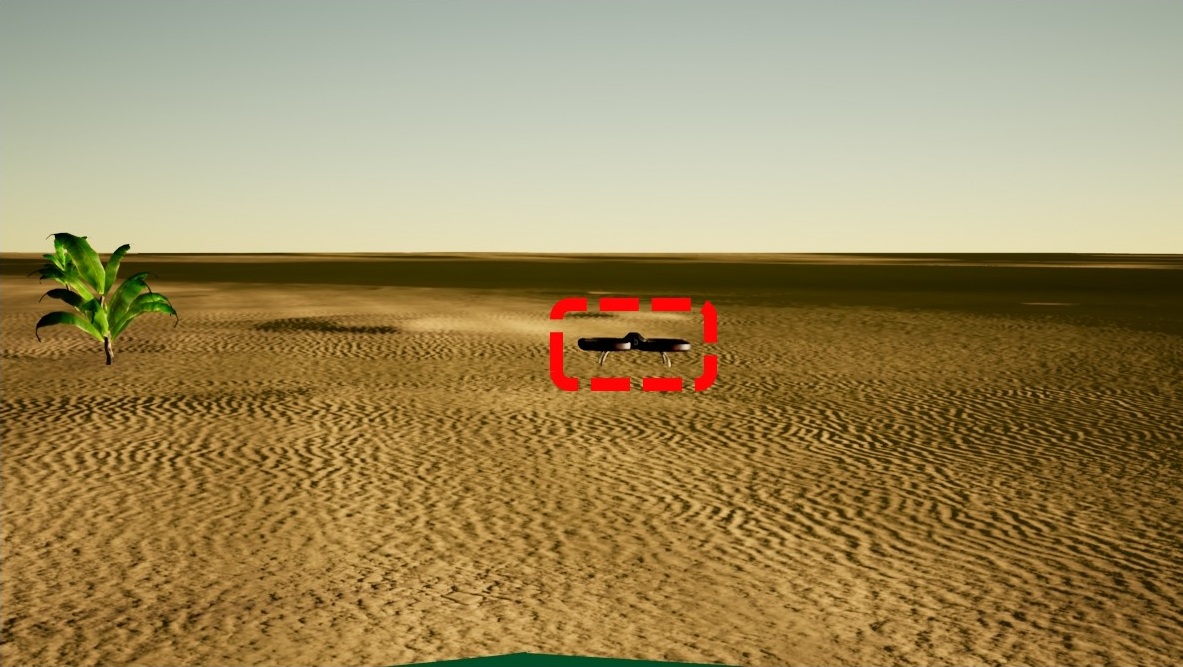} &
    \includegraphics[width=0.24\textwidth]{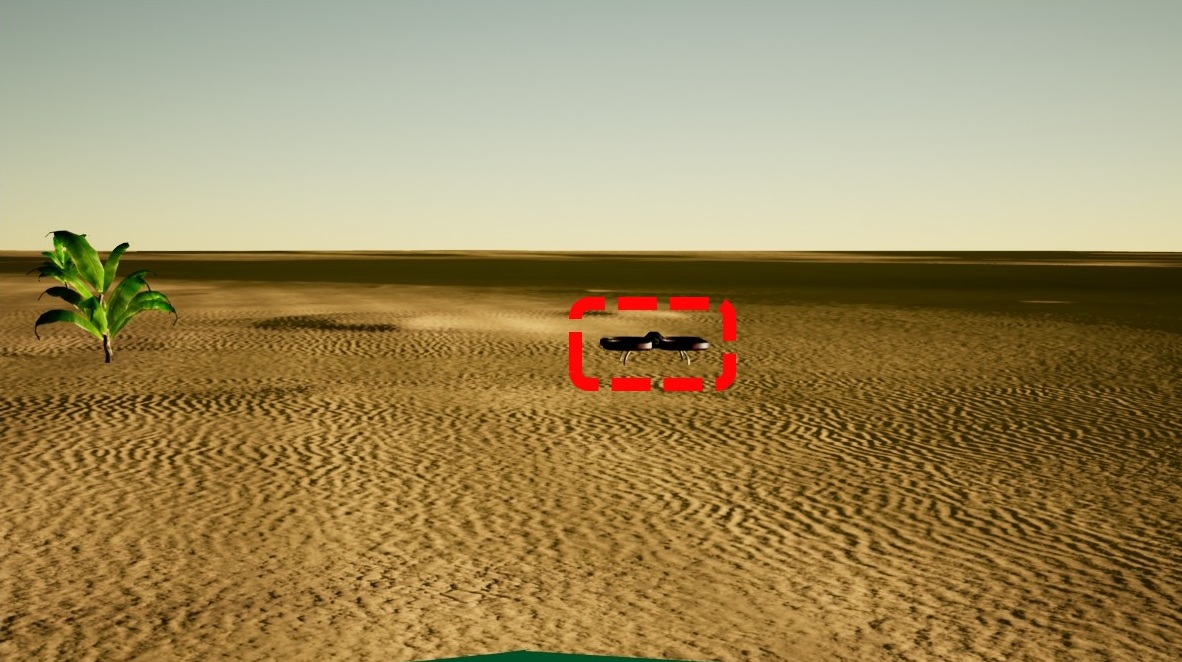} &
    \includegraphics[width=0.24\textwidth]{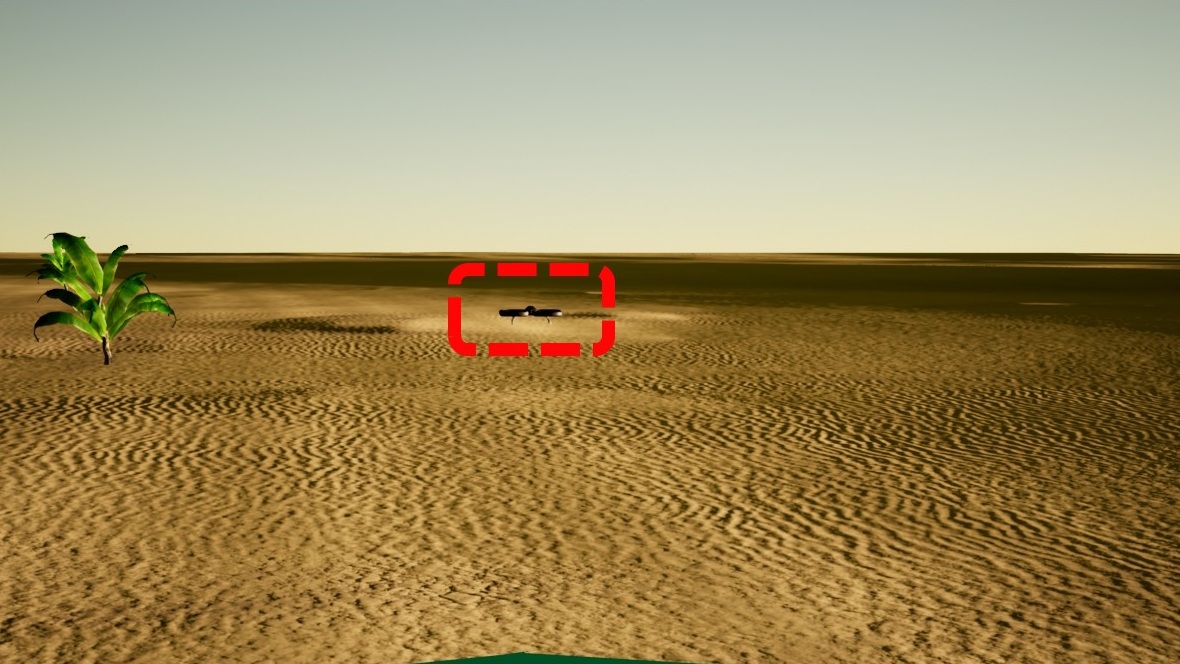} \\
    \multicolumn{4}{@{}p{\dimexpr\textwidth-3pt}@{}}{\centering\footnotesize {\textbf{CineWild} (top) and \textbf{CineMPC} (bottom): view from the tiger's eyes.}} \\
\end{tabular}
\caption{\footnotesize{\rev{\textbf{Experiment~2. Qualitative comparison of CineWild and CineMPC.} Frames are ordered left to right in time (start, and end of Sequences~1, 2 and~3). The onboard framing of the tiger is similar with both methods; however, the view from the tiger's eyes shows that, while very close to the animal, CineWild moves the drone to the border of the animal's field of view, whereas CineMPC keeps it in the centre, which is not appropriate. When drones are beyond $d_{\text{vis}}$ (third column on), a frontal shot is admissible, as shown for both methods in the frame in the last column.}}}
\label{fig:exp2_combined}
\end{figure*}

\paragraph{Sequence 1: Recovery and Frontal Framing}
The recovery behavior and the effect of the visibility cost $J_{\text{FoV}}$ are observed. Framing is enforced by $J_{\text{im}}$, while $J_p$ commands a frontal view at $d_{dt}^*=10\,\mathrm{m}$, illustrating the impact of visibility constraints. \rev{The start and the end of the sequence are shown in the first and second columns of Fig.~\ref{fig:exp2_combined}.}

\paragraph{Sequence 2: Closer Framing}
Screen presence is increased by adjusting $J_{\text{im}}$, achieved through focal zoom while maintaining visibility compliance. \rev{The end of the sequence is shown in the third column of Fig.~\ref{fig:exp2_combined}.}

\paragraph{Sequence 3: Frontal Shot from Beyond $d_{\text{vis}}$}
The drone records from a farther distance ($d_{dt}^*=15\,\mathrm{m}$), allowing frontal footage without visual disturbance when operating beyond $d_{\text{vis}}$. \rev{The end of the sequence is shown in the fourth column of Fig.~\ref{fig:exp2_combined}.}

\begin{figure*}[!t]
\centering
\includegraphics[width=0.99\textwidth]{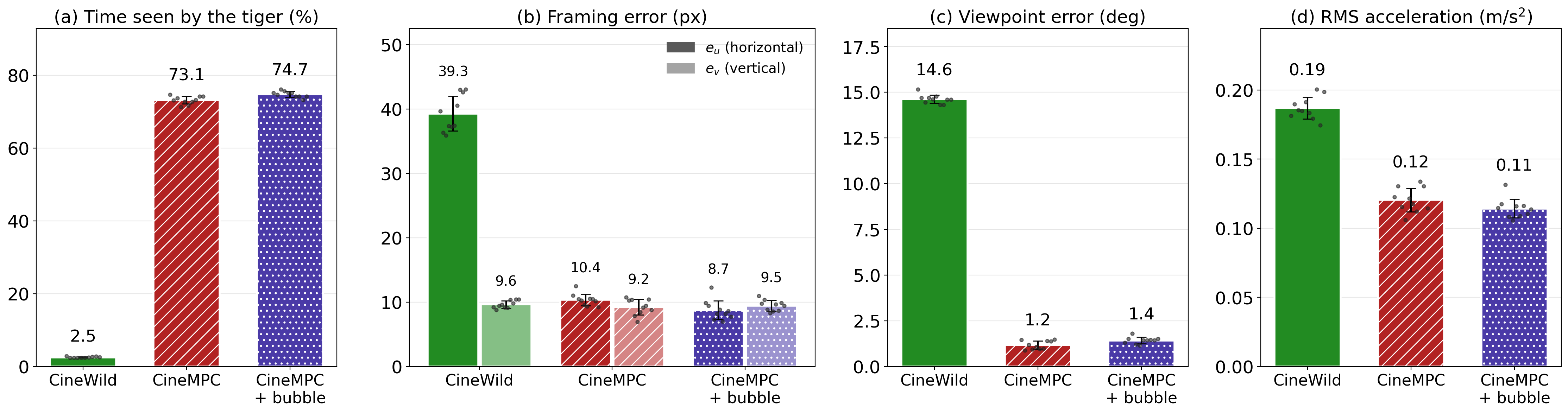}
\caption{\footnotesize{\rev{\textbf{Experiment~2. Comparison of the three methods} (mean $\pm$ std over 10 runs; dots: individual runs; lower is better in every panel). (a) Time in the tiger's field of view; (b) framing error, horizontal $e_u$ and, lighter, vertical $e_v$; (c) viewpoint error; (d) RMS acceleration. A safety distance does not keep the drone out of the animal's view: the bubble behaves as CineMPC. CineWild stays out of view at the cost of a larger viewpoint error, a transient framing error and more acceleration.}}}
\label{fig:exp2_bars}
\end{figure*}

\begin{figure*}[!t]
\centering
\includegraphics[width=0.9\textwidth]{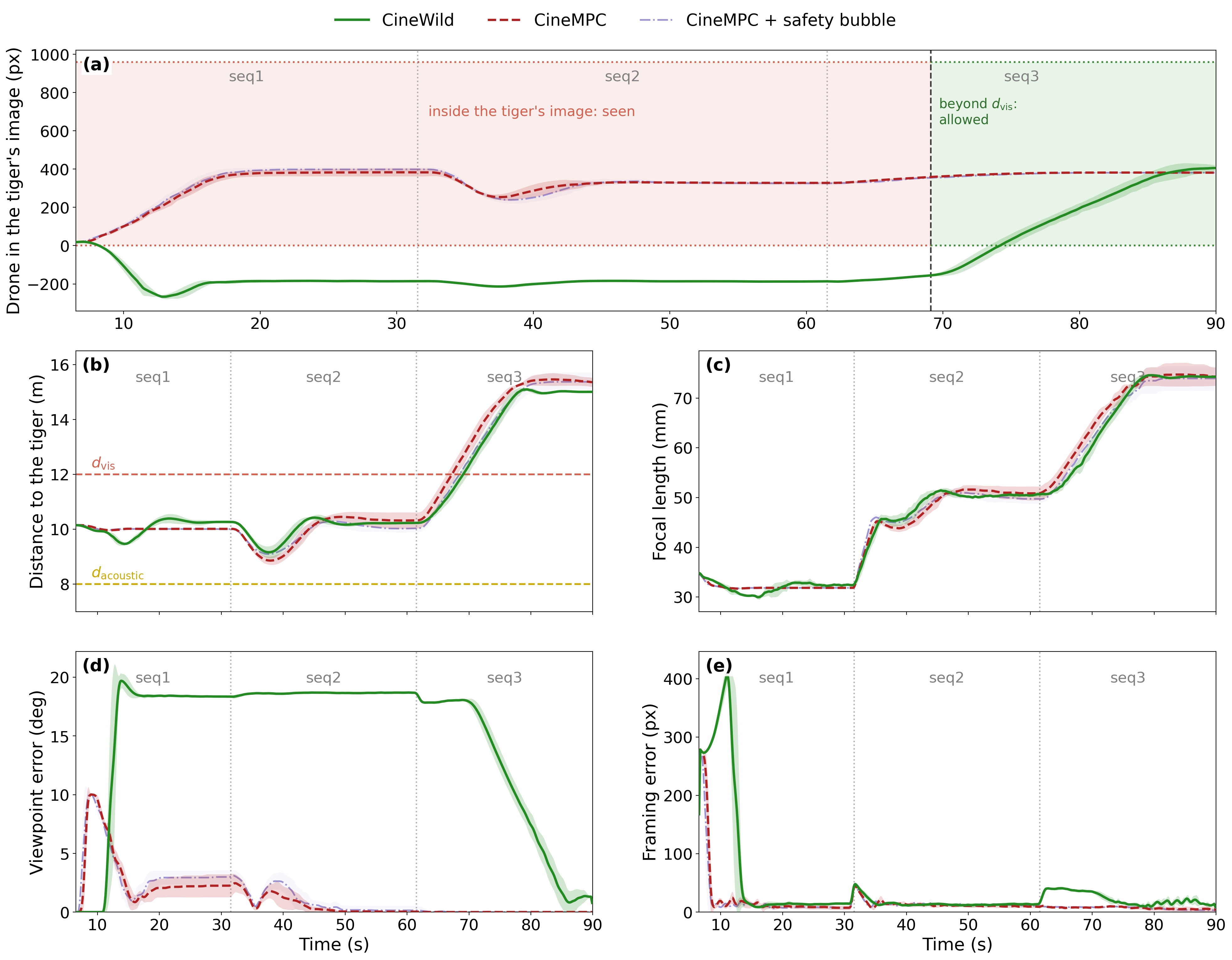}
\caption{\footnotesize{\rev{\textbf{Experiment~2. Quantitative comparison over time} (mean $\pm$ std over 10 runs; vertical grey dotted lines: sequence changes). (a) Horizontal position of the drone in the tiger's image (0--960\,px): red, inside the image while closer than $d_{\text{vis}}$, i.e.\ seen by the animal; green, beyond $d_{\text{vis}}$ (dashed line: median time at which CineWild moves beyond it; the baselines do so up to 2\,s earlier), where a frontal shot is allowed. (b) Distance to the tiger. (c) Focal length, which follows the distance to keep the requested framing. (d) Viewpoint error, difference between the requested and the achieved viewing angle around the animal. (e) Framing error, image distance between the target and its requested position. The RMS acceleration of each method is compared in Fig.~\ref{fig:exp2_bars}.}}}

\label{fig:exp2_quant}
\end{figure*}

\begin{table}[!t]
\centering
\caption{\rev{Experiment~2: mean over the 10 runs of each method (the spread across runs is shown in Fig.~\ref{fig:exp2_bars}), fixed weights (Table~\ref{table:weights}). Best value per row in bold. Drone position in the animal's image: averaged while in front of the animal and closer than $d_{\text{vis}}$; the image spans 0--960\,px.}}

\footnotesize
\setlength{\tabcolsep}{4pt}
\begin{tabular}{lccc}
\toprule
 & CineWild & CineMPC & CineMPC + bubble \\
\midrule
time in animal FoV [\%] & \textbf{2.5} & 73.1 & 74.7 \\
drone position in animal image [px] & \textbf{$-$180} & 310 & 314 \\
$d_{dt}$ min [m] & \textbf{9.1} & 8.8 & \textbf{9.1} \\
$e_u$ / $e_v$ [px] & 39.3 / 9.6 & 10.4 / \textbf{9.2} & \textbf{8.7} / 9.5 \\
$e_{\text{yaw}}$ [deg] & 14.6 & \textbf{1.2} & 1.4 \\
focal length [mm] & 49 & 50 & 49 \\
RMS acceleration [m/s$^2$] & 0.19 & 0.12 & \textbf{0.11} \\
\bottomrule
\end{tabular}
\label{table:exp2}
\end{table}

\subsubsection{Results}
\label{sec:e2_results}
\rev{Figure~\ref{fig:exp2_combined} compares onboard views and views from the tiger's eyes for CineWild and CineMPC. Both obtain a similar framing of the tiger, but CineWild leaves the centre of the animal's view after the intentionally infeasible initialization, whereas CineMPC remains in front of it, in its field of view, for most of the experiment.}

\begin{revblock}
Quantitative results are summarized in Table~\ref{table:exp2} and Figs.~\ref{fig:topdown}-b, \ref{fig:exp2_bars} and~\ref{fig:exp2_quant}. All methods start just inside the animal's field of view. CineWild leaves it after about 8\,s in all runs and only enters it again for the frontal shot of Sequence~3, once it is beyond $d_{\text{vis}}$ (Fig.~\ref{fig:topdown}-b). It spends 2.5\% of the time in the field of view, all of it during the initial recovery, against 73.1\% for CineMPC, which is seen in every run until Sequence~3 (Fig.~\ref{fig:exp2_quant}-a). Table~\ref{table:exp2} summarizes the difference: while close to the animal, CineWild keeps the drone projected at $-180$\,px, outside the animal's image [0--960\,px], whereas CineMPC stays at 310\,px, inside it, near its centre (480\,px).

The shots themselves are comparable. The same requested distances are compensated with the same focal lengths (49--50\,mm on average, Fig.~\ref{fig:exp2_quant}-b,c); the framing errors are similar once CineWild is out of the field of view; its larger mean horizontal error (39.3 against 10.4\,px) comes from the first 12\,s, while the drone leaves the field of view (Fig.~\ref{fig:exp2_quant}-e). 
The price of staying out of view is the viewpoint: since visibility is a hard constraint and the viewpoint only a soft objective, CineWild relaxes the frontal viewpoint while it is close to the animal (mean error 14.6$^\circ$ against 1.2$^\circ$) and reaches it only in Sequence~3 (Fig.~\ref{fig:exp2_quant}-d). Staying at the edge of the view cone also requires small, frequent corrections, which raise the RMS acceleration when using CineWild (0.19 against 0.12\,m/s$^2$, Fig.~\ref{fig:exp2_bars}).

The welfare-aware baseline behaves as CineMPC (74.7\% of the time in the field of view): the requested filming distance (10\,m) is beyond $d_{\text{ac}} = 8$\,m, so its safety distance never becomes active. A distance rule does not prevent the animal from seeing the drone; the visibility terms of CineWild do.
\end{revblock}

\begin{revblock}
\subsection{Ablation Study}
\label{sec:ablation}
To isolate the contribution of each welfare term, we remove them one at a time from CineWild and repeat both experiments with the protocol of Sec.~\ref{sec:sim_setup} and the weights of Table~\ref{table:weights}. 
\subsubsection{Ablations}
\begin{itemize}
\item \emph{w/o $J_{\text{prox}}$}: $w_{\text{prox}} = 0$; the hard constraint $d_{dt} \ge d_{\text{sf}}$ is kept.
\item \emph{w/o FoV terms}: $w_{\text{FoV}} = 0$ and  hard visibility constraint removed. Both are removed together; the soft term only acts next to the boundary.
\item \emph{w/o $J_{\text{smooth}}$}: $w_{\text{smooth}} = 0$.
\end{itemize}

\begin{figure*}[!t]
\centering
\includegraphics[width=\textwidth]{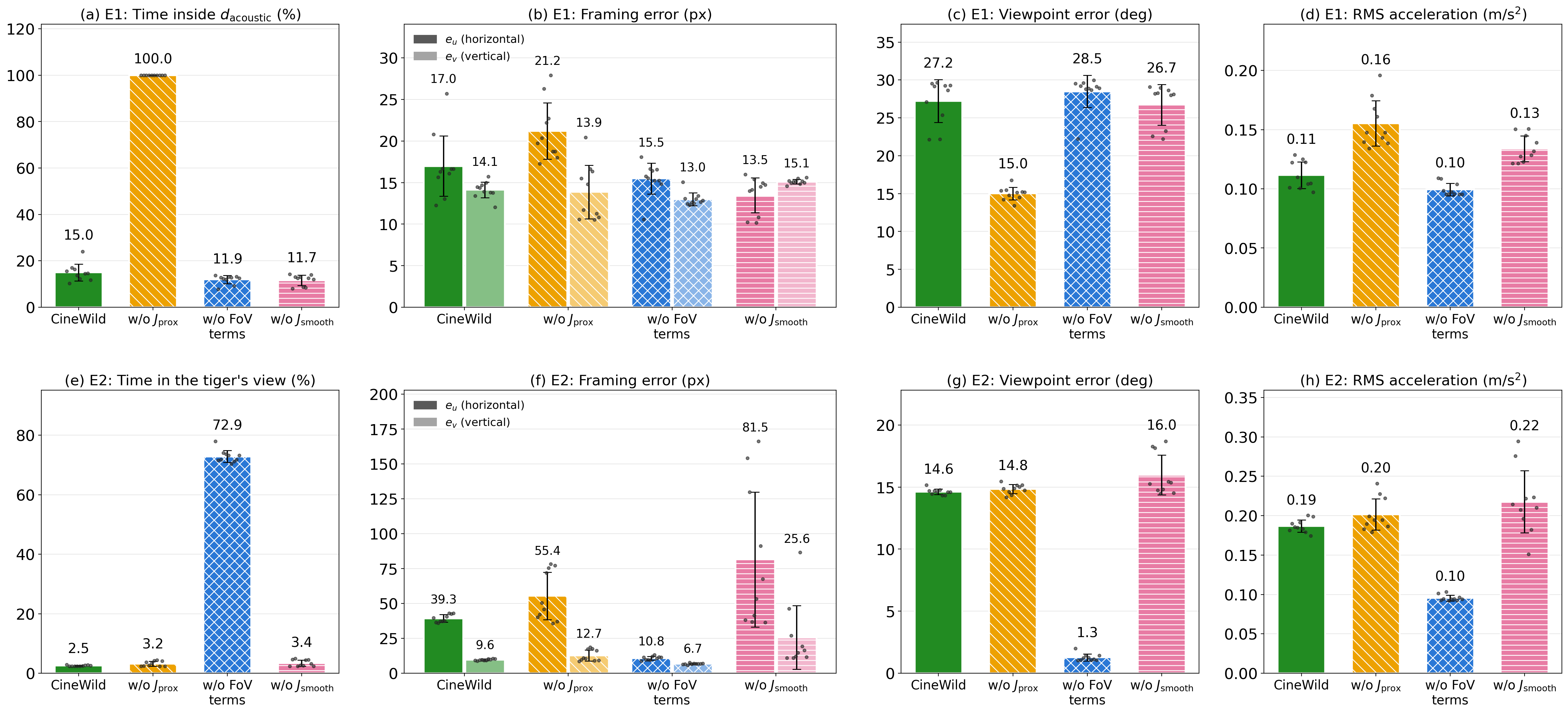}
\caption{\footnotesize{\rev{\textbf{Ablation study} (mean $\pm$ std over 10 runs per configuration; dots: individual runs). Top row: Experiment~1; bottom row: Experiment~2. Each welfare term acts on its own metric: $J_{\text{prox}}$ on the time inside the acoustic distance (a), the visibility terms on the time in the animal's field of view (e), and $J_{\text{smooth}}$ on the acceleration (d, h).}}}
\label{fig:ablation}
\end{figure*}

\begin{table}[!t]
\centering
\caption{\rev{Ablation study: mean over the 10 runs of each configuration (the spread is shown in Fig.~\ref{fig:ablation}). Best value of each column within each experiment in bold.}}
\label{table:ablation}

\footnotesize
\setlength{\tabcolsep}{3pt}
\begin{tabular}{llccccccc}
\toprule
 & Config & $d_{dt}$ min [m] & $t_{<d_{\text{ac}}}$ [\%] & $t_{\text{FoV}}$ [\%] & $e_u$ [px] & $e_v$ [px] & $e_{\text{yaw}}$ [deg] & $a_{\text{RMS}}$ [m/s$^2$] \\
\midrule
\multirow{4}{*}{E1} & CineWild & \textbf{14.6} & 15.0 & 0 & 17.0 & 14.1 & 27.2 & 0.11 \\
 & w/o $J_{\text{prox}}$ & 6.5 & 100 & 0 & 21.2 & 13.9 & \textbf{15.0} & 0.16 \\
 & w/o FoV & 14.5 & 11.9 & 0 & 15.5 & \textbf{13.0} & 28.5 & \textbf{0.10} \\
 & w/o $J_{\text{smooth}}$ & \textbf{14.6} & \textbf{11.7} & 0 & \textbf{13.5} & 15.1 & 26.7 & 0.13 \\
\midrule
\multirow{4}{*}{E2} & CineWild & \textbf{9.1} & \textbf{0} & \textbf{2.5} & 39.3 & 9.6 & 14.6 & 0.19 \\
 & w/o $J_{\text{prox}}$ & 8.6 & 1.5 & 3.2 & 55.4 & 12.7 & 14.8 & 0.20 \\
 & w/o FoV & \textbf{9.1} & \textbf{0} & 72.9 & \textbf{10.8} & \textbf{6.7} & \textbf{1.3} & \textbf{0.10} \\
 & w/o $J_{\text{smooth}}$ & 8.8 & \textbf{0} & 3.4 & 81.5 & 25.6 & 16.0 & 0.22 \\
\bottomrule
\end{tabular}
\end{table}

\subsubsection{Results}

Results are summarized in Table~\ref{table:ablation} and Fig.~\ref{fig:ablation}. 

\emph{Proximity.} Without $J_{\text{prox}}$ the drone stays within the acoustic distance for the whole of Experiment~1 (100\% against 15.0\%, mean distance 11.9 against 23.5\,m, Fig.~\ref{fig:ablation}-a). This also causes worse footage, with a larger framing error (21.2 against 17.0\,px) and more acceleration (0.16 against 0.11\,m/s$^2$). In Experiment~2 the requested distance (10\,m) is already beyond $d_{\text{ac}} = 8$\,m, so $J_{\text{prox}}$ is rarely active.

\emph{Visibility.} In Experiment~1 the drone is never in the animal's field of view, so removing the visibility terms has no clear effect. In Experiment~2, which is designed to test the visibility terms, removing them raises the time of the drone inside of the field of view from 2.5\% to 72.9\% (Fig.~\ref{fig:ablation}-e); it also removes the drawback of staying out of view (see \ref{sec:e2_results}), bringing the viewpoint error from 14.6$^\circ$ to 1.3$^\circ$ and lowering the acceleration. 

\emph{Smoothness.} Without $J_{\text{smooth}}$ the drone accelerates more in both experiments: the RMS acceleration rises from 0.11 to 0.13\,m/s$^2$ in Experiment~1 and from 0.19 to 0.22\,m/s$^2$ in Experiment~2 (Fig.~\ref{fig:ablation}-d,h). Since rotor noise grows with rotor effort, this term keeps the drone quieter, reducing the disturbance to the animal. Moving more smoothly also makes the shot more stable: in Experiment~2, without it, the drone escapes the animal's field of view more abruptly, increasing the framing error (Fig.~\ref{fig:ablation}-f).

\paragraph{Prediction horizon}
Table~\ref{table:horizon} compares CineWild in Experiment~1 with $N = 5$ and $N = 10$ (2 and 4\,s), from the same start as the main runs. With $N = 10$ the rear view of Sequence~1 is still reached in all runs, and the distance and the acceleration barely change, but the shot degrades: the mean viewpoint error rises from 27.2$^\circ$ to 45.6$^\circ$, mostly in Sequences~2 and~3, and the framing error from 17.0/14.1 to 26.7/16.2\,px. Over 4\,s, the constant-velocity prediction of the animal and of the requested shot is no longer accurate, and the controller commits to trajectories that no longer match the shot. A 6\,s horizon ($N = 15$) exceeded the real-time budget of our laptop, which also runs the simulator, and was not evaluated. $N = 5$ is therefore used throughout.

\begin{table}[!t]
\centering
\caption{\rev{Prediction horizon (Experiment~1, CineWild, same start as the main runs; $N = 5$: the 10 main runs, $N = 10$: 5 runs). Best value per column in bold.}}
\label{table:horizon}

\footnotesize
\setlength{\tabcolsep}{3pt}
\begin{tabular}{lccccc}
\toprule
$N$ & rear view & $e_{\text{yaw}}$ [deg] & $e_u$/$e_v$ [px] & $a_{\text{RMS}}$ [m/s$^2$] & $t_{<d_{\text{ac}}}$ [\%] \\
\midrule
5 (2\,s) & 10/10 & \textbf{27.2} & \textbf{17.0} / \textbf{14.1} & 0.11 & 15.0 \\
10 (4\,s) & 5/5 & 45.6 & 26.7 / 16.2 & \textbf{0.10} & \textbf{13.3} \\
\bottomrule
\end{tabular}
\end{table}

\paragraph{Sensitivity}
Table~\ref{table:robustness} varies $w_{\text{prox}}$ and $d_{\text{ac}}$ (5 runs each, starting 9\,m from the animal and $45^\circ$ off its rear axis). The weight barely matters: from 5 to 45, the time within $d_{\text{ac}}$ changes by less than two points, and the framing and acceleration stay the same, so it does not need tuning per scenario. The species threshold does: with $d_{\text{ac}} = 10$, 20 and 30\,m, the drone retreats to each threshold and stays outside it, zooming in further the farther it has to stay (to compensate the distances, the mean focal lengths increase from 53, to 75 and 88\,mm). The time within $d_{\text{ac}}$ is the time it takes to get there from the start.

\begin{table}[!t]
\centering
\caption{\rev{Sensitivity to $w_{\text{prox}}$ and $d_{\text{ac}}$ (Experiment~1 from a start 9\,m from the animal, full configuration, mean over the runs; $t_{<d_{\text{ac}}}$ computed with the $d_{\text{ac}}$ of each row).}}
\label{table:robustness}

\footnotesize
\setlength{\tabcolsep}{2.5pt}
\begin{tabular}{llccccc}
\toprule
 & Config & runs & $t_{<d_{\text{ac}}}$ [\%] & $\bar d_{dt}$ [m] & $e_u$/$e_v$ [px] & $a_{\text{RMS}}$ [m/s$^2$] \\
\midrule
reference & $w_{\text{prox}} = 15$, $d_{\text{ac}} = 20$\,m & 10 & 22.5 & 24.0 & 13.7 / 15.8 & 0.10 \\
\midrule
\multirow{2}{*}{$w_{\text{prox}}$} & 5 & 5 & 23.8 & 23.8 & 14.2 / 14.8 & 0.09 \\
 & 45 & 5 & 22.0 & 25.5 & 13.7 / 15.1 & 0.10 \\
\midrule
\multirow{2}{*}{$d_{\text{ac}}$} & 10\,m & 5 & 3.8 & 16.3 & 21.4 / 11.9 & 0.11 \\
 & 30\,m & 5 & 48.9 & 27.2 & 18.3 / 16.3 & 0.11 \\
\bottomrule
\end{tabular}
\end{table}
\begin{figure}[!b]
\centering
\includegraphics[width=0.7\columnwidth]{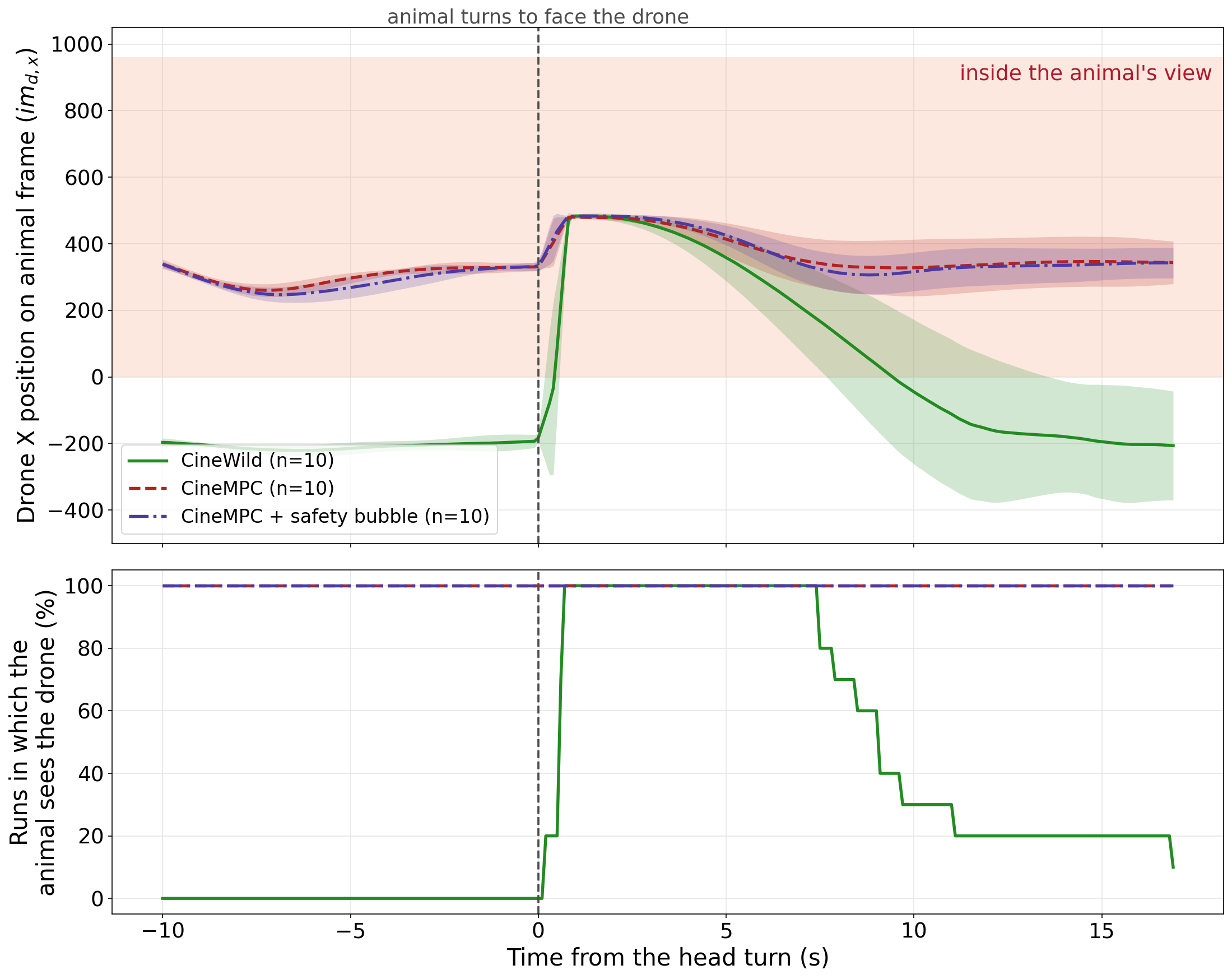}
\caption{\textbf{Head turn} (Experiment~2, 10 runs per method): the tiger turns to face the drone at $t = 0$. Top: horizontal position of the drone in the animal's image (mean $\pm$ std; shaded: inside the animal's view). Bottom: share of runs in which the animal sees the drone. Only CineWild leaves the field of view again.}
\label{fig:headturn}
\end{figure}

\paragraph{Head turn}
Animals do not keep a fixed gaze: they turn their heads frequently to scan their surroundings \cite{wilson2020orientation}, and they may orient towards a sound or a moving object such as an approaching drone, sometimes abruptly. A drone that is outside the field of view can therefore end up in it, and a hard visibility constraint then becomes infeasible. To evaluate this case, we repeat Experiment~2 and apply an instantaneous head turn at $t=45$\,s during Sequence~2, rotating the tiger so that the drone appears at the centre of its image. This represents a worst-case scenario, since the controller receives no advance notice and only tracks the measured animal orientation. As in the infeasible initialization case, the heuristic moves the drone left until the visibility constraint becomes feasible again. CineWild exits the animal's field of view in all runs, after $10.7 \pm 3.8$\,s on average (Fig.~\ref{fig:headturn}). During recovery, no feasible solution is found in 71\% of the steps and the framing error increases to $252 \pm 44$\,px, as leaving the field of view takes priority. In contrast, CineMPC and the welfare-aware baseline remain visible throughout all runs because they do not enforce visibility constraints.

\end{revblock}

\subsection*{\textbf{Real-World Validation}}
\rev{We conduct two real-world proof-of-concept experiments. Their goal  to show that the cost terms can be evaluated from real perception (RGB-D detection of the animal and motion-capture localisation of the drone), that the controller runs in real time on this data, and that its camera commands (focal length) are executed by a real zoom camera. They are not a closed-loop validation of the flight behavior, which is evaluated in simulation, due to hardware limitations.}
Due to ethical and practical constraints, controlled experiments with live wildlife are infeasible; instead, a lion puppet is used to enable safe and repeatable testing.
For conciseness, we report only representative experimental conditions and outcomes, while full runs and transitions are shown in the accompanying videos.
To isolate controller behavior, translational commands generated by the MPC are executed manually by the operator. \rev{The translational loop is therefore closed by a human pilot, and the experiments do not demonstrate autonomous flight; closing this loop with the autopilot is left to future field tests.}
\begin{figure}[!t]
    \centering
    \begin{tabular}{cc}
    \includegraphics[width=0.35\textwidth, height=3cm]{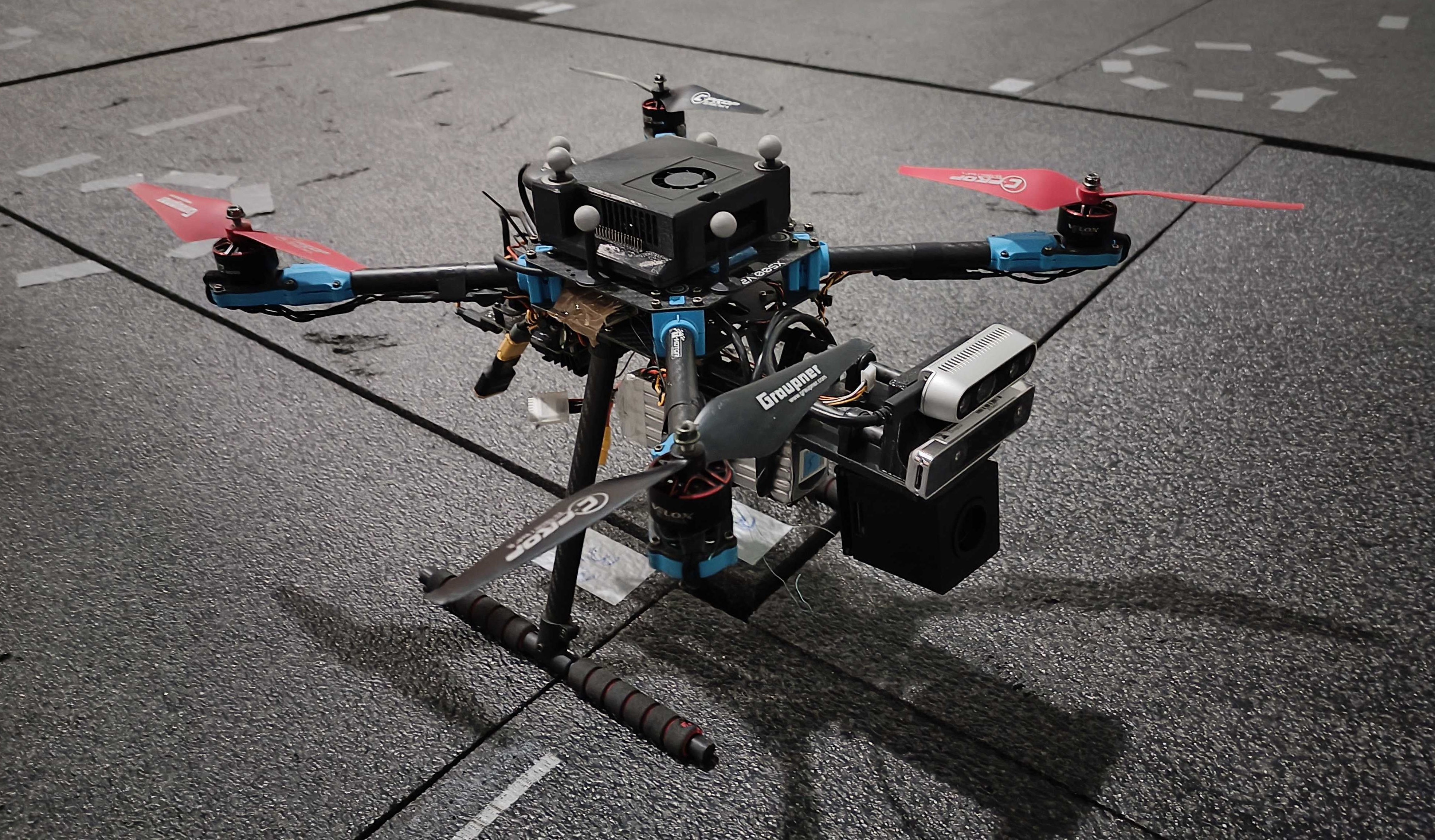} &
    \includegraphics[width=0.35\textwidth,height=3cm]{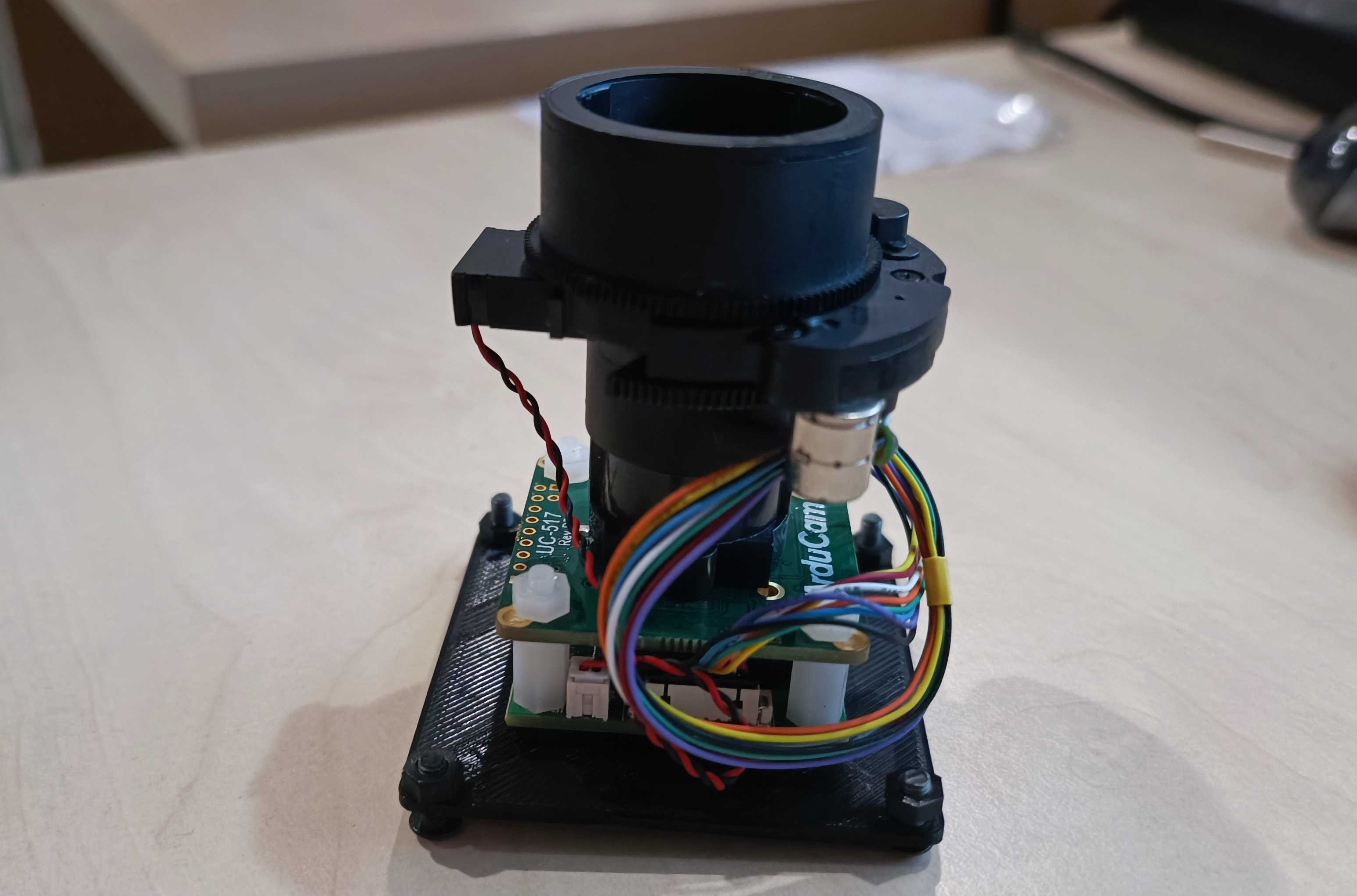} \\   
    \includegraphics[width=0.35\textwidth,height=3cm]{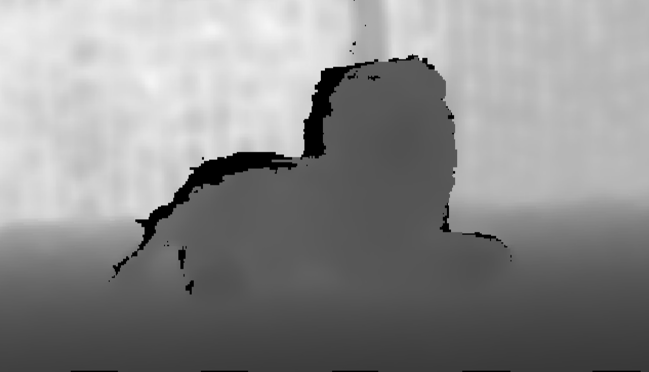} &    \includegraphics[width=0.35\textwidth,height=3cm]{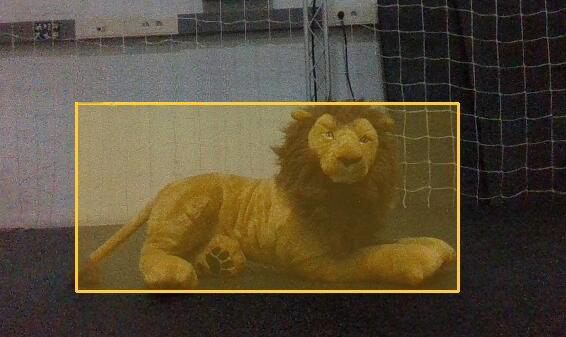} \\
    \end{tabular}
\caption{\footnotesize{\textbf{Real-world hardware and perception pipeline.}  
Top-left: The quadrotor platform used in the real experiments, equipped with an Intel RealSense RGB-D sensor, a Jetson Nano, and motion-capture markers.  
Top-right: The ArduCam PTZ camera used for intrinsic-parameter control (zoom and focus).  
Bottom: RGB-D outputs from the onboard perception pipeline, showing the detected puppet and the associated depth information used for target pose estimation.}}
    \label{fig:real_depth}
\end{figure}

\subsection{Real Experimental Setup}

The MPC runs offboard on a workstation (described in Sec.~\ref{sec:sim_setup}) and communicates with the platform via ROS. The drone (Fig.~\ref{fig:real_depth}-a) carries an ArduCam PTZ camera with controllable focal length and focus (Fig.~\ref{fig:real_depth}-b), an Intel RealSense RGB-D camera, and a Jetson Nano for target pose estimation and camera control. Drone pose is provided by an OptiTrack system, while the puppet pose is obtained from the RGB-D perception pipeline (Fig.~\ref{fig:real_depth}). The controller is configured following the setup described in Sec.~\ref{sec:sim_setup}.

\begin{figure}[!t] \centering \begin{tabular}{ccc} \includegraphics[width=0.30\columnwidth]{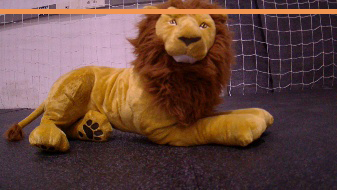} & \includegraphics[width=0.30\columnwidth]{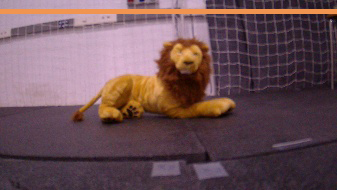} & \includegraphics[width=0.30\columnwidth]{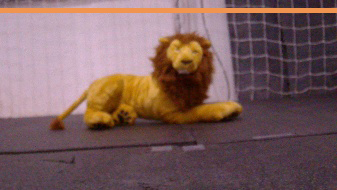} \\ 
\includegraphics[width=0.30\textwidth]{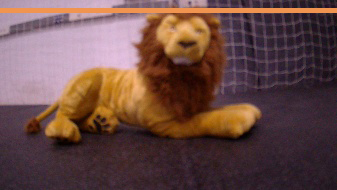} & \includegraphics[width=0.30\textwidth]{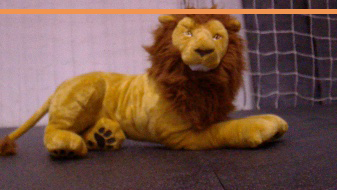} & \includegraphics[width=0.30\textwidth]{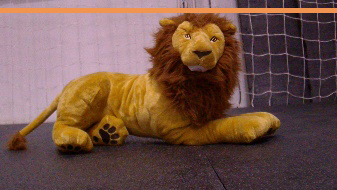} \\ 
\includegraphics[width=0.30\textwidth]{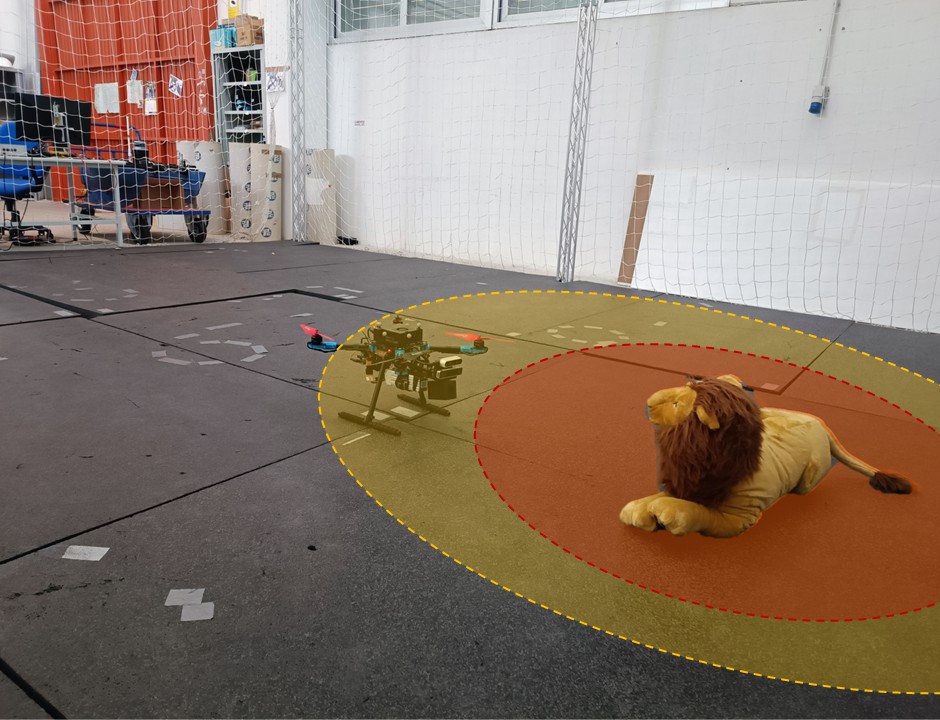} & \includegraphics[width=0.30\textwidth]{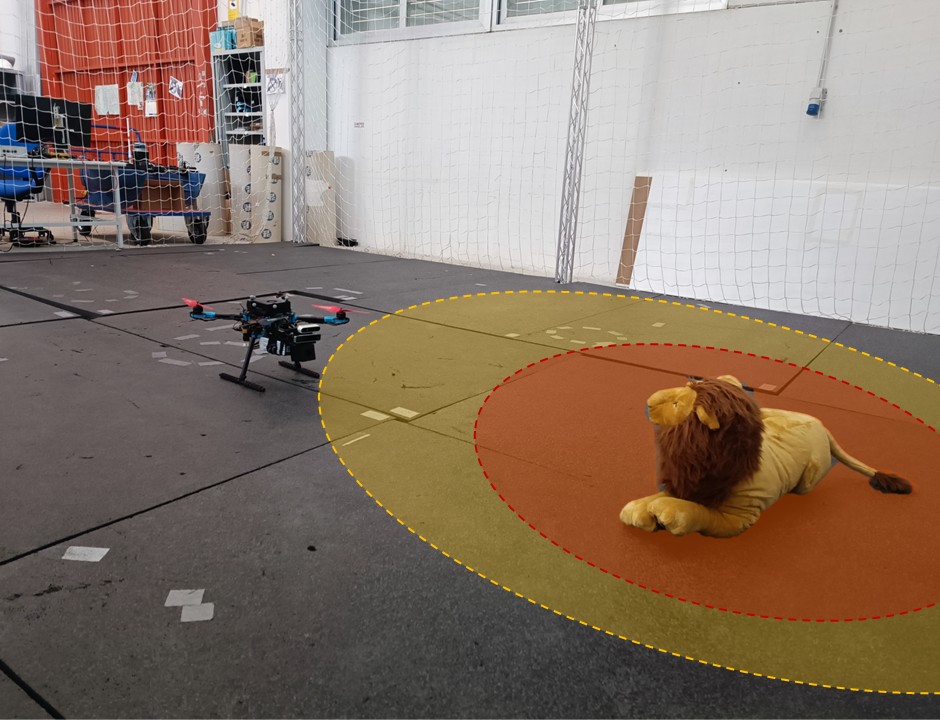} & \includegraphics[width=0.30\textwidth]{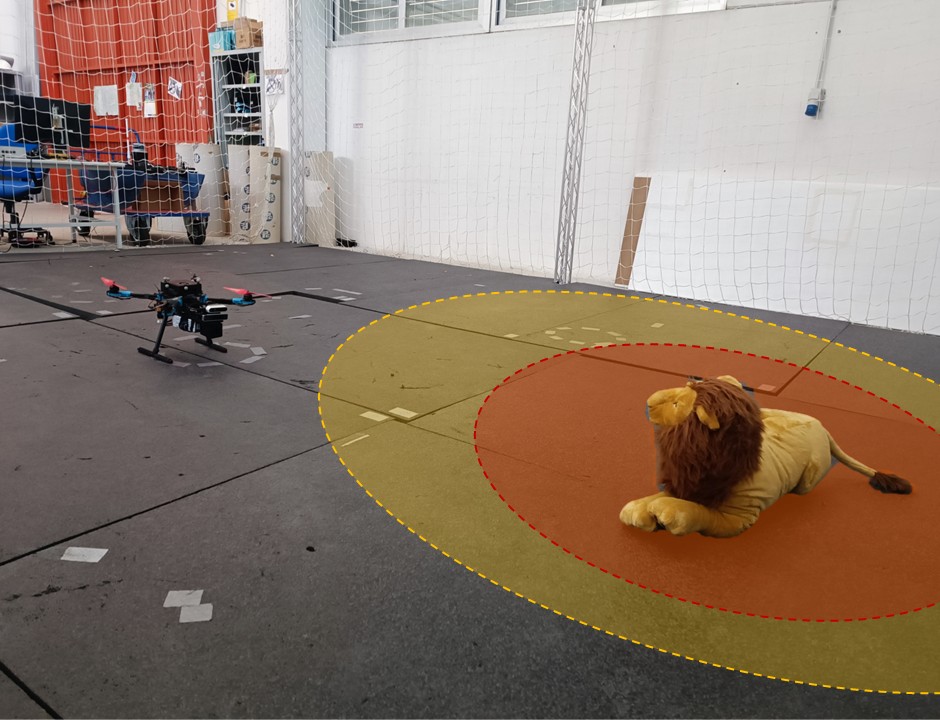} \\ 
\end{tabular} 
\caption{\footnotesize{\textbf{Views from Experiment~3.}
Frames are ordered left to right. 
Top: initial onboard views showing target framing and the pixel guideline (orange).
Middle: framing after automatic focal-length adjustment as drone--target distance increases.
Bottom: third-person views with physical distance and post-processed No-Fly (red) and Caution (yellow) zones, following Fig.~\ref{fig:costs_objectives}.}}
\label{fig:exp3_qual}
\end{figure}

\subsubsection{Experiment 3: Real-World Validation of Proximity-Aware Control}
\begin{figure}[t]
    \centering
    \begin{tabular}{cc}
        \includegraphics[width=0.48\columnwidth]{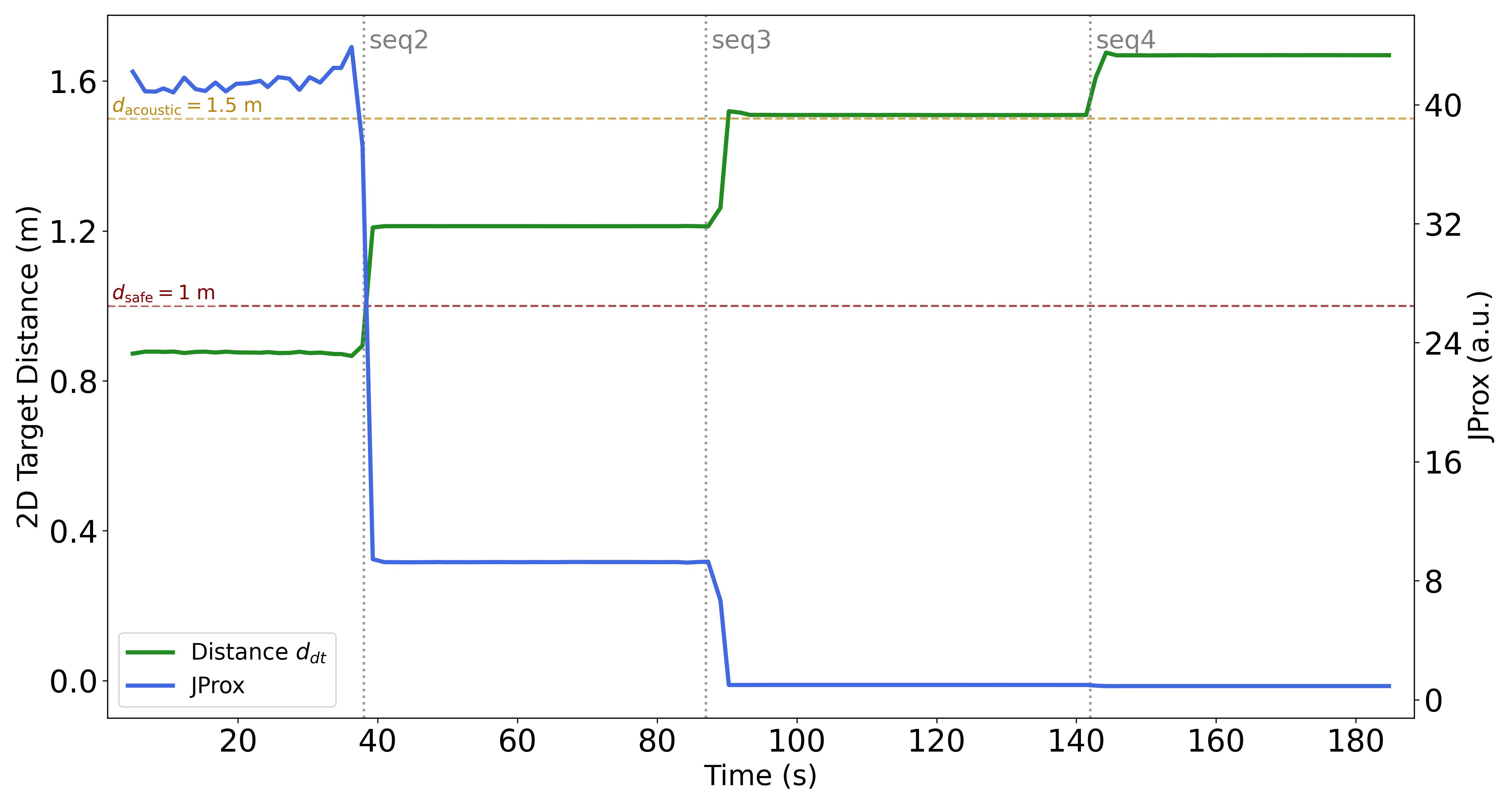} &
        \includegraphics[width=0.48\columnwidth]{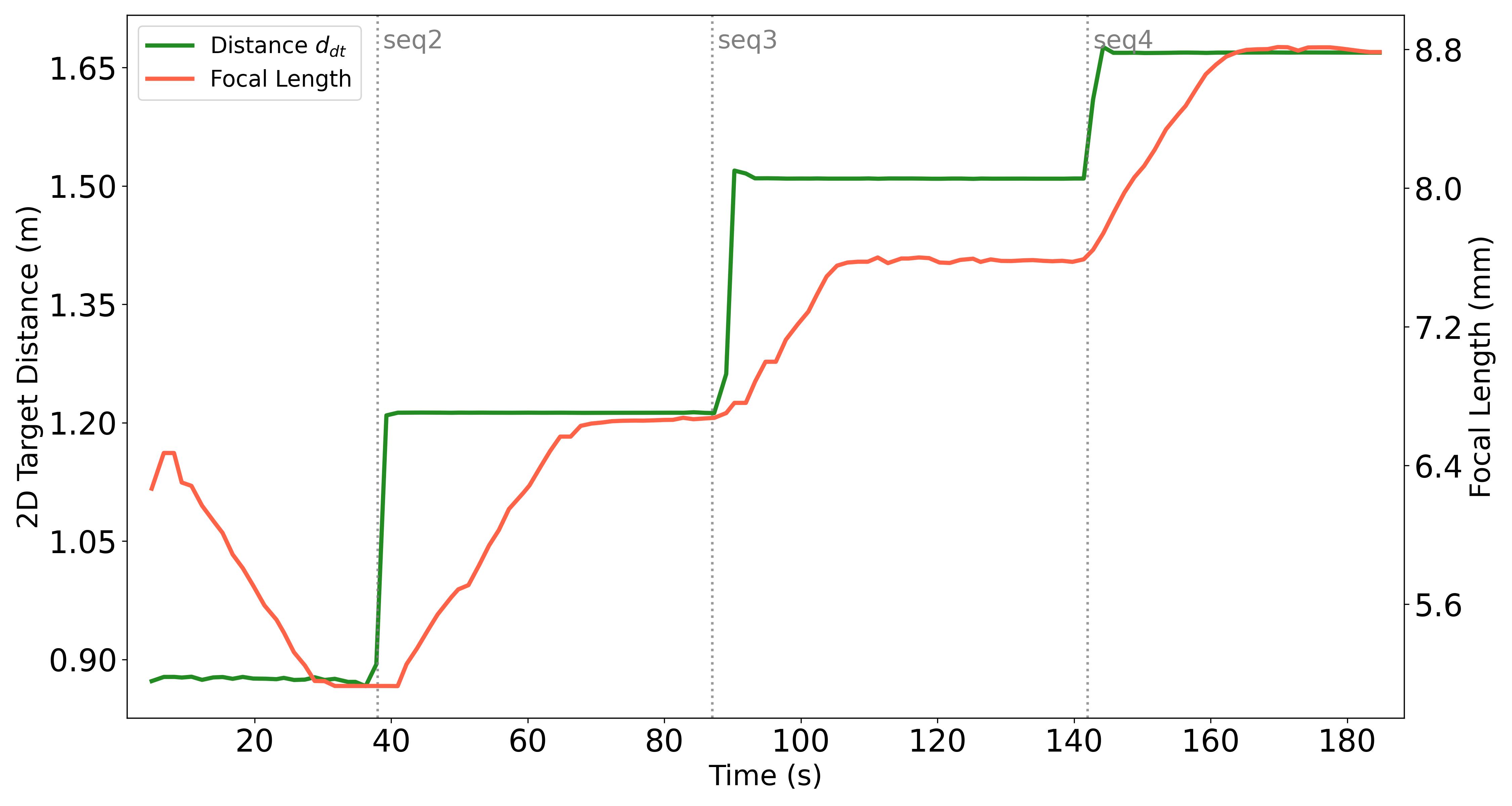} \\
        \multicolumn{2}{c}{\includegraphics[width=0.48\columnwidth]{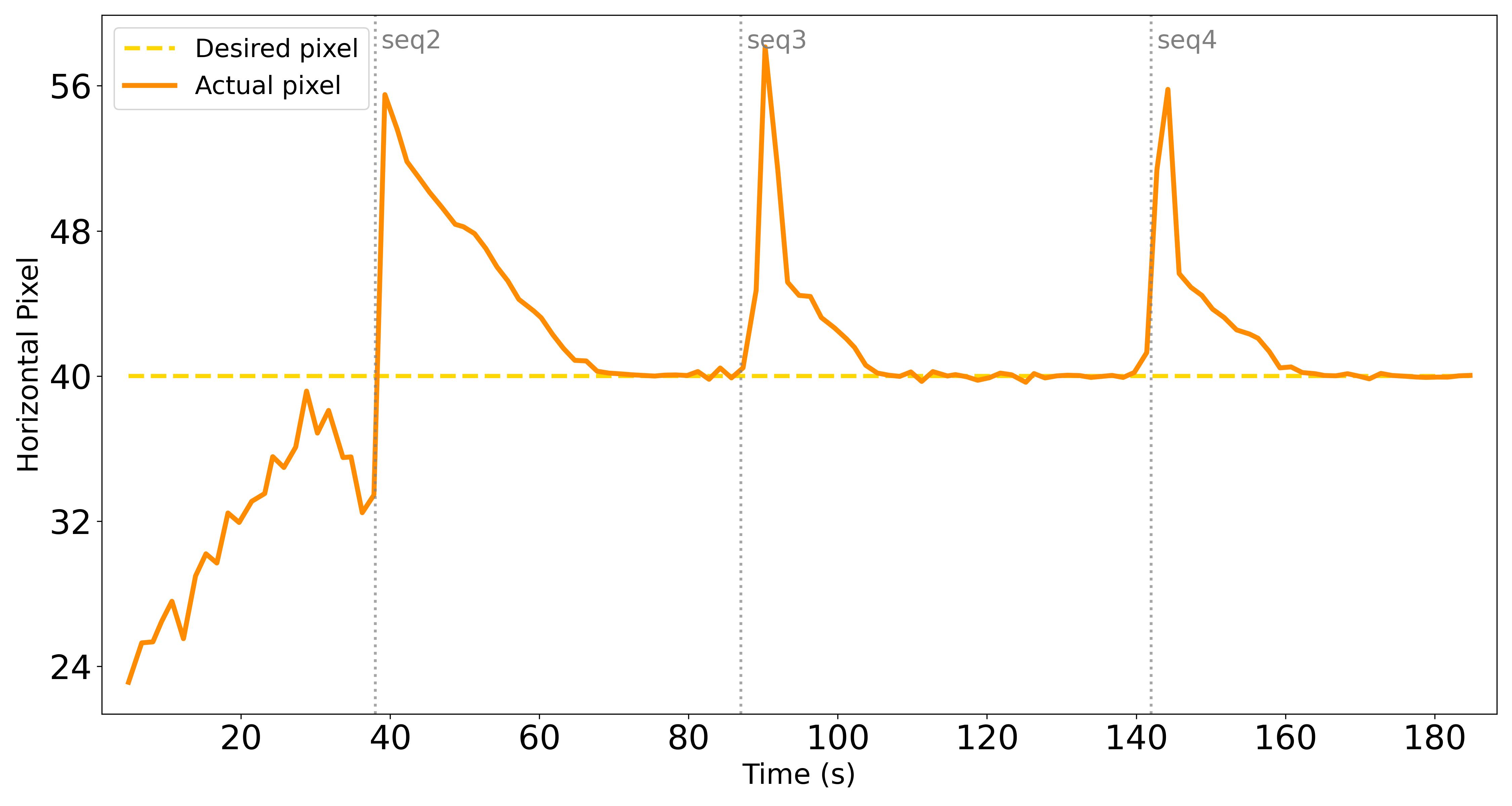}}
    \end{tabular}
    \caption{\footnotesize{\textbf{Experiment 3. Quantitative results.}
    Top-left: Drone--target distance (green) and proximity cost $J_{\mathrm{prox}}$ (blue), illustrating infeasibility in Sequence~1, active cost in Sequence~2, and saturation beyond $d_{\mathrm{ac}}$ in Sequences~3--4.
    Top-right: Evolution of focal length, showing compensation of physical distance to preserve framing.
    Bottom: Desired vs.\ actual horizontal pixel position, demonstrating stable framing.}}
    \label{fig:exp3_quant}
\end{figure}

This experiment tests  the proximity-aware control on real hardware, evaluating distance regulation and zoom compensation under physical sensing and actuation.
The platform demonstrates three key behaviors: increasing distance when operating within the acoustic region, compensating framing via focal-length adjustment, and exhibiting saturation of the proximity cost beyond the acoustic threshold.
Qualitative and quantitative results (Figs.~\ref{fig:exp3_qual} and \ref{fig:exp3_quant}) confirm behavior consistent with simulation, with full experimental runs provided in the accompanying videos.

\begin{figure}[!ht]
\centering
\begin{tabular}{cccc}
      \includegraphics[width=0.30\columnwidth]{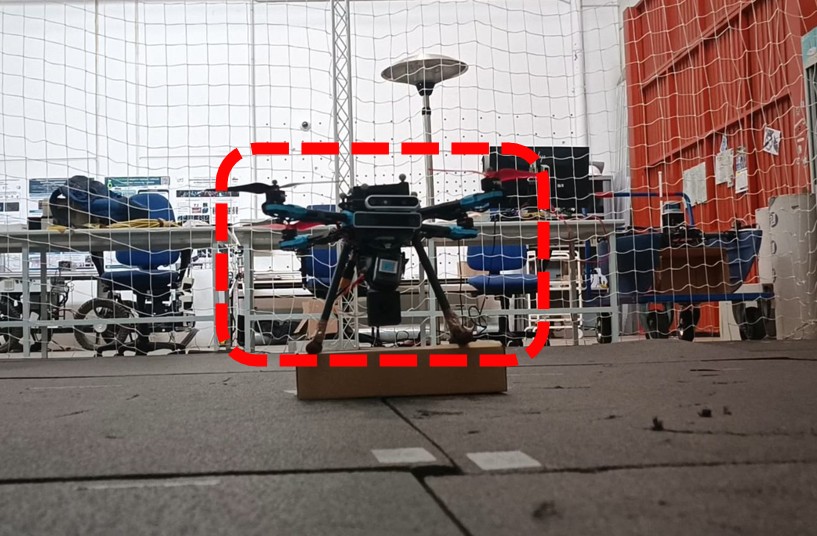} &
    \includegraphics[width=0.30\columnwidth]{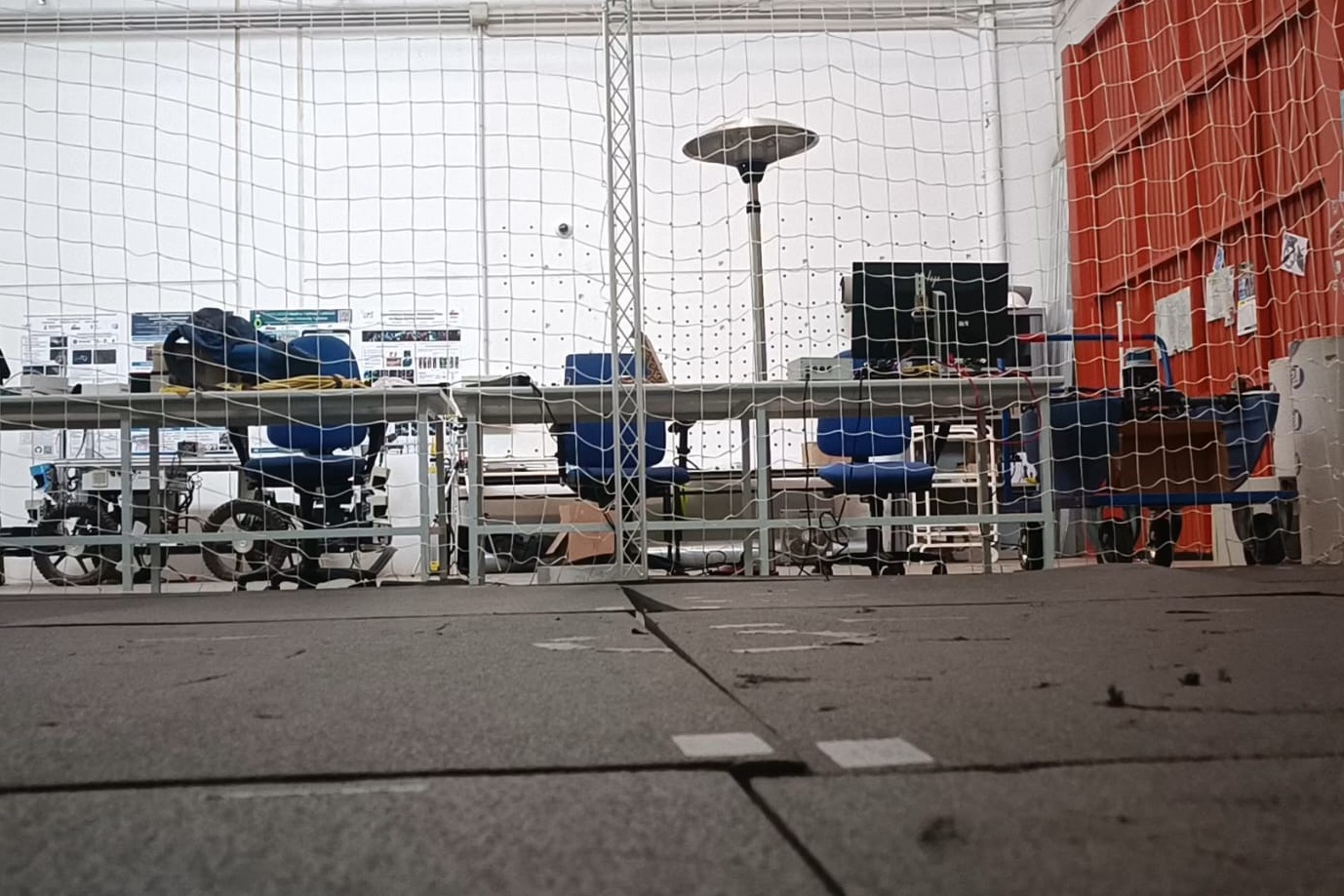} &
    \includegraphics[width=0.30\columnwidth]{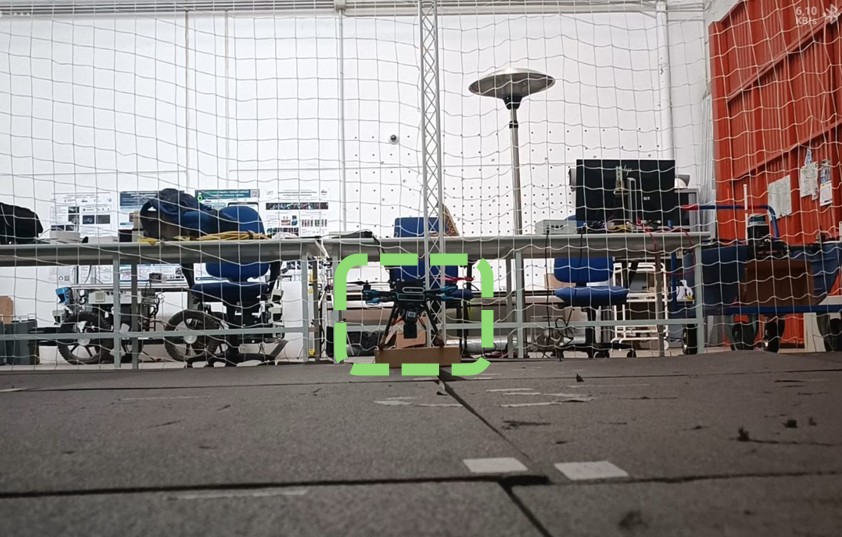} \\
    \includegraphics[width=0.30\columnwidth]{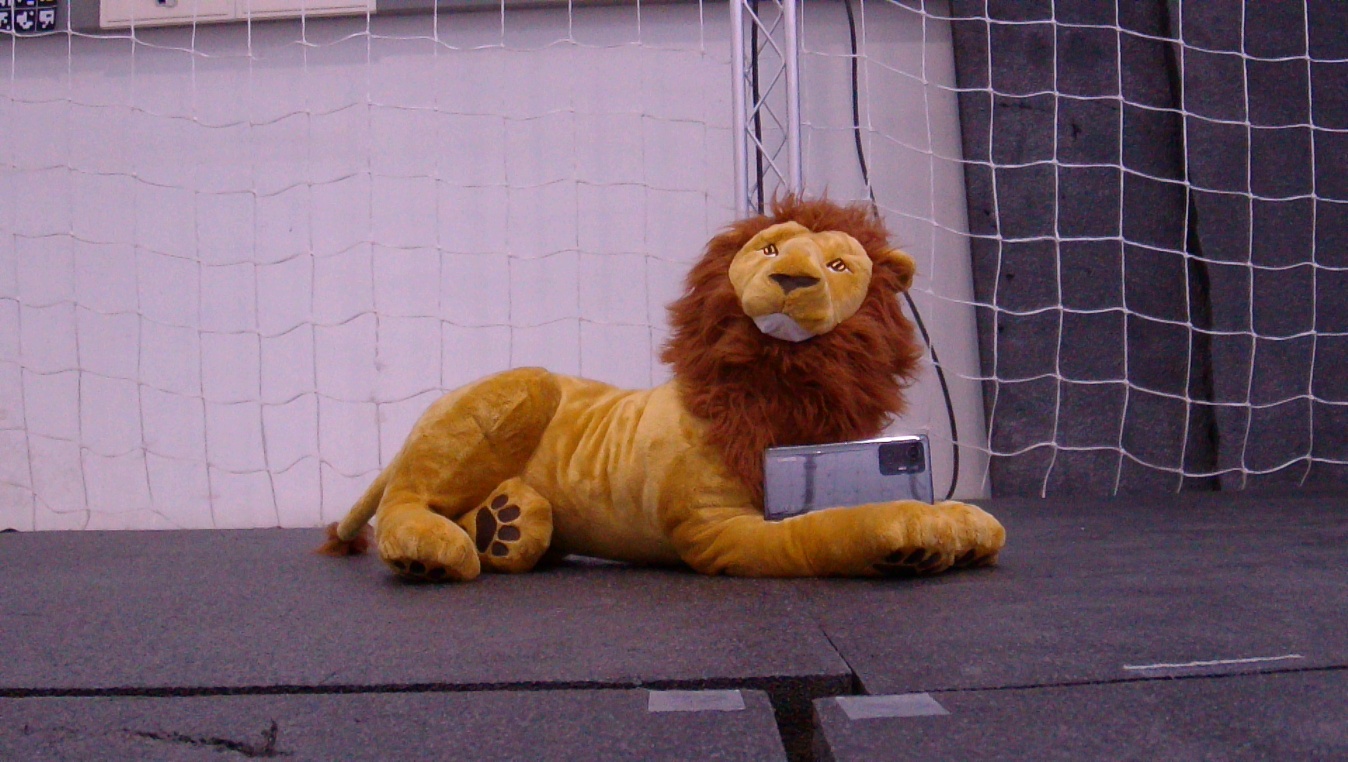} &
    \includegraphics[width=0.30\columnwidth]{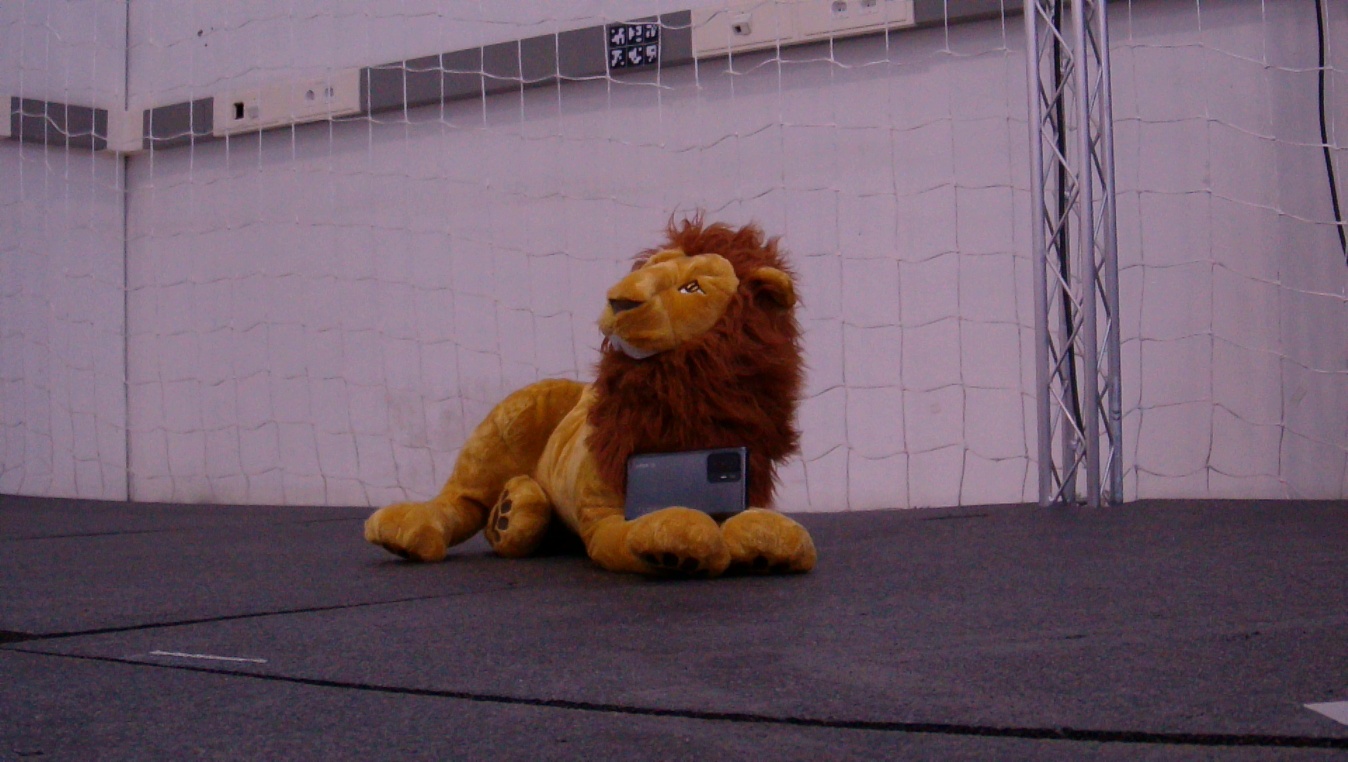} &
    \includegraphics[width=0.30\columnwidth]{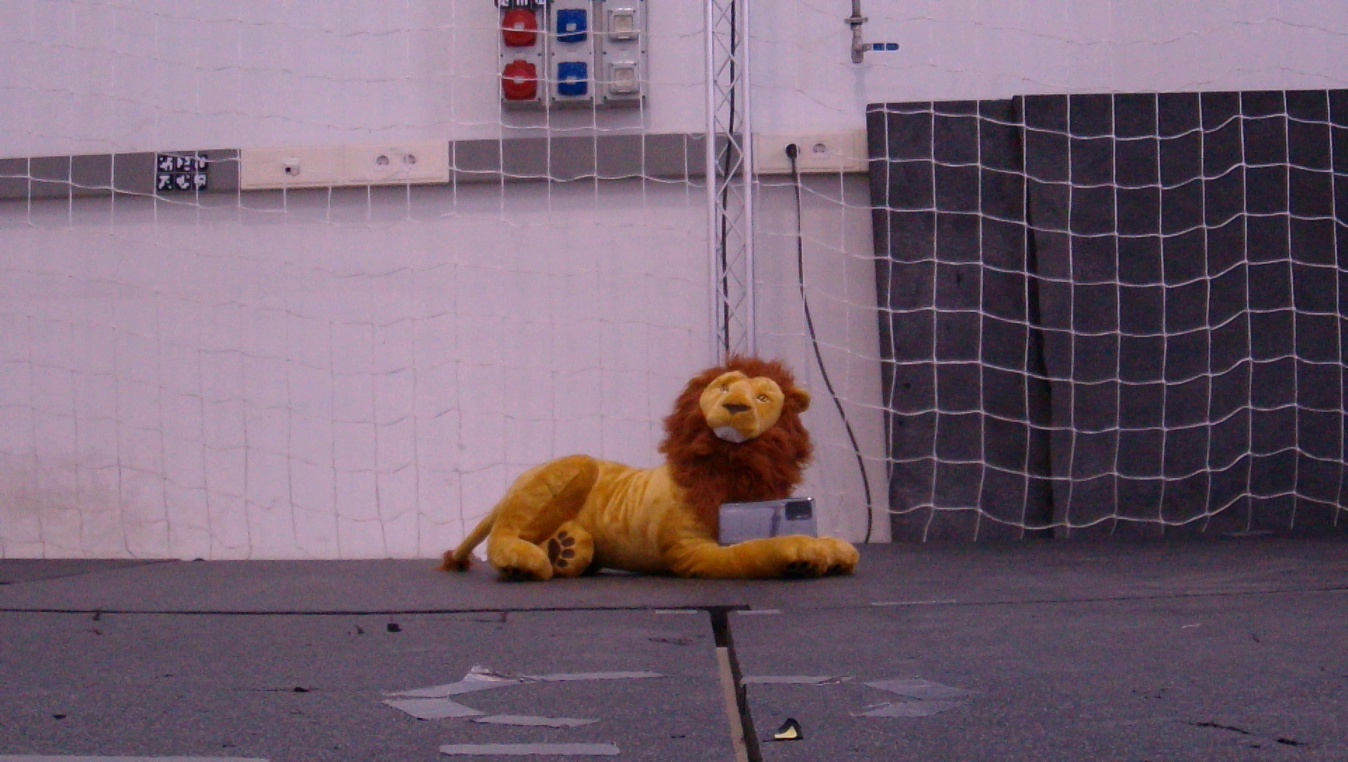} \\

    \includegraphics[width=0.30\columnwidth]{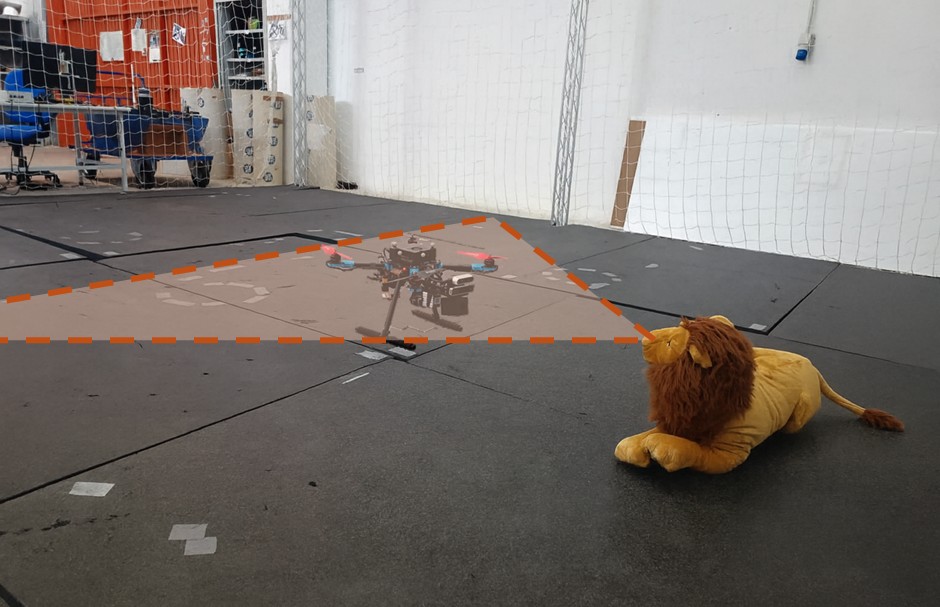} &
    \includegraphics[width=0.30\columnwidth]{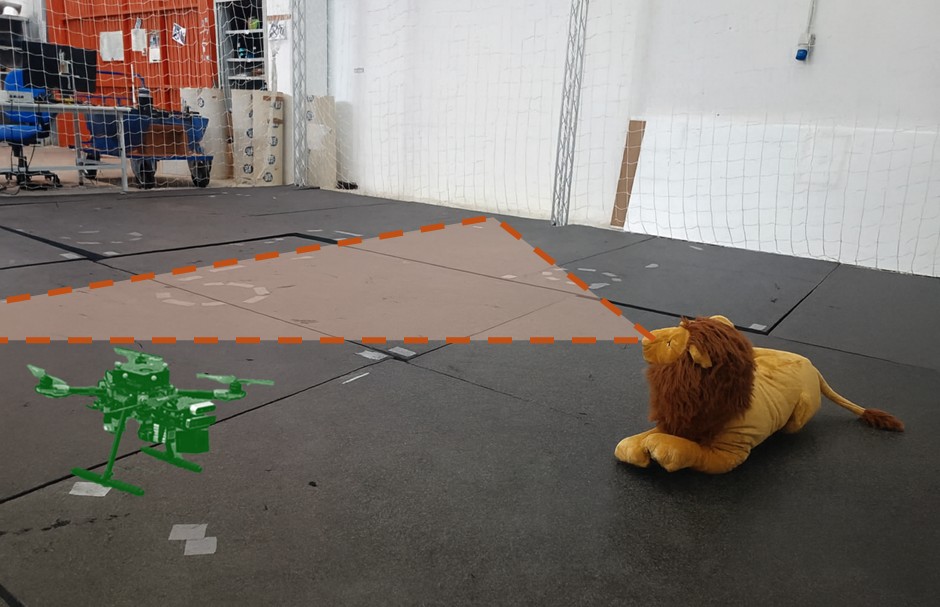} &
    \includegraphics[width=0.30\columnwidth]{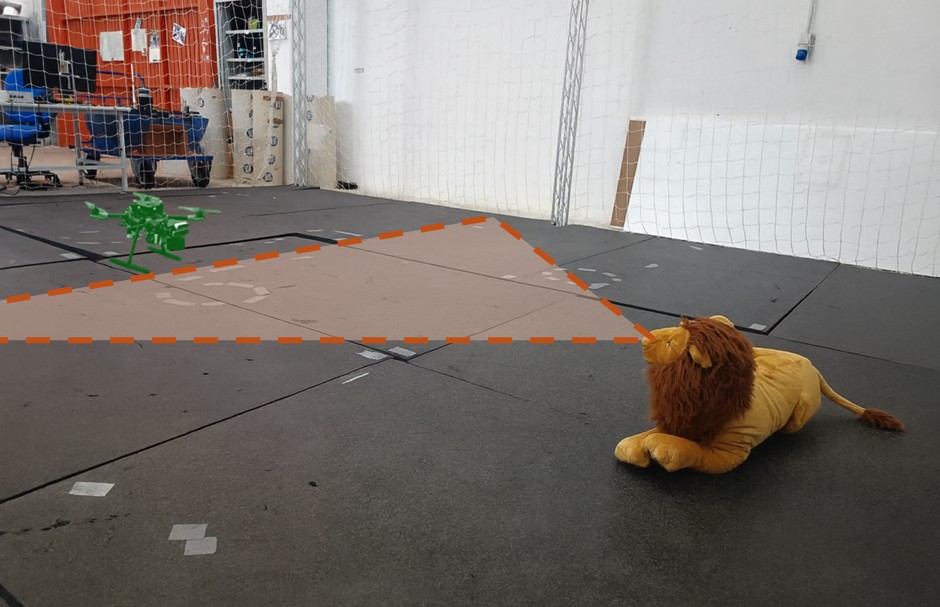} \\
  
\end{tabular}
\caption{\footnotesize{\textbf{Experiment 4 - Real-World Visibility-Aware Behavior. Qualitative results.}
First person view of the animal (first row), onboard camera views from the drone (second row), and third-person views showing the FoV representation (third row) during Experiment~4.
The animal’s field of view is added in post-processing.
The drone is shown in red when inside the FoV and in green when operating in visually safe regions.}}
\label{fig:exp4_qual}
\end{figure}

\begin{figure}[!h]
\centering
\begin{tabular}{cc}
    \includegraphics[width=0.49\columnwidth]{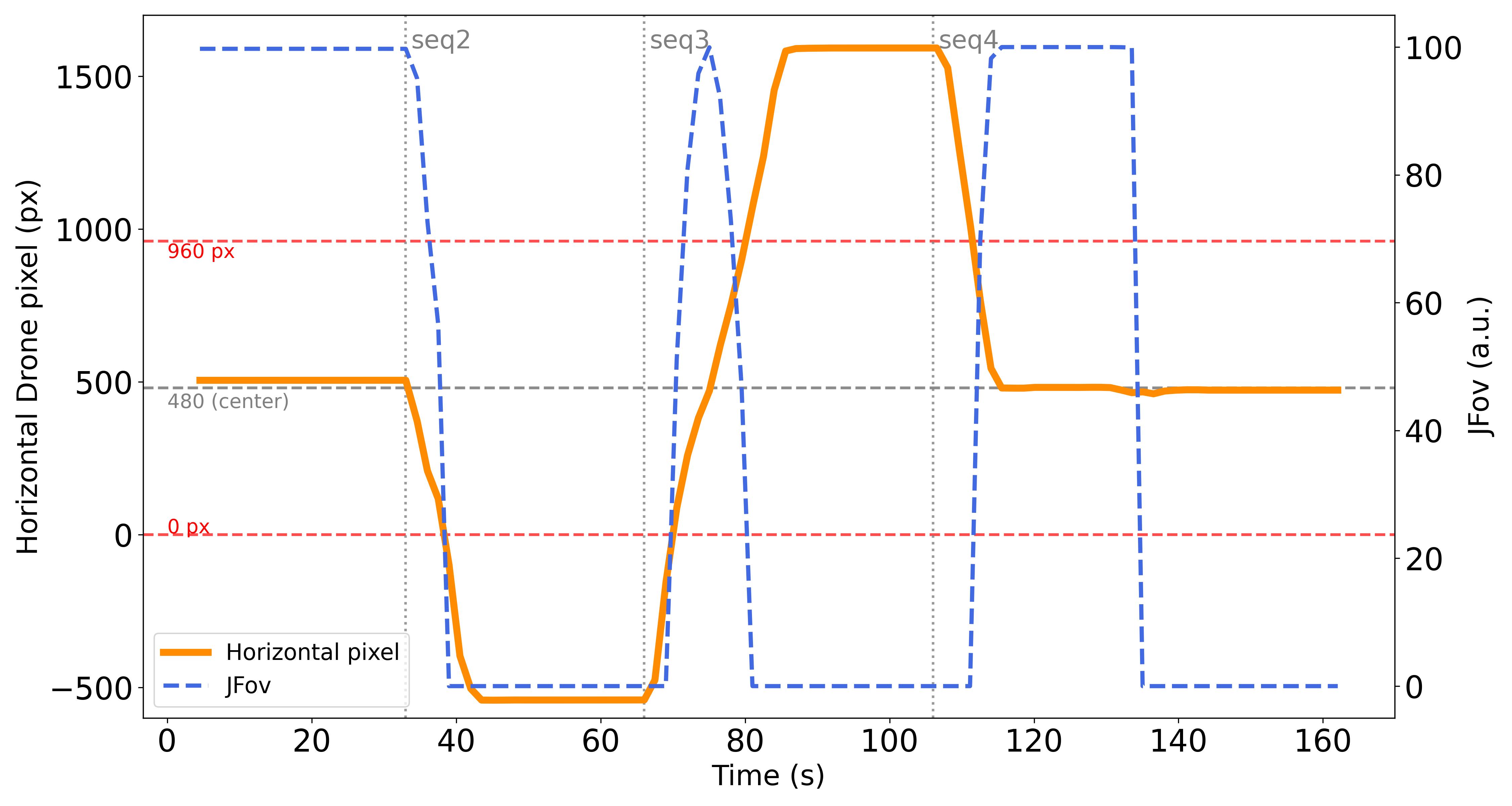} &   
    \includegraphics[width=0.49\columnwidth]{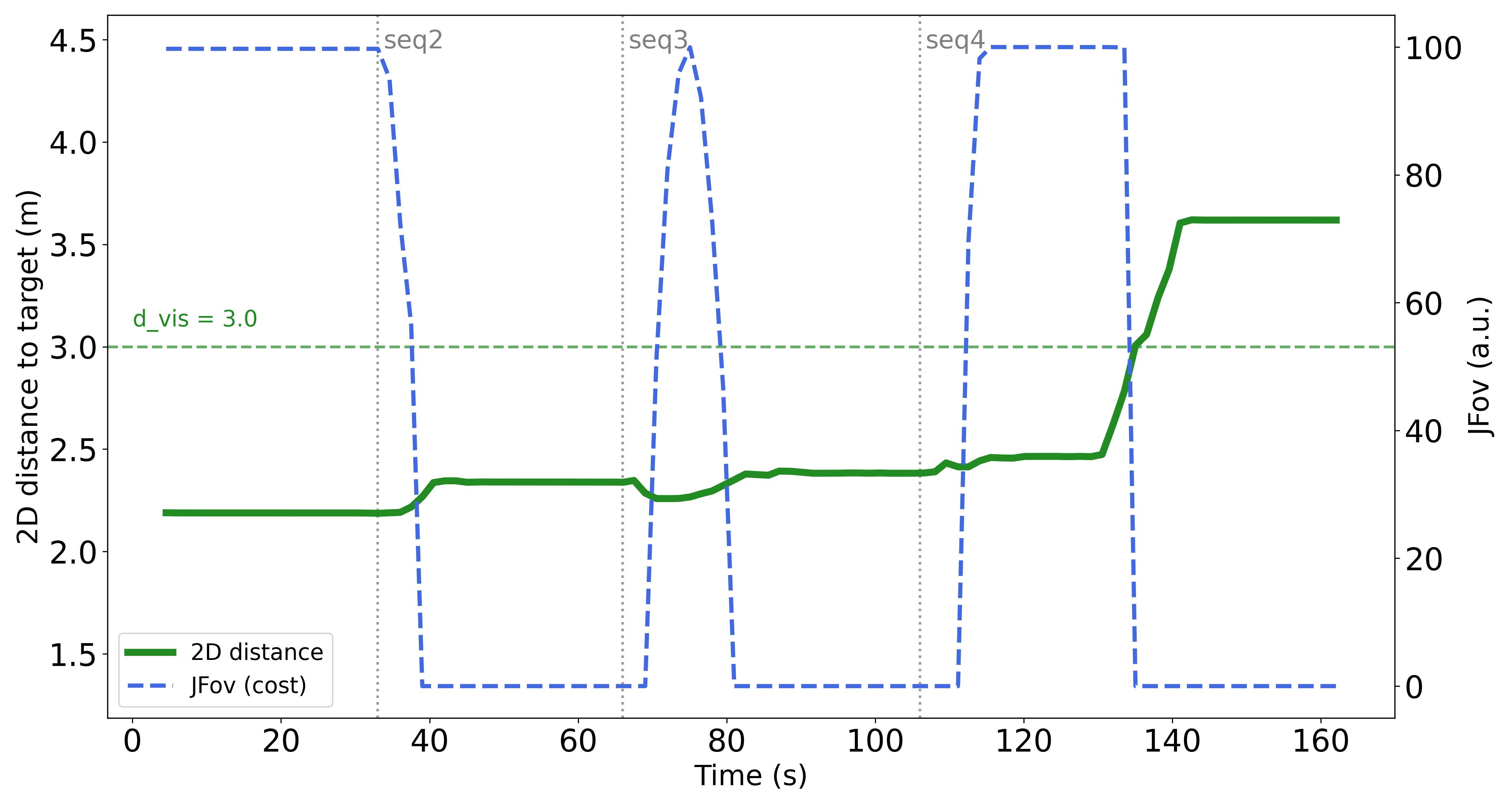} \\

\end{tabular}
\caption{\footnotesize{\textbf{Experiment 4. Quantitative results.}
(left) Horizontal image-plane position of the drone (orange) and visibility cost $J_{\text{FoV}}$ (blue).
(right) Drone--target distance (green) and visibility cost (blue), showing cost increase during FoV intrusions and decrease in safe configurations.}}
\label{fig:exp4_quant}
\end{figure}

\subsubsection{Experiment 4: Real-World Visibility Behavior}

This experiment provides a qualitative hardware validation of the visibility-aware behavior observed in Experiment~2.
Due to indoor space constraints and safety considerations, the hard visibility constraint \rev{$\mathcal{X}_{\mathrm{FoV}}$} is disabled in this experiment, and visual disturbance is regulated only through the soft cost $J_{\text{FoV}}$.
As a result, the strict enforcement of visibility constraints is primarily validated in simulation.
Nevertheless, qualitative results (Fig.\ref{fig:exp4_qual}) and quantitative measurements (Fig.\ref{fig:exp4_quant}) show trends consistent with simulation: the visibility cost increases during field-of-view intrusions and decays as the drone moves toward visually safe regions, confirming consistent controller behavior under real sensing and camera actuation.

\section{Declaration of generative AI}
During the preparation of this work the authors used Claude Code in order to clean the code, fix bugs and prepare some of the included graphs. After using this tool, the authors reviewed and edited the content as needed and take full responsibility for the content of the published article.

\section{Conclusions}
In this paper, we introduced CineWild, a cinematographic drone platform that combines the creative and visual goals of wildlife filmmaking with ethical operational constraints. By integrating model predictive control with disturbance‑aware and cinematographic objectives, CineWild jointly controls drone motion and camera parameters to enable automatic adaptive framing, zoom, and smooth trajectories that respect animal welfare.
Simulation results show that CineWild produces stable, high‑quality cinematography while adhering to non‑invasive guidelines and outperforming \rev{both a cinematographic controller and a safety distance combined with zoom, at the price of reaching some requested viewpoints later}. \rev{The ablation study shows that each welfare term is responsible for its own effect, keeping the distance, staying out of view and smoothing the motion, and the behavior holds across start poses, welfare weights and species thresholds with the same weights in all scenarios, and after a sudden head turn of the animal.} Real‑world proof‑of‑concept experiments further indicate that these principles transfer to physical hardware, with the system \rev{evaluating the disturbance terms from real perception and adapting the zoom to preserve the framing, while the translational commands are executed by a human pilot}.
Future work will focus on field deployments in natural environments \rev{with autonomous flight and real animals, where the disturbance thresholds can be calibrated from the observed responses of each species}, and on extending the framework with improved perception and multi‑drone coordination to support richer visual narratives and broader ecological monitoring. \rev{On the modelling side, we plan angular visibility models for species with panoramic vision, agile targets and groups of animals, and a faster recovery when the animal turns towards the drone.}

\bibliographystyle{elsarticle-num}
\bibliography{refs}

\end{document}